\documentclass[10pt,onecolumn]{article}

\usepackage{geometry}
\usepackage{authblk} % For authors
\usepackage[sort,square,numbers]{natbib}

\usepackage[counterclockwise]{rotating}
\usepackage[T1]{fontenc}
\usepackage{amsmath}
\usepackage{amsfonts}
\usepackage{pifont}
\usepackage{tikz}
\usetikzlibrary{shapes}

\usepackage{hyperref}
\usepackage{array}
\usepackage{subcaption}
\usepackage{graphicx}
\usepackage{enumitem}
\usepackage{multirow}
\usepackage{lscape}
\usepackage{tabularray}
\usepackage{rotating}
\usepackage{booktabs}

\newcommand{\argmax}[1]{\arg \max_{#1}} %{\arg\,\underset{\, #1}{\max} }
\newcommand{\figref}[1]{\hyperref[#1]{\textup{Fig.~\ref{#1}}}}
\newcolumntype{L}[1]{>{\raggedright\let\newline\\\arraybackslash\hspace{0pt}}p{#1}}
\newcommand{\vt}[1]{\rotatebox[origin=c]{90}{#1}}
\newcommand{\nomark}{-}
\newcommand{\cmark}{\ding{51}}
\newcommand{\xmark}{\ding{55}}

\DeclareRobustCommand{\mytriangle}[3]{\tikz{\node[draw=#1,fill=#1,isosceles
triangle,isosceles triangle stretches,shape border rotate=#2,minimum
width=#3cm,minimum height=#3cm,inner sep=0pt] at (0,0) {};}}
\DeclareRobustCommand{\uptri}[1]{\mytriangle{blue}{90}{#1}}
\DeclareRobustCommand{\downtri}[1]{\mytriangle{red}{-90}{#1}}
\newcommand\setItemnumber[1]{\setcounter{enumi}{\numexpr#1-1\relax}}
\renewcommand{\eqref}[1]{Eq.~(\ref{#1})}

\title{Towards Interactive Reinforcement Learning with Intrinsic Feedback}

\author{Benjamin Poole$^1$\thanks{Corresponding author}\,}
\author{Minwoo Lee$^1$}

\affil{$^1$Department of Computer Science \\
University of North Carolina at Charlotte, 9201 University City Blvd, Charlotte NC, 28223}
\date{\textit {\{bpoole16, Minwoo.Lee\}@charlotte.edu}}

\begin{document}

\maketitle

\begin{abstract}
Reinforcement learning (RL) and brain-computer interfaces (BCI) have experienced significant growth over the past decade. With rising interest in human-in-the-loop (HITL), incorporating human input with RL algorithms has given rise to the sub-field of interactive RL. Adjacently, the field of BCI has long been interested in extracting informative brain signals from neural activity for use in human-computer interactions. A key link between these fields lies in the interpretation of neural activity as feedback such that interactive RL approaches can be employed. We denote this new and emerging medium of feedback as \textit{intrinsic feedback}. Despite intrinsic feedback's ability to be conveyed automatically and even unconsciously, proper exploration surrounding this key link has largely gone unaddressed by both communities. Thus, to help facilitate a deeper understanding and a more effective utilization, we provide a tutorial-style review covering the motivations, approaches, and open problems of intrinsic feedback and its foundational concepts.

\end{abstract} 

%%%%%%%%%%%%%%%%%%%%%%%%%%%%%%%%%%%%%%%%%%%%%%%%%%%%%%%%%%%%%%%%%%%%%%%%%%%%%%%%%%%%%%%%%%%%%%%%%%%%%%%%%%%%%%%%%%%%%%%%%%%%%%%%%%%%%%%
\section{Introduction}\label{sec:introduction}

The combination of reinforcement learning (RL) and human-in-the-loop (HITL) ideas through integration of human input into the RL framework has led to the emergence of the field of interactive RL \citep{zhang_irlsurvey_2019, li_irlsurvey_2019, arzate_irlsurvey_2020, najar_irlsurvey_2021}. Interactive RL can be seen as a method for facilitating human-to-agent knowledge transfer. This knowledge transfer inherently helps to alleviate various aptitude and alignment challenges an agent may face. Challenges encompassed by the term \textit{aptitude} correspond to challenges for simply being able to learn. This includes ideas such as \textit{robustness}, the ability of an agent to perform a task (e.g., asymptotic performance) and generalize within/between environments of similar complexity; \textit{scalability}, the ability of an agent to scale up to more complex environments; and \textit{aptness}, the rate at which an agent can learn a desired behavior. On the other hand, \textit{alignment} encompasses challenges concerned with learning as intended (i.e., learning to behave as a human dictates) \citep{amodei_aisafty_2016, leike_alignment_2018, dulacarnold_challenges_2020, gabriel_artificial_2020, rodriguez_instilling_2022}. The hypothetical Paperclip Maximizer \citep{bostrom_ethical_2020} is a classical example of misalignment where an agent maximizes its goal of making paperclips in an unintended manner by turning the world into a paperclip factory. While not as extreme, RL algorithms often display lesser forms of misalignment \citep{amodei_aisafty_2016, leike_alignment_2018, gabriel_artificial_2020}. Therefore, it is not unreasonable to anticipate an increase in misaligned behaviors as the problem complexity increases, i.e., the complexity of the environment and behaviors \citep{leike_alignment_2018}. 

The field of brain-computer interfaces (BCI) aims to develop systems for monitoring the brain and utilizing neural activity as a direct means of communication between the brain and external devices. BCI research has primarily focused on medical applications such as fatigue detection, monitoring sleep, analyzing emotion, and facilitating sensory-motor rehabilitation. However, particular interest has been directed towards addressing paralysis conditions like amyotrophic lateral sclerosis and locked-in-state. These paralysis conditions necessitate innovative technologies that can assist patients in interacting with the external world. Therefore, traditional BCI applications have been primarily concentrated on control problems such as wheelchair and interface control \citep{rashid_bcisurvey_2020, schmidt_mentaltype_2012}. Yet, as the medical BCI research has progressed, it has begun to spill out into the non-medical realm \citep{dabas_bciappsurvey_2020}. Applications in robotics \citep{kirchner_robointerface_2019, wirth_goal_2020, salazar_errprobot_2017}, VR and gaming \citep{kerous_bcigames_2018}, smart environments \citep{kosmyna_bcienv_2016}, bio-metrics and security \citep{gui_bcisecuritysurvey_2019}, and brain-to-brain interfaces \citep{rao_braintobrain_2014, jiang_brainnet_2019} have provided opportunities to further explore this form of intrinsic communication. Moreover, there has been growing interest in using BCI to communicate with artificial agents to facilitate human-AI interactions. One promising avenue involves the integration of RL and BCI, where particular neural activity is interpreted as a reward in the RL framework \citep{iturrate_robot_2010, iturrate_errprl_2015, ehrlich_errpcoadapt_2018, kim_intrinsic_2017}. Interestingly, as we will soon see, this fundamental integration can actually be more broadly formulated using interactive RL concepts.

In the realm of interactive RL, a crucial concern revolves around seamlessly integrating and effectively communicating human intentions to an agent \citep{knox_thesis_2012, li_thesis_2016}. To this end, human input is often categorized into two main channels of communication: advice and demonstrations \citep{najar_irlsurvey_2021}. Of particular interest is a specific type of advice denoted as \textit{feedback}, which aims to convey the "goodness" or "badness" of an agent's behavior. Classically, the de facto medium for conveying feedback has been what we refer to as \textit{explicit feedback}. This type of feedback involves the human communicating directly with the agent via a button push or very basic natural language \citep{knox_interactively_2009, knox_thesis_2012, arumugam_dlcoach_2019, christiano_preferences_2017}. More recently, \textit{implicit feedback} provided through human social cues (e.g., facial expressions, body language, or vocal tone) has begun to see wider use in conveying feedback as it offers a more natural and potentially automatic means of communication \citep{lin_irlsocial_2020}. 

Yet, both explicit and implicit feedback do not cover the entire range of potential human feedback mediums. Another untapped source of valuable information lies in intrinsically occurring and automatically elicited biological signals. These raw and largely unconsciously elicited biological signals have the potential to provide a wealth of information concerning a human's internal state including feelings and even beliefs. Thus, we denote \textit{intrinsic feedback} as a new medium for providing feedback where feedback is implied through a human's intrinsically occurring biological signals\footnote{Originally termed as ``intrinsic interactive RL'' by \citet{kim_intrinsic_2017}, we now extend the term ``intrinsic" to potentially encompass all biological signals, not just brain signals, for the sake of generalization.}. As recent works \citep{xu_accelerating_2021, luo_drlpref_2018, chakraborti_hitlbci_2017, wang_maximizing_2020, akinola_accelerated_2020} have begun to recognize that neural activity can be interpreted more broadly as feedback, we will focus particularly on the brain, the central organ in the human nervous system, as it harbors a dense concentration of biological signals. For brevity, we will henceforth use the terms intrinsic feedback and neurological intrinsic feedback interchangeably. 

\begin{table*}[t]
    \footnotesize
    \caption{Summary of interactive RL reviews categorized by the human input categories of advice and demonstrations  (denoted as ``Demos''). Advice is further broken into categories of feedback and other (i.e., all other types of advice such as general advice, guidance, ect). Feedback is further broken into its three different mediums. A category with a check mark \cmark indicates the review is \textit{primarily} focused on a particular category of human input while no check mark indicates a given category is not discussed or only briefly mentioned. For the reader's reference, we have included a single comprehensive learning from demonstrations review although many more exist.}
    \begin{tblr}{
        width=\textwidth,
        colspec={|X[1.2,l]|X[5,l]|X[0.55,c]|X[0.55,c]|X[0.55,c]|X[0.55,c]|X[0.55,c]|},
        cell{1}{1} = {r=1}{c},
        cell{1}{2} = {r=1}{c},
        cell{1}{3} = {c=4}{c},
        cell{2}{3} = {c=3}{c},
        cell{2}{6} = {r=1}{c},
        cell{1}{7} = {r=1}{c},
        hline{1,4} = {-}{},
        hline{2} = {3-6}{},
        hline{3} = {3-5}{},
    }
        Review
        & Summary
        & Advice & & & 
        & Demos \\ &         
        & Feedback & &          
        & Other & \\ & 
        & \scriptsize{Intrinsic} 
        & \scriptsize{Explicit} 
        & \scriptsize{Implicit} &  & \\
        \citet{osa_imitation_2018}( \citeyear{osa_imitation_2018}) 
        & An in-depth technical review of various learning from demonstration approaches (e.g., behavioral cloning and inverse RL). 
        & -
        & -
        & -
        & -
        &\cmark \\ 
        \hline
        \citet{zhang_irlsurvey_2019}( \citeyear{zhang_irlsurvey_2019}) & A high-level overview of basic interactive RL ideas and approaches.
        & -
        & \cmark
        & -
        & -
        &\cmark \\ 
        \hline
        \citet{li_irlsurvey_2019}( \citeyear{li_irlsurvey_2019})
        & A partially technical overview of feedback approaches and application from the perspective of myopic and non-myopic learning. 
        & -
        & \cmark 
        & \cmark 
        & -
        & - \\
        \hline
        \citet{arzate_irlsurvey_2020}( \citeyear{arzate_irlsurvey_2020})
        & A high-level review directed at human-computer interaction researchers which covers various contextual advice approaches and application design principles. 
        & -
        & \cmark 
        & -
        & \cmark
        & - \\
        \hline
        \citet{najar_irlsurvey_2021}( \citeyear{najar_irlsurvey_2021})
        & Introduces taxonomy for interactive RL and provides a high-level review of learning from advice approaches. 
        & -
        & \cmark 
        & -
        & \cmark
        & - \\
        \hline
        \textbf{Ours} 
        & \textbf{Technical review of intrinsic feedback with a focus on its motivations and foundational concepts. }
        & \textbf{\cmark }
        & \textbf{ \cmark }
        & \textbf{-}
        & \textbf{-}
        & \textbf{-} \\
        \hline
    \end{tblr}
    \label{tab:irl_surveys}
    \normalsize
\end{table*}

By formulating and interpreting neural activity as an alternative feedback medium, we can employ the integrative RL setup by utilizing learning from feedback (LfF) approaches. This key linking idea between the fields of BCIs and interactive RL has motivated the goal of this paper: to provide a tutorial-style review of intrinsic feedback and its foundational concepts (e.g., interactive RL and LfF). Due to the rather chaotic state of the interactive RL field, we believe a unified review that includes foundational concepts will help to more clearly convey intrinsic feedback as a concept. Our hope is that this paper will equip both BCI and RL researchers with the necessary background to apply and build upon existing LfF approaches using intrinsic feedback. To achieve this, we provide the following contributions:
\begin{itemize}
\item We briefly introduce interactive RL and a taxonomy of human input while also elucidating the general motivations behind incorporating human input into the RL framework.
\item We introduce LfF and provide a technical review of its foundational approaches to show how explicit feedback has traditionally been used for learning. Furthermore, we analyze these approaches in terms of their viability, enabling researchers who intend to leverage intrinsic feedback to build upon them while addressing their limitations.
\item We review intrinsic feedback, its motivations, LfF inspired approaches and their viability.
\item We discuss challenges related to intrinsic feedback while also highlighting future directions of interest.
\end{itemize}
Although previous reviews have covered a variety of the interactive RL and explicit feedback literature, as summarized in Table~\ref{tab:irl_surveys}, this paper aims to complement these reviews. By concentrating on concepts related to intrinsic feedback, we offer a narrower and more technical perspective on interactive RL.

As such, this paper is structured as follows. The \nameref{sec:background} section provides a general overview and background for RL and BCI. To build the foundations of intrinsic feedback the \nameref{sec:irl} section provides an overview of the human input taxonomy, with a focus on advice, feedback, and the general motivations for incorporating human input. The \nameref{sec:foundations-LfF} section then provides a technical review of foundational LfF approaches and their validity to demonstrate how feedback can be used for learning. The \nameref{sec:iirl} section then provides an overview of intrinsic feedback, its motivations, what signals can be utilized, along with recent approaches. Finally, the \nameref{sec:challenges} section concludes by discussing challenges and future directions of interest relating to intrinsic feedback.

%%%%%%%%%%%%%%%%%%%%%%%%%%%%%%%%%%%%%%%%%%%%%%%%%%%%%%%%%%%%%%%%%%%%%%%%%%%%%%%%%%%%%%%%%%%%%%%%%%%%%%%%%%%%%%%%%%%%%%%%%%%%%%%%%%%%%%%

\section{Background}\label{sec:background}
The objective of this section is to provide a concise overview of RL and BCI while also introducing notation that will be utilized in the subsequent sections. Each respective section contains additional sources for those readers wishing to pursue particular ideas or concepts in greater detail.

%%%%%%%%%%%%%%%%%%%%%%%%%%%%%%%%%%%%%%%%%%%%%%%%%%%%%%%%%%%%%%%%%%%%%%%%%%%%%%%%%%%%%%%%%%%%%%%%%%%%%%%%%%%%%%%%%%%%%%%%%%%%%%%%%%%%%%%

\subsection{Reinforcement Learning}\label{sec:rl}

\begin{figure}[t]
    \includegraphics[scale=.5]{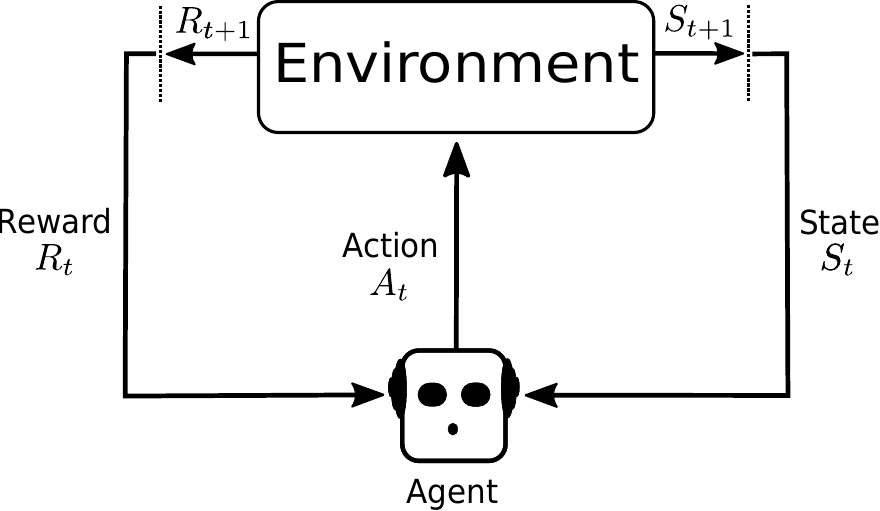}
    \centering
    \caption{Reinforcement learning setup where agent interacts with the environment at every time step $t$ by observing the current state $S_t$ and performing an action $A_t$ which then produces a reward $R_{t+1}$ and a new state $S_{t+1}$.}
    \label{fig:rl_formulation}
\end{figure}

Reinforcement learning (RL) explores how agents ought to make a sequence of decisions by discovering a mapping of actions to observations through the maximization of a numerical reinforcement signal \citep{sutton_rl_2018} as depicted in \figref{fig:rl_formulation}. RL can be formalized as a Markov decision process (MDP), given by the tuple $(\mathcal{S}, \mathcal{A}, r, p, \gamma)$. We denote the set of all possible states (i.e., observations), actions, and rewards as $\mathcal{S}$, $\mathcal{A}$, and $\mathcal{R}$. Discrete time steps are used to divide states from one another where at each time step $t$, the agent perceives state $S_t \in \mathcal{S}$, takes an action $A_t \in \mathcal{A}(s)$ and then receives a real-valued reward $R_{t+1} \in \mathcal{R}$. When time $t$ is not specified, $s$ denotes a state, $a$ denotes an action, $s'$ denotes the next state and $r$ denotes the reward received. Furthermore, The reward function $r: \mathcal{S} \times \mathcal{A} \times \mathcal{S} \rightarrow \mathbb{R}$ is denoted as $r(s, a)$. The transition function $p: \mathcal{S} \times \mathcal{A} \times \mathcal{S} \rightarrow  \mathbb{R}$ is denoted as $p(S_{t+1} = s' | S_t = s, A_t = a)$ and articulated as the probability of transitioning to the next state $S_{t_+1}$ given the current state $S_t$ and action $A_t$. The transition function is used to model the environment's dynamics (i.e., the probability of transitioning to another state). Algorithms can either attempt to model these dynamics (i.e., model-based) or completely forgo them (i.e., model-free). Finally, the discount factor $\gamma \in [0, 1]$ is used to determine how the agent values future reward.

Given the formal definition of an MDP, the agent's goal is to maximize the total discounted return $G_t$ from the current time step $t$:
\begin{equation}\label{expected_discounted_return}
G_t = R_{t+1} + \gamma R_{t+2} +  \gamma^2 R_{t+3} + ... = \sum_{k=0}^{\infty} \gamma^k R_{t+k+1},
\end{equation}
where $\gamma$ is a discounting factor. Here $\gamma$ represents the uncertainty in the model as the sum reaches further into the future and to ensure the math remains bounded. Given \eqref{expected_discounted_return}, as $\gamma \rightarrow 0$ the agent becomes myopic and begins to only value immediate reward. Likewise, as $\gamma \rightarrow 1$ the agent becomes farsighted and begins to value all future rewards equally.

Maximizing $G_t$, the agent learns behaviors represented as a state-to-actions map. This mapping defines the policy $\pi: S \times A \rightarrow \mathbb{R}$, where $\pi(a | s)$ denotes the probability of taking an action in a given state under the policy $\pi$. Therefore, a policy $\pi$ determines how the agent should behave in the given environment. One approach to learn $\pi$ is known as value-based learning or generalized policy iteration \citep{sutton_rl_2018}, which involves evaluating states or state-action pairs. The state-value function denoted as $V_\pi(s)$ estimates the expected discounted return of state $s$, thenceforth following policy $\pi$ \citep{sutton_rl_2018}, such that
\begin{equation}\label{eq:state_value_function_bellman}
V_{\pi}(s) = \mathbb{E}_{\pi}  \big[ R_{t+1} + \gamma V_{\pi}(S_{t+1}) \, \big | \, S_t = s \big],
\end{equation}
where $R_{t+1}$ is the most recently received reward and $\gamma V_{\pi}(S_{t+1})$ is the estimated value of the future discounted return when starting from the next state. Alternatively, the action-value function (i.e., Q-function or Q-values) denoted as $Q_\pi(s, a)$ estimates the expected discounted return of state $s$ when taking action $a$ and thenceforth following policy $\pi$ \citep{sutton_rl_2018}, such that
\begin{equation}\label{eq:action_value_function_bellman}
Q_{\pi}(s, a) = \mathbb{E}_{\pi}  \big[ R_{t+1} + \gamma Q_{\pi}(S_{t+1}, A_{t+1}) \, \big | \, S_t = s, A_t = a \big].
\end{equation}

A popular partial sampling method for updating state-value or action-value functions is referred to as Temporal Difference (TD) learning. TD learning is considered a bootstrapping method as it learns a value function using the current estimate of the value function. The following formalizes the TD error using a state-value function:
\begin{equation}\label{eq:temporal_difference_error}
    \delta_t = R_{t+1} + \gamma V(S_{t+1}) - V(S_t),
\end{equation}
where same formalization applies to action-value function. The TD error can be interpreted as measuring the difference between the next state's value along with the reward received $R_{t+1} + \gamma V(s_{t+1})$ and the current state's estimated value $V(s_t)$. By minimizing the TD error, the agent learns a (sub-)optimal policy, which leads to greedy action selection as follows:
\begin{equation}\label{eq:greedy}
    \pi(s) =\argmax{a}Q(s, a).
\end{equation}

On the other hand, policy gradient methods directly optimize the parameterized policy $\pi_\theta$ using the policy gradient theorem \citep{sutton_rl_2018}:
 \begin{equation}\label{eq:actor_update}
\theta_{t+1} = \theta_t + \alpha \nabla \log \pi_{ \theta_t}(A_t|S_t) V(S_t).
\end{equation}
To further decrease the variance of the update, an advantage function $\mathbb{A}(s, a)$ can be used to replace $V(S_t)$ such that
\begin{equation}\label{eq:advantage_function}
\mathbb{A}(s, a) =  Q(s, a) - V(s).
\end{equation}
 The advantage function can be interpreted as computing whether the selected action is performing better or worse than expected \citep{sutton_rl_2018}. \eqref{eq:actor_update} can be rewritten using the advantage function as 
\begin{equation}\label{eq:actor_advantage_update}
\theta_{t+1} = \theta_t + \alpha \nabla \log \pi_{\theta_t}(A_t|S_t) \mathbb{A}(S_t, A_t).
\end{equation}

Finally, value-based and policy gradient methods can be combined using actor-critic methods where the critic estimates the value using TD learning and the actor learns a policy using the policy gradient theorem. For readers who are not familiar with RL, \citet{sutton_rl_2018} offer a more comprehensive review of these methods.

%%%%%%%%%%%%%%%%%%%%%%%%%%%%%%%%%%%%%%%%%%%%%%%%%%%%%%%%%%%%%%%%%%%%%%%%%%%%%%%%%%%%%%%%%%%%%%%%%%%%%%%%%%%%%%%%%%%%%%%%%%%%%%%%%%%%%%%

\subsection{Brain-Computer Interfaces}\label{sec:bci}

BCI allows for monitoring, communication, and translation of brain signals elicited by cognitive and sensory-motor processes. While we provide a brief overview of a typical BCI pipeline and various informative brain signals utilized by BCI applications, we will defer readers to the plethora of BCI reviews covering concepts from signals and acquisition \citep{ramadan_bcisurvey_2017, gu_bcisurvey_2020}, to classification techniques \citep{lotte_bcisurvey_2018, zhang_bcisurvey_2019, craik_bcisurvey_2019}, to general reviews \citep{tiwari_bcisurvey_2018, rashid_bcisurvey_2020}, and future directions of the field \citep{vansteensel_questionnaire_2017}. 

\begin{figure*}[t]
    \centering
    \includegraphics[scale=.28]{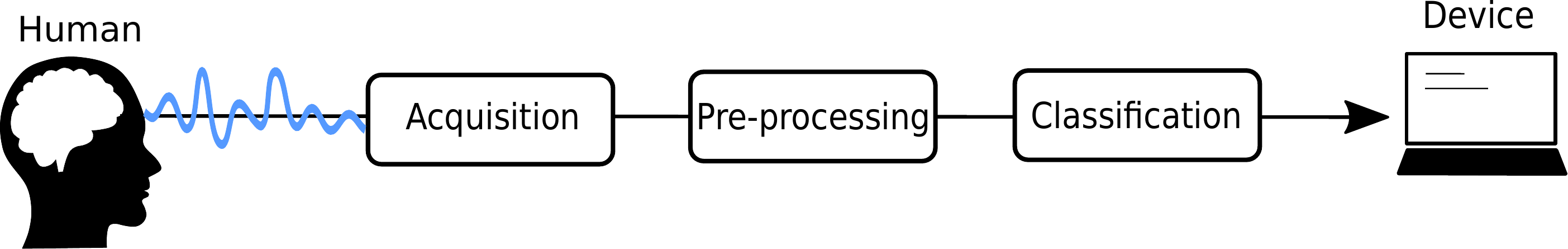}
    \caption{An example of a simplistic BCI pipeline which includes the stages of acquisition, pre-processing, and classification before being passed to an external device (e.g., computer) for utilization.}
    \label{fig:bci_pipeline}
\end{figure*}

Most BCI applications follow the \textit{BCI pipeline} which roughly outlines the steps required to acquire, pre-process, and decode neural activity, as seen in \figref{fig:bci_pipeline}. The first stage of the pipe is signal acquisition. Acquisition of informative brain signals involves the monitoring and recording of neural activity. BCI acquisition is subdivided into non-invasive, invasive, and semi-invasive methods. Each method is defined by where the neural activity is being recorded in relation to the cortex. Non-invasive methods involve monitoring neural activity from outside the skull and scalp. In particular, we will focus on electroencephalographs (EEG) throughout this paper. This is because EEGs are the most common non-invasive device as they have desirable properties such as being safe to use, portable, cost efficient, and relatively easy to work with \citep{tiwari_bcisurvey_2018, gu_bcisurvey_2020, rashid_bcisurvey_2020}. EEGs utilize electrodes that measure electrical activity which orientates from the summation of many postsynaptic potentials (PSPs) that occur simultaneously. PSPs are generated when neurotransmitters bind to receptors, thus causing a change in the flow of ions across the cell membranes. 

\begin{figure*}[t]
    \centering
    \begin{minipage}[c][9cm][c]{0.53\textwidth}
        \centering
            \includegraphics[scale=.50]{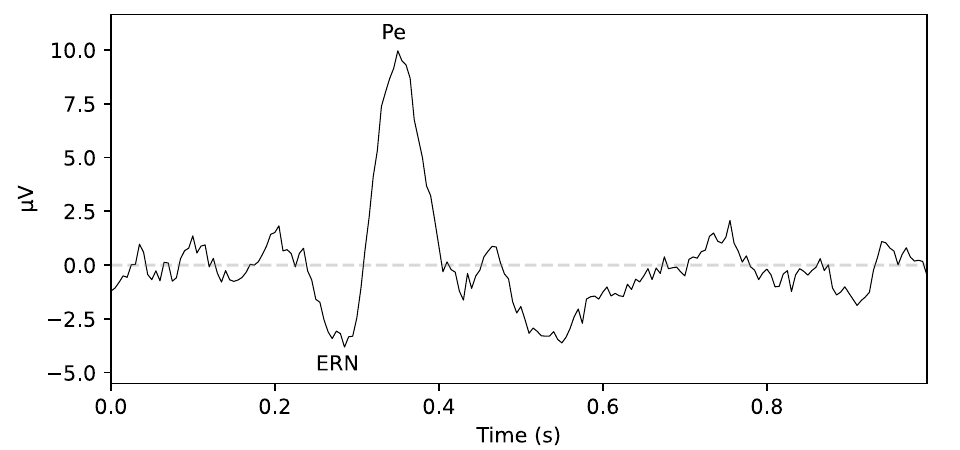}
            \subcaption{Time domain}
            \label{fig:time-domain}
            \includegraphics[scale=.52]{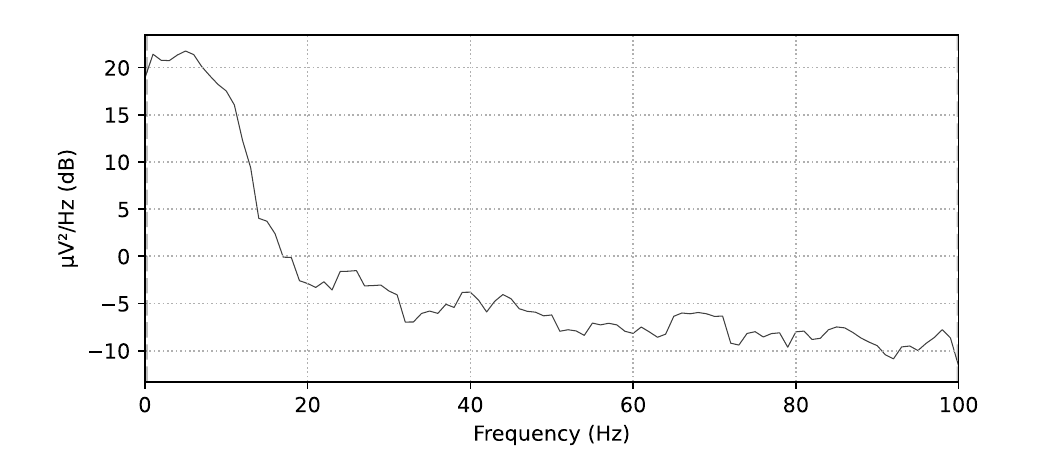}
            \subcaption{Frequency domain}
            \label{fig:freq-domain}
    \end{minipage}%
    \begin{minipage}[c][9cm][c]{0.45\textwidth}
        \centering
        \hspace{10cm}
        \begin{center}
            \includegraphics[scale=.5]{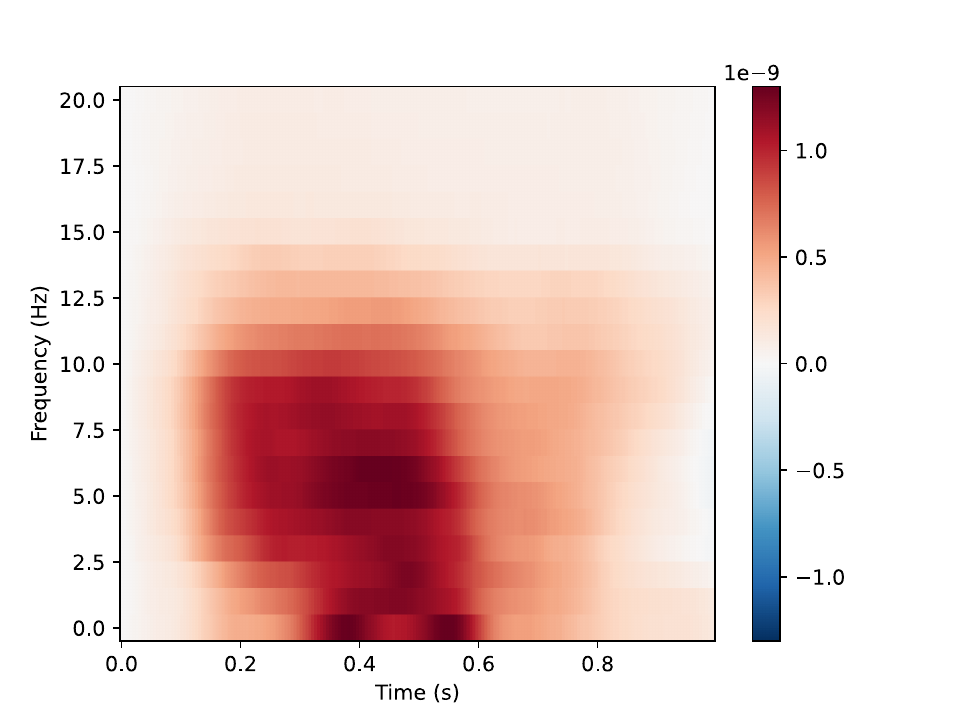}
            \subcaption{Time-Frequency domain}
            \label{fig:time-freq-domain}
        \end{center}
    \end{minipage}
    \caption{Provides an example of a subject's error-related potential grand average (i.e., average across many trial) in the time, frequency and time-frequency domains. The grand average is computed using 185 trails from subject 1's BGSInt task in the dataset provided by \citet{poole_errp_2022}. (\subref{fig:time-domain}) shows the time domain features for ErrP grand average where the ERN peak is given around 250 ms and the Pe peak is given around 350 ms. (\subref{fig:freq-domain}) shows the frequency domain features for the ErrP grand average.  (\subref{fig:time-freq-domain}) shows the time-frequency domain features for the ErrP grand average.}
    \label{fig:signal-domains}
\end{figure*}

As shown in \figref{fig:signal-domains}, this captured electrical activity can be represented either in the time domain, frequency domain, or both (i.e, time-frequency domain). EEGs are characterized by their high temporal resolution but poor spatial resolution. High temporal resolution is desirable for most BCI applications as it allows for changes in neural activity to be captured at the level of milliseconds \citep{luck_erpbook_2014}. The next stage of pre-processing serves to filter and extract latent features associated with the desired brain signal of interest in order to increase the signal-to-noise ratio (SNR). To do so, two different methods are often employed: temporal filtering and feature extraction \citep{luck_erpbook_2014, lotte_tutorial_2014, lotte_bcisurvey_2018}. The last stage of classification consists of decoding informative brain signals from the general neural activity which is often done through employing various machine learning methods \citep{lotte_bcisurvey_2018, zhang_bcisurvey_2019, craik_bcisurvey_2019, yger_rgbcisurvey_2017}.

While the BCI pipeline is highly modular, allowing for the introduction or removal of various different steps, the core of BCI pipeline revolves around the extraction and utilization of informative brain signals. Therefore, it is worth briefly exploring a few different types of brain signals that are could be pertinent to intrinsic feedback.

Event-related potentials (ERPs) are transient brain signals characterized by their unique series of positive and negative peaks that are elicited due to changes over time in scalp-recorded voltage. Elicitation of ERPs can reflect sensory, cognitive, affective, or motor processes in response to either internal (e.g., emotional states) or external stimuli (e.g., perceiving an erroneous behavior) \citep{luck_erpbook_2014}. From a general perspective, ERPs can be seen as the brain's automatic response to a given stimulus or event. ERPs are composed of ERP components which are associated with cognitive processes. An ERP component is characterized by its positivity or negativity at a given point in time. However, this does not mean that there is a one-to-one relationship between ERP components and the observed polarity at a given point in time \citep{luck_erpbook_2014}. As mentioned before, this is due to the fact that many known other components, artifacts, and various forms of noise are all occurring simultaneously. Thus, the observed component is always influenced by these other sources.

A commonly occurring ERP component is P3 or P300 that is characterized by a large positive peak that occurs approximately 300 ms after the stimulus. The P3 component is thought to be associated with cognitive processes such as context updating, surprise, and attention \citep{polich_p3review_2007, luck_erpbook_2014}. Rare-frequent tasks (e.g., the oddball paradigm) are a classical example of P3 elicitation. These tasks entail the repeated elicitation of frequent and infrequent stimuli, where the P3 amplitude inversely scales with the rarity of the infrequent stimulus \citep{polich_p3review_2007, luck_erpbook_2014}. P3 is often applied to decision making or interface control tasks like the P3 speller \citep{farwell_p3_1988}. 

Error-related negativity (ERN) is another ERP component that is associated with errors and occurs approximately 200 ms after a stimulus \citep{falkenstein_ern_2000}. Experiments frequently use ERN to determine whether a subject has either committed or observed an erroneous event \citep{chavarriaga_errplearning_2010, ferrez_errpwrong_2005, salazar_errprobot_2017}. ERPs which contain different variations of the ERN component are referred to as error-related potentials (ErrPs). ErrPs are characterized by a negative peak (i.e., ERN or Ne) that occurs between 80-300 ms and a positive peak (i.e., Pe) that occurs between 200-500 ms as seen in \figref{fig:time-domain} \citep{abu_errpinvariance_2017, falkenstein_ern_2000, luck_erpbook_2014, van_ern_2004, holroyd_frn_2002}. ErrPs are frequently used to provide feedback to an interface or an agent regarding erroneous events \citep{chavarriage_errpbci_2014, schmidt_mentaltype_2012}.

Lastly, extracting cognitive and affective states (e.g., fatigue, mental workload, valence-arousal, or attention) from neural activity has been a popular choice for many BCI applications \citep{arico_pbci_2018, alnafjan_emotionsurvey_2017}. Doing so often entails the utilization of frequency band features where each state is associated with a different combination of bands \citep{li_emotionsurvey_2022}. For instance, \citet{myrden_cognitivestates_2017} find that attention is associated with posterior alpha band activity while frustration is associated with posterior alpha and frontal beta band activity.

%%%%%%%%%%%%%%%%%%%%%%%%%%%%%%%%%%%%%%%%%%%%%%%%%%%%%%%%%%%%%%%%%%%%%%%%%%%%%%%%%%%%%%%%%%%%%%%%%%%%%%%%%%%%%%%%%%%%%%%%%%%%%%%%%%%%%%%

\subsection{Notation}
Table~\ref{tab:notation} summarizes the common notation that will be used throughout the paper.

\begin{table}[h!]
    \footnotesize
    % \captionsetup{singlelinecheck = false}
    \centering
    \caption{Notation summary.}
    % \begin{tabular*}{\linewidth}{@{\extracolsep{\fill}} l|l|l|l} 
    \begin{tabular}{l|p{2cm}|l|p{4cm}} 
        \toprule
        $s$          & a state                & $s'$           & a next state  \\
        $a$          & an action              &  $r$           & a reward \\
        $t$          & discrete time step     & $R_{t+1}$      & reward at time $t+1$ due to $S_{t}$ and $A_{t}$ \\
        $S_t$        & state at time $t$      & $S_{t+1}$      & next state at time $t+1$  \\ 
        $A_t$        & action at time $t$     & $H_t$          & human feedback at time $t$  \\
        $\tau$       & trajectory             & $\sigma$       & trajectory segment \\
        $\pi$        & policy                 & $\pi(a | s)$ & probability of taking an action in state $s$ under a policy $\pi$  \\
        $\mathbb{A}$   & advantage function & $\mathbb{H}$ & entropy \\
        $h$          & human feedback &  $\hat{h}$    & modeled human feedback\\ 
        $\pi_{h}$ & human policy & &\\
        \bottomrule
    \end{tabular}
    \label{tab:notation}
    \normalsize
\end{table}

%%%%%%%%%%%%%%%%%%%%%%%%%%%%%%%%%%%%%%%%%%%%%%%%%%%%%%%%%%%%%%%%%%%%%%%%%%%%%%%%%%%%%%%%%%%%%%%%%%%%%%%%%%%%%%%%%%%%%%%%%%

\section{Interactive Reinforcement Learning}\label{sec:irl}

\begin{figure}[t!]
    \centering
    \includegraphics[scale=.4]{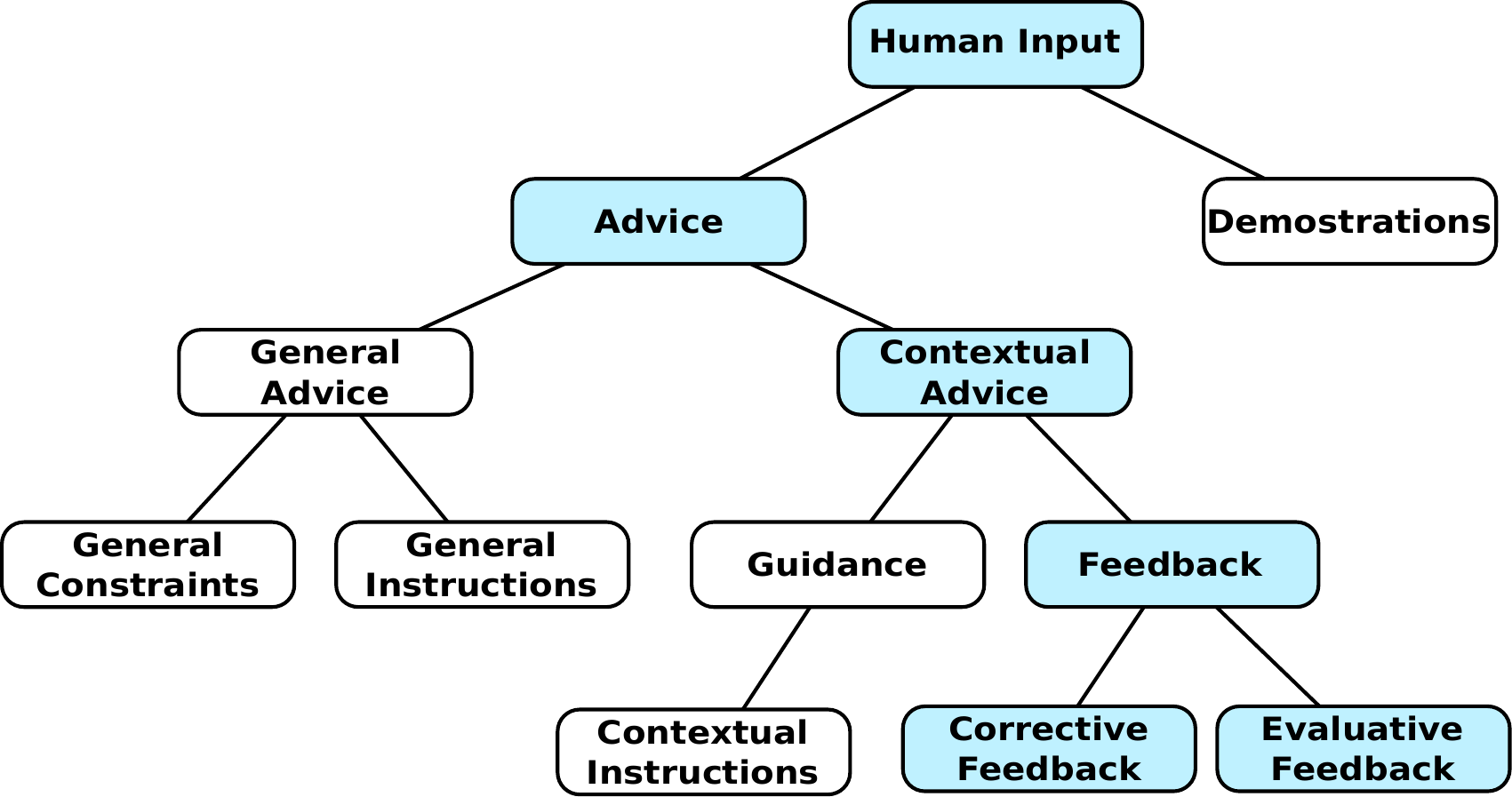}
    \caption{Taxonomy of human input for interactive RL \citep{najar_irlsurvey_2021}. To emphasize feedback, the communication channel of interest for this paper, the path through the taxonomy to reach feedback is highlighted.}
    \label{fig:irl_tax}
\end{figure}

HITL systems leverage the capabilities of humans in order to incorporate human input into intelligent agents. Thus, the objective of interactive RL is to integrate human input directly into the RL framework, achieving an effective HITL setup. By doing so, interactive RL inherently enhances the human-to-agent knowledge transfer process. Thus, we define \textit{human input} as any process or act that assists an agent in its learning, allowing for a transfer of knowledge from the human to the agent. As the terminology of the interactive RL fields has been quite chaotically defined, we mostly follow the taxonomy of \citet{najar_irlsurvey_2021} as they present a recent attempt to better organize the field. Similar to how humans communicate with one another, human input is categorized into the two main communication channels: demonstrations (i.e., showing) and advice (i.e., explaining or critiquing). This taxonomy of human input is illustrated in \figref{fig:irl_tax}. Notice that the demonstration channel has no sub-taxonomy while the advice channel has a more complex and nuanced sub-taxonomy to encompass the diverse methods humans employ for explaining and critiquing.

\begin{figure}[t!]
    \centering
    \includegraphics[scale=.5]{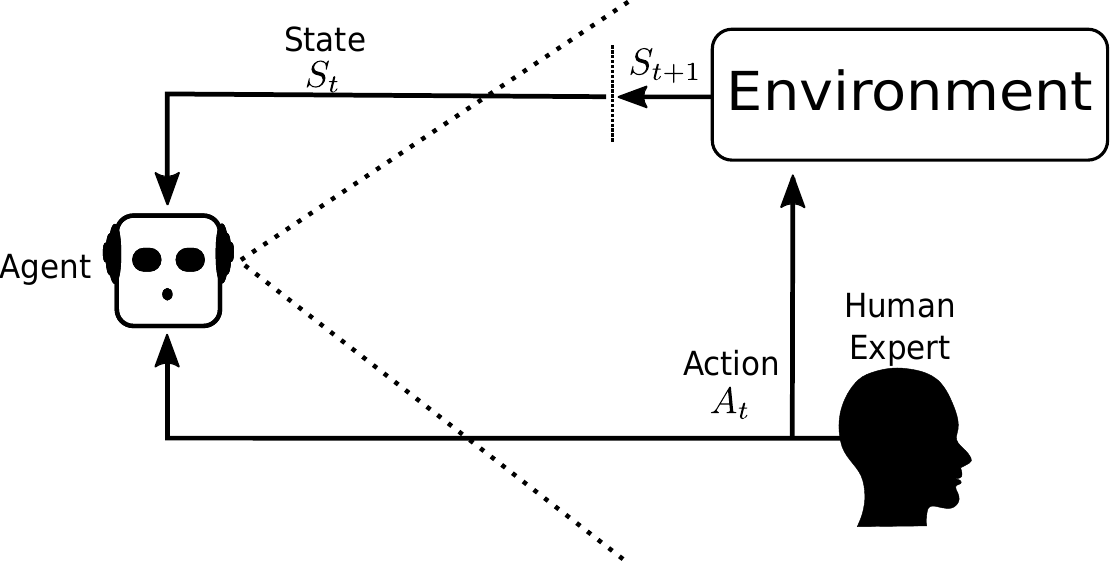}
    \vspace{-0.2cm}
    \caption{The interactive RL setup where human input is provided through demonstrations. Notice that the reward is not included as it is typically not available.
    }
    \label{fig:lfd-all}
\end{figure}

The demonstration channel encompasses information that is communicated by having a human show the agent how to perform a behavior by actually executing the task. Learning from demonstrations (LfD) or imitation learning is a relatively established field that focuses on learning from "expert" demonstrations using methodologies such as behavior cloning, inverse learning, or offline learning \citep{osa_imitation_2018}. For later comparison with advice setups, \figref{fig:lfd-all} shows how the classical RL setup, given in \figref{fig:rl_formulation}, is modified to depict the LfD setup where the agent can be thought of as observing and learning from the human interactions with the environment. 

Conversely, this paper will primarily concentrate on the advice channel, within which feedback serves as a sub-category. The advice channel encompasses any information that is communicated to an agent by the human where the human does not actually have to execute the task. To reach the sub-category of feedback, advice is first broken into the two sub-categories of general advice and contextual advice. Here general advice is any form of general information that can be provided about a task prior to the agent learning while contextual advice is any information that is state-specific and thus provided when the agent is executing the task. Contextual advice is then further decomposed into the categories of guidance and feedback. Guidance focuses on informing an agent about future actions while feedback focuses on proving information about past actions. Lastly, feedback is broken into corrective and evaluative feedback. While both types of feedback aim to evaluate the quality of an agent's actions, corrective feedback is also concerned with providing information about the potentially correct action(s). Table~\ref{tab:irl_definitions} contains a summary of these definitions and other interactive RL definitions used throughout the paper.

As the paper continues we will attempt to narrow our focus from human input, to feedback, and then finally intrinsic feedback. To provide proper exploration and understanding of interactive RL, and in turn intrinsic feedback, we have formulated three critical questions which we deem necessary to address:
\begin{enumerate}[label=(Q\arabic*)] %, ref={Q}.\roman{*}]
    \item What are the motivations for pursuing interactive RL? \label{item:irl-Q1}
    \item How can human input $h$ be integrated into an RL algorithm such that an agent learns a good policy $\pi$? \label{item:irl-Q2}
    \item How well do interactive RL works fulfill the motivations of \ref{item:irl-Q1}? \label{item:irl-Q3}
\end{enumerate}
These three critical questions aim to structure the reaming sections and will be refined in their scope as the paper progresses. First, the \nameref{sec:irl_motivation} section will attempt to address \ref{item:irl-Q1} by laying out the general motivations of interactive RL and human input. In turn, these motivations will carry over to feedback and instinct feedback. Next, the \nameref{sec:LfF} section will address and refine \ref{item:irl-Q2} and \ref{item:irl-Q3} by discussing the idea of feedback as a channel of communication, exploring how feedback has been used for learning, and discussing the validity of the approaches. This is done to provide readers with a solid foundation for how intrinsic feedback might be properly formulated and integrated. Finally, using the established interactive RL and LfF foundations, the \nameref{sec:iirl} section will attempt to refine and re-explore all critical questions from the perspective of intrinsic feedback.

\begin{table*}
    \footnotesize
    \captionsetup{singlelinecheck = false}
    \caption{Definitions for various interactive RL terms used throughout the paper.}
    \begin{tabular}{|L{1.5cm}L{6cm}|L{1.5cm}L{6cm}|@{}}
        \hline
        \textbf{Term} & \textbf{Definition} & \textbf{Term} & \textbf{Definition} \\
        \hline
        Human input
        & Any process or act that assists an agent in its learning, allowing for a transfer of knowledge from the human to the agent.
        & Demonstrations
        & Human input that shows an agent how to perform a task through execution of the task.\\
        \hline
        Advice
        & Human input which communicates information about the task to an agent without a human actually executing the task.
        & Contextual Advice
        & A form of advice that communicates information that is state specific and thus provided as the agent is executing the task.\\
        \hline
        Feedback
        & A form of contextual advice that communicates information about previous actions. This is typically done by evaluating the "goodness" or "badness" of an agent’s action or behavior.
        & Evaluative Feedback
        & A form of feedback that aims to evaluate the quality of an agent’s action or behavior.\\
        \hline
        Corrective Feedback
        & A form of feedback that aims to evaluate the quality of an agent’s action or behavior while also providing information regarding the potentially correct action(s) or behavior(s).
        & Explicit Feedback
        & Feedback conveyed directly to the agent, typically through a manual means (e.g., pushing a button).\\
        \hline
        Implicit Feedback
        & Feedback that is implied through a human's observable social feedback (e.g., social cues, facial expressions, body language, or vocal tone). 
        & Intrinsic Feedback
        & Feedback that is implied through a human's intrinsically occurring biological signals  (e.g., brain signals, heart rate, or blood pressure). We focus specifically on brain signals in this paper.\\
        \hline
        Feedback Modeling
        & A method where feedback is modeled. Typically, the predicted feedback from this model is then used with a shaping methodology.
        & Reward shaping
        & A shaping methodology for integrating advice (e.g.,feedback) where it is used to augment or replace the agent’s reward function.\\
        \hline
        Value shaping
        & A shaping methodology for integrating advice (e.g.,feedback) where it is used to augment or replace the agent’s value function.
        & Policy shaping
        & A shaping methodology for integrating advice (e.g.,feedback) where it is used to augment or replace the agent’s policy.\\
        \hline
        Quantitative Feedback
        & An interpretation of feedback in which the representation conveyed to the agent is numerical such that a magnitude (i.e., specific value) of how good or bad an agent's action or behavior was is measured. When performing feedback modeling, this is akin to a regression problem.
        & Qualitative Feedback
        & An interpretation of feedback in which the representation conveyed to the agent is categorical such that no specific value regarding how good or bad an agent's action or behavior was is provided. When performing feedback modeling, this is akin to a classification problem. \\
        \hline
    \end{tabular}
    \label{tab:irl_definitions}
    \normalsize
\end{table*}

%%%%%%%%%%%%%%%%%%%%%%%%%%%%%%%%%%%%%%%%%%%%%%%%%%%%%%%%%%%%%%%%%%%%%%%%%%%%%%%%%%%%%%%%%%%%%%%%%%%%%%%%%%%%%%%%%%%%%%%%%%

\subsection{Human Input Motivations}\label{sec:irl_motivation}
To address \ref{item:irl-Q1}, we briefly examine the general motivations for utilizing human input, which will in turn apply as motivations for not only feedback but also intrinsic feedback. The main motivation for integrating human input is to enable human-to-agent knowledge transfer, thereby helping to alleviate issues concerning aptitude and alignment. A core assumption is that human input will often provide a more information-rich signal than that of the typical reward function \citep{knox_thesis_2012}. If utilized properly, this can have various positive effects on aptitude and alignment issues \citep{warnell_deeptamer_2018, arumugam_dlcoach_2019, ouyang_training_2022}. Even if the human input is only on par with the reward signal in its information density, it can still serve as an additional source of information that the reward function may not adequately encompass \citep{knox_combining_2010, knox_reinforcement_2012}. Below, we delve into why these alleviations tend to occur. 

Before we do so, it is important to recall some if the impediments to traditional RL so that we can better understand the impact human input can have. One of the largest impediments to RL is the agent's lack of prior knowledge, requiring the agent to learn from scratch. This challenge is further exacerbated by trial-and-error search which necessitates exploring numerous possibilities, the majority of which are likely to be ineffective or even useless. In addition, reward function poses additional limitations. For instance, the reward function provides only a single channel of communication, constraining humans to convey their knowledge through a signal communication channel and thereby limiting the amount of information transfer. Another limitation stems from a poorly specified reward function, often due to its rule-based formulation. To create a reward function, one must encode their intentions into a set of rules. However, during the encoding process, there is the possibility of losing certain aspects of the human intentions due to the difficulty in translating abstract intentions into a set of rules. As a result, the reward function may become too ambiguous or poorly correlated with the human's ``true'' intentions \citep{amodei_aisafty_2016}. Moreover, the reward function is usually sparse (i.e., very few reward signals are given), leading to the credit assignment problem where the reward is temporally misaligned with its contributing states \citep{sutton_temporal_1984}. These factors make RL inherently sample inefficient as learning can require millions of samples to discover an effective solution, even for environments humans might consider simplistic (e.g., Atari games) \citep{mnih_dqn_2015, mnih_a3c_2016}. This indicates that scaling up to more complex real-world environments could demand an unimaginable amount of samples, assuming learning is even feasible.
 
The primary motivation for integrating human input is to alleviate aptitude issues, which encompass various sub-motivations. One of these sub-motivations is to increase aptness (i.e., the rate at which an agent learns) \citep{knox_interactively_2009, christiano_preferences_2017, zhang_irlsurvey_2019, arzate_irlsurvey_2020}. Human input inherently constrains the agent's exploration space towards actions or behaviors that the human believes to be effective. In turn, this bias towards human-guided exploration significantly reduces the amount of trial-and-error search required by the agent \citep{warnell_deeptamer_2018, knox_interactively_2009,  macglashan_coach_2017, arumugam_dlcoach_2019, christiano_preferences_2017, vinyals_alphastar_2019}. Additionally, human input overcomes the limitation of having only one channel of communication by allowing for a combination of multiple channels (e.g., demonstrations, advice, and the reward function). The integration of human input with a reward function is a clear example of how this can be achieved, as it often leads to increased aptness by providing additional information more easily \citep{knox_combining_2010, knox_reinforcement_2012, arakawa_dqntamer_2018, cruz_multimodal_2016}. Yet another sub-motivation arises when the reward function is unknown as leveraging human input enables the derivation of a new reward function that can be used by traditional RL algorithms \citep{knox_tamer_2008, christiano_preferences_2017, ng_inverse_2000}. Oftentimes, a reward function can be learned using two different sources of human input, such as demonstrations and feedback. This ensures a greater range of information coverage that might be lacking in any one channel \citep{li_interactive_2018, ibarz_rewardlearning_2018}. Lastly, certain types of human input like feedback can present a lesser credit assignment problem compared to that of a delayed sparse reward function. This arises due to humans often being able to provide feedback shortly after the behavior of interest \citep{knox_interactively_2009, macglashan_coach_2017}.
 
A secondary motivation of integrating human input is to alleviate alignment issues, which arise from the challenge of specifying a ``complete'' reward function that fully captures the human's intentions. Alignment issues, such as negative side effects and reward hacking, tend to occur more frequently in complex problems where formulating a reward function can be quite difficult \citep{amodei_aisafty_2016, leike_alignment_2018, hendrycks_what_2021, nahian_training_2021, cruz_interactive_2021}. Negative side effects entail any negative and unintended behaviors that occur when the agent seeks to maximize its reward. For example, a cleaning robot may inadvertently disrupt its environment by knocking over furniture in the pursuit of maximizing its goal of cleaning. Similarly, rewarding hacking involves any exploits or loopholes in a reward function that the agent uses to efficiently maximize its reward. For instance, a cleaning robot may clean the environment and then intentionally dirty the environment so that it can continuously clean and receive a reward. 

Human input inherently offers various benefits that can alleviate alignment issues. One of these benefits is the introduction of communication channels that tend to be more intuitive and natural for humans to express \citep{thomaz_teachable_2008, thomaz_reinforcement_2006}. The intuitive and straightforward nature of human input reduces the chances of losing the intended meaning during the encoding process. This is largely because of the shifted emphasis on an agent's (i.e., algorithm's) ability to interpret and utilize the human input rather than on the human's ability to convert their intentions into a reward function, which may not be feasible for complex problems. For instance, both InstructGPT and ChatGPT use interactive RL and human input to generate responses that are more aligned with human expectations and intentions \citep{ouyang_training_2022}. Another benefit, as mentioned previously, entails the combination of multiple channels of communication. If one channel fails to fully convey a particular intention, another or the combination of channels potentially can \citep{li_interactive_2018, ibarz_rewardlearning_2018}. On top of this, human input possesses different qualities from that of the reward function which can be valuable for learning. For example, human input tends to be non-stationary and policy-dependent, allowing the adaptation to misaligned agent behaviors during online learning \citep{macglashan_coach_2017}.

%%%%%%%%%%%%%%%%%%%%%%%%%%%%%%%%%%%%%%%%%%%%%%%%%%%%%%%%%%%%%%%%%%%%%%%%%%%%%%%%%%%%%%%%%%%%%%%%%%%%%%%%%%%%%%%%%%%%%%%%%%

\section{Learning from Feedback}\label{sec:LfF}
\begin{figure}[t!]
    \centering
    \subfloat[General feedback-based setup]{\includegraphics[scale=.4]{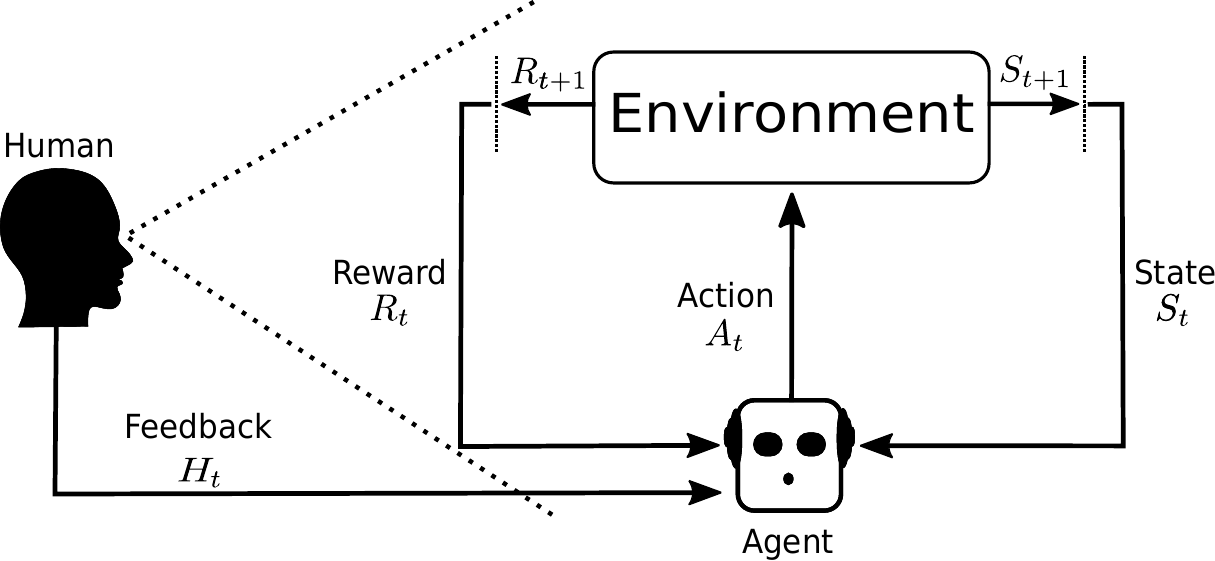}
        \label{fig:fb}}
        \hfil
    \subfloat[Preference-based setup]{\includegraphics[scale=.4]{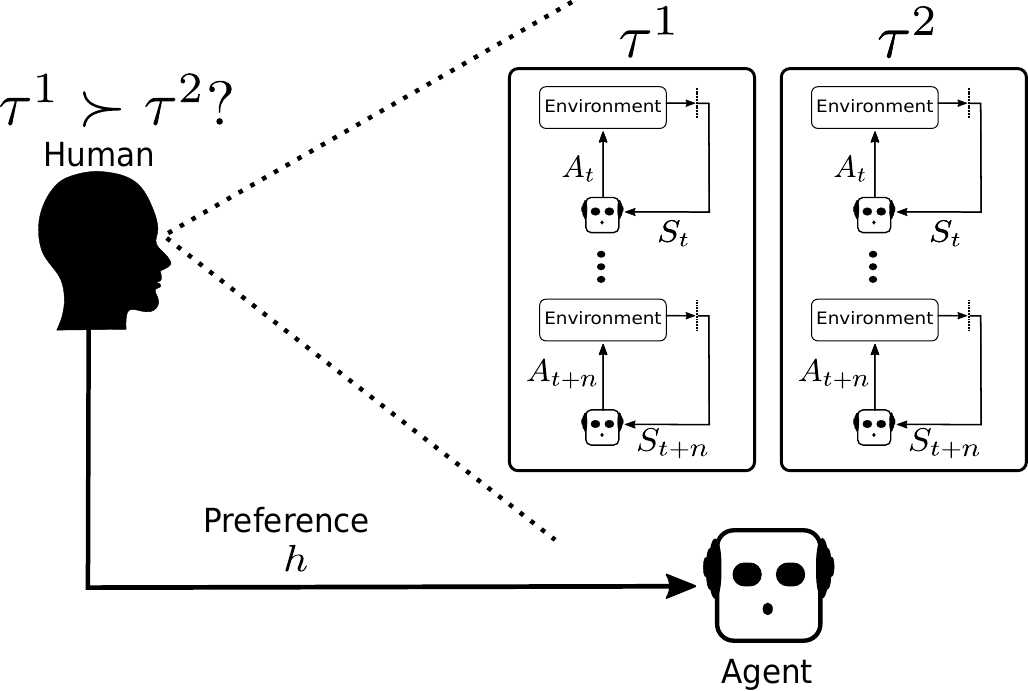}
        \label{fig:pbl}}
        \hfil
    \caption{LfF setups that can use either raw human feedback $h$ or modeled human feedback $\hat{h}$ to influence the agent. (\subref{fig:fb}) The general LfF setup used for either quantitative or qualitative feedback. (\subref{fig:pbl}) The LfF setup for preference-based qualitative feedback where preference is given in regards to pairwise agent trajectories $\tau^1$ and $\tau^2$. To display a short behavior, each trajectory contains a series of state-action tuples up to a given horizon $n$.}
     \label{fig:irl-qual}
\end{figure}

With the general motivations of incorporating human input, and in turn feedback, examined, we can now look at addressing \ref{item:irl-Q2} and \ref{item:irl-Q3}. To do so, we will further narrow the scope of these two questions to focus on feedback. This is crucial to do as intrinsically occurring neural activity can be most intuitively formulated as an alternative feedback medium. Thus, this marks LfF as a second foundational topic for formulating, understanding, and integrating intrinsic feedback. To visualize LfF at a high-level, \figref{fig:fb} shows how the original RL framework can be adapted to account to depict the LfF setup. Here the human can be seen observing the agent and providing feedback denoted as $h$ to the agent. The agent then uses this feedback to assist in its learning. Later on, we will take this general LfF setup and refine it for intrinsic feedback. 

\begin{figure}[!t]
    \centering
    \includegraphics[scale=.4]{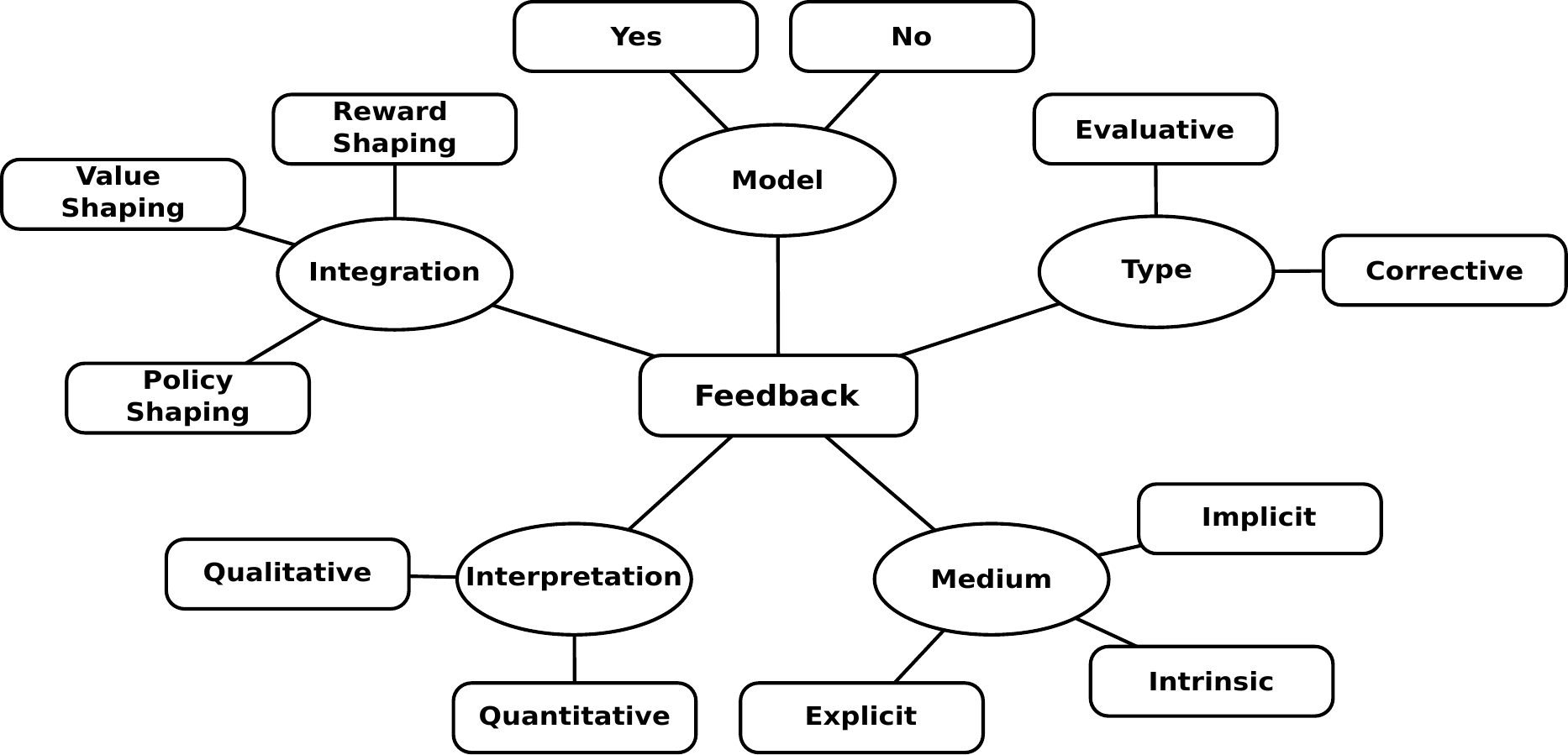}
    \caption{A bubble map summarizing the components to consider when using feedback. When using feedback one must consider what medium will be used to communicate the feedback (i.e., medium), what the agent's internalized representation of the feedback will look like (i.e., interpretation), what the types of feedback will be used (i.e., type), if the feedback will be modeled or not (i.e., model), and ways for integrating of feedback into the RL framework (i.e., integration). All five components interplay and can influence the selection of each other. The circle nodes indicate the feedback components while the rectangle nodes indicate a potential choice for a given component.}
    \label{fig:feedback-components}
\end{figure}

%%%%%%%%%%%%%%%%%%%%%%%%%%%%%%%%%%%%%%%%%%%%%%%%%%%%%%%%%%%%%%%%%%%%%%%%%%%%%%%%%%%%%%%%%%%%%%%%%%%%%%%%%%%%%%%%%%%%%%%%%%
\subsection{Feedback Components}
Before discussing LfF approaches, we can begin to address \ref{item:irl-Q2} by first looking at the five core components require when developing a LfF approach \citep{najar_irlsurvey_2021}. While \figref{fig:feedback-components} provides an overview of these components, we will provide a brief discussion of each component, including its role and how it may influence other components. While we discuss the components in a top-down approach in relation to component influence, this does not necessarily dictate the order in which components must be considered.

%%%%%%%%%%%%%%%%%%%%%%%%%%%%%%%%%%%%%%%%%%%%%%%%%%%%%%%%%%%%%%%%%%%%%%%%%%%%%%%%%%%%%%%%%%%%%%%%%%%%%%%%%%%%%%%%%%%%%%%%%%
\paragraph{Type}
One of the first components to consider is the type of feedback that is going to be used. According to the taxonomy of human input, two types of feedback exist: evaluative and corrective. It is worth noting that both types of feedback aim to assess the quality of an agent's actions, but corrective feedback goes further by providing information about potentially correct actions. The choice of feedback largely dictates the formulation of the LfF approach. More specifically, it influences decisions regarding the communication medium for feedback (i.e., medium), the agent's feedback interpretation process (i.e., interpretation), or the integration of feedback into the RL framework (i.e., integration). Notably, corrective feedback is often combined with demonstrations, wherein the correct action is provided as the feedback \citep{najar_irlsurvey_2021}. However, these types of corrective feedback are not particularly relevant to intrinsic feedback. Therefore, this paper primarily focuses on evaluative feedback, considering that intrinsic feedback can readily be interpreted as such. Nevertheless, some works involving corrective feedback hold promise for being compatible with intrinsic feedback \citep{celemin_interactive_2019}. 

%%%%%%%%%%%%%%%%%%%%%%%%%%%%%%%%%%%%%%%%%%%%%%%%%%%%%%%%%%%%%%%%%%%%%%%%%%%%%%%%%%%%%%%%%%%%%%%%%%%%%%%%%%%%%%%%%%%%%%%%%%
\paragraph{Medium}
The medium component determines what type of feedback medium is going to be used. As mentioned, there are three options: explicit, implicit, and intrinsic. Historically, explicit feedback has been the default choice as it can be conveyed through direct means such as a button press. While explicit feedback is relatively straightforward to work with, implicit feedback, and now intrinsic feedback, can be considered ``indriect'' or ``implied'' mediums of feedback, making them slightly more complex to use. For instance, to detect social (e.g., facial expression) or biological (e.g., neural activity) signals as positive or negative feedback, a classifier is required. Due to the inherent bias in any given classifier, resulting in potential misclassification, the feedback provided is prone to noise. To compensate for these noisy feedback signals, additional mechanisms are often added to an interactive RL algorithm to make it more functional \citep{xu_accelerating_2021, wang_maximizing_2020}. The choice of the medium often impacts the component choice for how the agent will need to interpret the feedback (i.e., interpretation). This is largely due to implicit and intrinsic feedback can be harder to represent using numerical values than categorical values.

%%%%%%%%%%%%%%%%%%%%%%%%%%%%%%%%%%%%%%%%%%%%%%%%%%%%%%%%%%%%%%%%%%%%%%%%%%%%%%%%%%%%%%%%%%%%%%%%%%%%%%%%%%%%%%%%%%%%%%%%%%
\paragraph{Interpretation}
The interpretation component is responsible for defining how the agent will interpret or represent the feedback. This is done by representing feedback either quantitative (i.e., numerically) or qualitatively (i.e., categorically). Quantitative feedback attempts to capture the magnitude of the feedback. Typically, this can be accomplished by representing feedback as binary (i.e., +1 or -1) \citep{knox_interactively_2009, knox_thesis_2012, macglashan_coach_2017} where positive feedback (e.g., +1) suggests the agent's behavior was good and negative feedback (e.g., -1) suggests the agent's behavior was bad. By summing repeatedly elicited feedback using reward aggregation, it is possible to map the feedback to the corresponding state-action pairs, providing a representation of the magnitude of the quality of the behavior \citep{knox_interactively_2009, macglashan_coach_2017}. Alternatively, one can manually adjust the values of positive or negative feedback. However, this can act as a hyper-parameter that is problem dependent and challenging to choose.

When it is difficult to determine the magnitude of feedback that should be given, it is possible to formulate feedback more generally using qualitative feedback. Naively, this can be done by simply using one-hot encodings to represent the categorical variables of good or bad. Alternatively, rather than providing feedback representing the "goodness" or "badness", even more general feedback can be given. For instance, preference-based feedback can be given to indicate whether a behavior is simply preferred over another \citep{wirth_preference_2017}. In practice, preference can be given in relation to a pair of either two different states $S_i^1 \succ S_i^2$, actions  $A_i^1 \succ A_i^2$, or trajectories $\tau^1 \succ \tau^2$ \citep{wirth_preference_2017}. \figref{fig:pbl} shows how the original RL formulation is updated to account for preference-based learning with trajectories.  

The interpretation used often affects the component concerned with how the feedback is modeled (i.e., model). For instance, quantitative feedback modeling is formulated as a regression problem while qualitative feedback is formulated as a classification problem. Thus, it can be helpful to conceptualize quantitative feedback as the quality (good or bad) of a given behavior and qualitative feedback as the likelihood that a behavior is considered good or preferred \cite{knox_tamer_2008, griffith_policy_2013, christiano_preferences_2017}.

%%%%%%%%%%%%%%%%%%%%%%%%%%%%%%%%%%%%%%%%%%%%%%%%%%%%%%%%%%%%%%%%%%%%%%%%%%%%%%%%%%%%%%%%%%%%%%%%%%%%%%%%%%%%%%%%%%%%%%%%%%
\paragraph{Model}
Next, it is important to consider whether the provided feedback will need to be modeled or if the raw feedback will be used. In many cases, continually capturing human feedback can be considered expensive or impractical \citep{christiano_preferences_2017, leike_alignment_2018}. Thus, it can prove beneficial to model feedback, denoted as $\hat{h}$, in order to minimize the reliance on extensive human supervision \citep{knox_interactively_2009, wirth_preference_2017, ouyang_training_2022}. In the simplest cases, modeling quantitative feedback involves solving a regression problem to estimate the magnitude of feedback. Likewise, modeling qualitative feedback includes tackling a classification problem to estimate the probability of a behavior being good or preferred. However, there may be situations where it is desirable to extract a numerical representation from the qualitative feedback by modeling the qualitative feedback and extracting a latent quantitative value \citep{wirth_preference_2017, christiano_preferences_2017, xu_accelerating_2021}.

The manner in which the feedback is modeled is likely to affect the integration component, which determines how the feedback is incorporated into the RL framework. For example, qualitative feedback in which the model outputs probabilities, may not be well-suited for reward shaping without additional considerations \citep{christiano_preferences_2017, xu_accelerating_2021}.

%%%%%%%%%%%%%%%%%%%%%%%%%%%%%%%%%%%%%%%%%%%%%%%%%%%%%%%%%%%%%%%%%%%%%%%%%%%%%%%%%%%%%%%%%%%%%%%%%%%%%%%%%%%%%%%%%%%%%%%%%%%
\paragraph{Integration}\label{sec:shaping_methodologies}
% OLD: The final component relates to how feedback can be integrated into the RL framework and used to influence the agent's learning. First, it is crucial to recall that the MDP specification of an RL framework is made up of three primary components: the reward function, value function, and policy. Thus, it is possible to interpret feedback as \textit{shaping} any one of these components. The process of shaping entails reinforcing the agent's sub-behaviors to facilitate the agent's learning of a more complex desired behavior\citep{knox_interactively_2009, gullapalli_shaping_1992}. According to this definition, shaping in traditional RL is typically done using the reward function such that reward function acts as external reinforcement that shapes the agent by influencing either the value function, policy, or both. Likewise, we can generalize this idea of shaping for feedback (and other forms of advice). This allows for the formulation of three categories of shaping methodologies: reward shaping, value shaping, and policy shaping. While we specifically use the raw human feedback $h$ and action-value function $Q$, the below equations can be adjusted to account for modeled human feedback $\hat{h}$ or the state-value function $V$.
The final component relates to how feedback can be integrated into the RL framework. Recall that the MDP specification of an RL framework is made up of three primary components: the reward function, value function, and policy. Thus, it is possible to interpret feedback as \textit{shaping} (i.e., augmenting) any one of these components\footnote{Here ``shaping'' is synonymous with ``augmenting'', as indicated by \citet{ng_reward_1999}. However, the term ``shaping'' has also been defined in the interactive RL literature by \citet{knox_interactively_2009} to refer to the reinforcement of the agent's sub-behaviors to facilitate the agent's learning of a more complex desired behavior. Due to this overloaded usage, we will differ to the former definition as it fits the terms reward, value, and policy shaping better.}. This allows for the formulation of three categories of shaping methodologies: reward shaping, value shaping, and policy shaping. While we specifically use the raw human feedback $h$ and action-value function $Q$, the below equations can be adjusted to account for modeled human feedback $\hat{h}$ or the state-value function $V$.

% NEW but removed: Following \citet{knox_interactively_2009}, we define \textit{shaping} as the reinforcement of the agent's sub-behaviors to facilitate the agent's learning of a more complex desired behavior. Thus, shaping in traditional RL is done via the reward function. Recall that the reward function is usually an external signal provided by the environment. This reward signal then influences the agents behaviors by being taken into account when computing the value function or policy. Likewise, we can generalize this idea of shaping for feedback\footnote{This generalization can apply to other forms of advice as well.}. 

Reward shaping is an shaping methodology where the reward function $r(s, a)$ is augmented using feedback and the resulting reward $r^\prime$ is then utilized to influence the agent \citep{ng_reward_1999}. Augmentation of $r(s, a)$ is commonly done by simply perturbing the current reward $r$ by $h$ such that
\begin{equation}\label{eq:reward_shaping}
    r^\prime = r + \beta h,
\end{equation}
where $\beta$ is predefined parameter that scales the impact of the feedback. While quantitative feedback can be directly utilized with reward shaping \citep{kuri_dynamic_2010, knox_combining_2010, knox_reinforcement_2012}, qualitative feedback requires modeling by learning of a latent numerical function $\hat{h}$ \citep{christiano_preferences_2017, xu_accelerating_2021}. Moreover, when the reward function is unknown, reward shaping can entail completely replacing $r^\prime$ with $h$ \citep{thomaz_reinforcement_2006, thomaz_posneg_2007, thomaz_teachable_2008, knox_learning_2013, knox_myopic_2012, knox_framing_2015, christiano_preferences_2017, luo_drlpref_2018}.

Similarly, value shaping entails the augmentation of the agent's value function. For example, Q-augmentation \citep{knox_combining_2010, knox_reinforcement_2012} uses $h$ to perturb the value function $Q$ such that
\begin{equation}\label{eq:value_shaping}
    Q^\prime(s, a) = Q(s, a) + \beta h(s, a),
\end{equation}
where $\beta$ is predefined parameter that scales the impact of the feedback. Once again, while quantitative feedback can be directly utilized by value shaping \citep{knox_combining_2010, knox_reinforcement_2012}, qualitative feedback typically requires some sort of modeling. When the value function is unknown, value shaping can entail simply replacing $Q^\prime$ with $h$. 

Policy shaping entails the augmentation of the agent's policy \citep{griffith_policy_2013, macglashan_coach_2017}. In contrast to other shaping approaches, policy shaping offers a broader range of implementations \citep{knox_combining_2010, knox_reinforcement_2012, griffith_policy_2013, macglashan_coach_2017}. Two common approaches are referred to as control sharing and action biasing \citep{knox_combining_2010, knox_reinforcement_2012}. Control sharing is defined as 
\begin{equation}\label{eq:control_sharing}
    P(a\!= \pi_{h} )=\mathrm{min}(\beta, 1),
\end{equation}
where the agent takes an action according to the human derived policy $\pi_{h}$ given some probability $\beta$. On the other hand, action biasing is defined as 
\begin{equation}\label{eq:action_biasing}
    \pi(s) = \argmax{a}[Q(s, a) + \beta h(s, a) ],
\end{equation}
where the agent's value function $Q$ is perturbed by $h$ at the time of action selection. As before, if the agent policy is unknown, replacing $\pi$ with $\pi_{h}$ can be considered as policy shaping.

\begin{sidewaystable}
    \scriptsize
    \caption{Overview of each LfF works covered. Information regarding the authors, the five feedback components, RL algorithm(s) that are shaped, environments, baselines, and alleviation/degradation in aptitude and alignment issues when compared to the baselines. For the five feedback components, all works are formulated using the evaluative feedback, thus no feedback type is provided. For the environments, the \xmark \hspace{1pt} indicates environments that were used outside of the original work. For the baselines, the asterisks $^*$ denotes non-interactive RL baseline. For aptitude and alignment issues, a \uptri{.2} indicates alleviation while a \downtri{.2} indicates degradation. The size of the triangle gives a rough approximation to the degree of alleviation or degradation. Quan=quantitative, Qual=qualitative, RS=reward shaping, VS=value shaping, PS=policy shaping, QL=Q-learning, DQL=deep Q-learning.}
    \begin{tblr}{
    width=1\textwidth,
    colspec = {|X[0.1,c,m]|
            X[1.5,l,m]|
            X[0.25,c,m]X[0.1,c,m]|
            X[0.3,c,m]|
            X[0.1,c,m]X[0.1,c,m]X[0.1,c,m]|
            X[0.8,c,m]|
            X[0.1,c,m]X[0.1,c,m]X[0.1,c,m]X[0.1,c,m]X[0.1,c,m]X[0.1,c,m]X[0.1,c,m]X[0.1,c,m]|
            X[0.9,c,m]|
            X[0.1,c,m]X[0.1,c,m]X[0.1,c,m]X[0.1,c,m]|},
    row{1-2} = {c},
    cell{1}{1} = {r=2}{},
    cell{1}{2} = {r=2}{},
    cell{1}{3} = {c=2}{},
    cell{1}{5} = {r=2}{},
    cell{1}{6} = {c=3}{},
    cell{1}{9} = {r=2}{},
    cell{1}{10} = {c=8}{},
    cell{1}{18} = {r=2}{},
    cell{1}{19} = {c=4}{},
    cell{3}{1} = {r=8}{},
    cell{11}{1} = {r=7}{},
    vline{4,7-8,11-17,20-22} = {2}{},
    hline{1,3} = {-}{},
    hline{2} = {3-5,6-9,10-18,19-23}{},
    }

        \vt{Medium} 
        & Authors \& Year 
        & Interpretation &
        & Model
        & Integration & &
        & Shaped Algorithms
        & Environments & & & & & & &
        & Baselines
        & RL Challenges & & &\\
        &
        & {\scriptsize \vt{Quan}}
        & {\scriptsize \vt{Qual}}
        & 
        & {\scriptsize \vt{RS}}
        & {\scriptsize \vt{VS}}
        & {\scriptsize \vt{PS}}
        &   
        & {\tiny \vt{Mt Car}}
        & {\tiny \vt{Cart Pole}}
        & {\tiny \vt{GridWorld}}
        & {\tiny \vt{Tetris}} 
        & {\tiny \vt{Atari}}
        & {\tiny \vt{MuJoCo}} 
        & {\tiny \vt{Minecraft}} 
        & {\tiny \vt{Robotics}}
        & 
        & {\tiny \vt{Robustness}}
        & {\tiny \vt{Scalability}} 
        & {\tiny \vt{Aptness}} 
        & {\tiny \vt{Alignment}} \\ 
        
        \vt{Explicit} 
        & \citet{knox_interactively_2009}( \citeyear{knox_interactively_2009}) %TAMER 
        & \cmark 
        & \nomark
        & \cmark 
        & \cmark 
        & \nomark
        & \nomark 
        & Myopic RL
        & \cmark 
        & \xmark 
        & \xmark 
        & \cmark 
        & \xmark 
        & \nomark
        & \nomark 
        & \xmark  
        & {SARSA($\lambda$)$^*$}
        & \downtri{0.10}
        & \uptri{0.15}
        & \uptri{0.15}
        & \nomark  \\ 
        \cline{2-22}
        
        & \citet{knox_combining_2010, knox_reinforcement_2012}( \citeyear{knox_reinforcement_2012}) %TAMER+RL 
        & {\cmark} 
        & {\nomark} 
        & {\cmark} 
        & {\cmark} 
        & {\cmark}
        & {\cmark} 
        & SARSA($\lambda$) %  Myopic RL + SARSA($\lambda$)
        & {\cmark} 
        & {\cmark} 
        & {\nomark} 
        & {\nomark} 
        & {\nomark} 
        & {\nomark} 
        & {\nomark} 
        & {\nomark}  
        & \citep{knox_interactively_2009}, SARSA($\lambda$)$^*$
        & {\uptri{0.1}} 
        & {\uptri{0.1}}  
        & {\nomark}
        & {\nomark}  \\ 
        \cline{2-22}
        
        & \citet{knox_framing_2015, knox_myopic_2012, knox_learning_2013} ( \citeyear{knox_framing_2015}) %VI-TAMER
        & {\cmark} 
        & {\nomark} 
        & {\cmark} 
        & {\cmark} 
        & {\nomark}
        & {\nomark} 
        & SARSA($\lambda$) % Non-myopic + SARSA($\lambda$)
        & {\cmark} 
        & {\nomark} 
        & {\cmark} 
        & {\nomark} 
        & {\nomark} 
        & {\nomark} 
        & {\nomark} 
        & {\nomark}  
        & \citep{knox_combining_2010}
        & {\uptri{0.15}} 
        & {\nomark}  
        & {\nomark}
        & {\downtri{0.1}}  \\ 
        \cline{2-22}
        % Deep TAMER
        & \citet{warnell_deeptamer_2018}( \citeyear{warnell_deeptamer_2018}) %deep tamer  
        & {\cmark} 
        & {\nomark} 
        & {\cmark} 
        & {\cmark} 
        & {\nomark}
        & {\nomark} 
        & Myopic DQL 
        & {\nomark} 
        & {\nomark} 
        & {\xmark} 
        & {\nomark} 
        & {\cmark} 
        & {\nomark} 
        & {\xmark} 
        & {\nomark}  
        & 
       \citep{knox_interactively_2009}, \citep{mnih_a3c_2016}$^*$, \citep{hasselt_ddqn_2016}$^*$
        & {\uptri{0.2}} 
        & {\nomark}  
        & {\uptri{0.2}}
        & {\nomark}  \\ 
        \cline{2-22}
        
        % DQNTAMER
        & \citet{arakawa_dqntamer_2018}( \citeyear{arakawa_dqntamer_2018}) % DQN TAMER
        & {\cmark} 
        & {\nomark} 
        & {\cmark} 
        & {\nomark} 
        & {\nomark}
        & {\cmark} 
        & DQL % Myopic DQL + DQL
        & {\nomark} 
        & {\nomark} 
        & {\cmark} 
        & {\nomark} 
        & {\nomark} 
        & {\nomark} 
        & {\nomark} 
        & {\nomark}  
        & \citep{warnell_deeptamer_2018}, \citep{mnih_dqn_2015}
        & {\uptri{0.1}} 
        & {\uptri{0.1}}  
        & {\uptri{0.1}} 
        & {\nomark}  \\ 
        \cline{2-22}
        
        & \citet{macglashan_coach_2017}( \citeyear{macglashan_coach_2017}) %COACH 
        & {\cmark} 
        & {\nomark} 
        & {\nomark} 
        & {\nomark} 
        & {\nomark}
        & {\cmark} 
        & Actor-Critic
        & {\nomark} 
        & {\nomark} 
        & {\nomark} 
        & {\nomark} 
        & {\nomark} 
        & {\nomark} 
        & {\xmark} 
        & {\cmark}  
        &
       \citep{knox_interactively_2009}
        & {\uptri{0.15}} 
        & {\nomark}  
        & {\uptri{0.15}}
        & {\nomark}  \\ 
        \cline{2-22}
        
        & \citet{arumugam_dlcoach_2019}( \citeyear{arumugam_dlcoach_2019}) %Deep COACH
        & {\cmark} 
        & {\nomark} 
        & {\nomark} 
        & {\nomark} 
        & {\nomark}
        & {\cmark} 
        & Actor-Critic
        & {\nomark} 
        & {\nomark} 
        & {\nomark} 
        & {\nomark} 
        & {\nomark} 
        & {\nomark} 
        & {\cmark} 
        & {\nomark}  
        &
       \citep{warnell_deeptamer_2018}, \citep{macglashan_coach_2017}
        & {\uptri{.15}} 
        & {\uptri{.10}}  
        & {\uptri{.10}}
        & {\nomark}  \\ 
        \cline{2-22}
        
        & \citet{christiano_preferences_2017}( \citeyear{christiano_preferences_2017}) % Learning from prefences
        & {\nomark} 
        & {\cmark} 
        & {\cmark} 
        & {\cmark} 
        & {\nomark}
        & {\nomark} 
        & A2C, TRPO
        & {\nomark} 
        & {\nomark} 
        & {\nomark} 
        & {\nomark} 
        & {\cmark} 
        & {\cmark} 
        & {\nomark} 
        & {\nomark}  
        &
       \citep{mnih_a3c_2016}$^*$, \citep{schulman_trust_2017}$^*$
        & {\downtri{.1}} 
        & {\uptri{0.2}}  
        & {\downtri{0.1}}
        & {\nomark} \\ 
        \hline

        % Proof of concept
        \vt{Intrinsic} 
        & \citet{iturrate_errprl_2015, iturrate_robot_2010} ( \citeyear{iturrate_errprl_2015})
        & {\cmark} 
        & {\nomark} 
        & {\nomark} 
        & {\cmark} 
        & {\nomark}
        & {\nomark} 
        & QL
        & {\nomark} 
        & {\nomark} 
        & {\nomark} 
        & {\nomark} 
        & {\nomark} 
        & {\nomark}
        & {\nomark} 
        & {\cmark}  
        & --
        & {\nomark} 
        & {\nomark}  
        & {\nomark}
        & {\nomark}  \\ 
        \cline{2-22}

        % Proof of concept
        & \citet{kim_intrinsic_2017}( \citeyear{kim_intrinsic_2017})
        & {\cmark} 
        & {\nomark} 
        & {\nomark} 
        & {\cmark} 
        & {\nomark}
        & {\nomark} 
        & LibUCB
        & {\nomark} 
        & {\nomark} 
        & {\nomark} 
        & {\nomark} 
        & {\nomark} 
        & {\nomark} 
        & {\nomark} 
        & {\cmark}  
        & --
        & {\nomark} 
        & {\nomark}  
        & {\nomark}
        & {\nomark}  \\ 
        \cline{2-22}
        % Proof of concept
        & \citet{ehrlich_errpcoadapt_2018}( \citeyear{ehrlich_errpcoadapt_2018})
        & {\cmark} 
        & {\nomark} 
        & {\nomark} 
        & {\cmark} 
        & {\nomark}
        & {\nomark} 
        & Policy Gradient
        & {\nomark} 
        & {\nomark} 
        & {\nomark} 
        & {\nomark} 
        & {\nomark} 
        & {\nomark} 
        & {\nomark} 
        & {\cmark}  
        & --
        & {\nomark}
        & {\nomark}  
        & {\nomark}
        & {\nomark}  \\ 
        \cline{2-22}
        % Adaptive
        & \citet{luo_drlpref_2018}( \citeyear{luo_drlpref_2018})
        & {\nomark} 
        & {\cmark} 
        & {\cmark} 
        & {\cmark} 
        & {\nomark}
        & {\nomark} 
        & TRPO 
        & {\nomark} 
        & {\nomark} 
        & {\nomark} 
        & {\nomark} 
        & {\cmark} 
        & {\nomark} 
        & {\nomark} 
        & {\nomark}  
        & \citep{christiano_preferences_2017}
        & {\downtri{.1}} 
        & {\uptri{.1}}  
        & {\downtri{.1}}
        & {\nomark} \\ [3pt] 
        \cline{2-22}
        
        % Explorative
        & \citet{akinola_accelerated_2020}( \citeyear{akinola_accelerated_2020})
        & {\nomark} 
        & {\cmark}
        & {\cmark} 
        & {\nomark} 
        & {\nomark}
        & {\cmark} 
        & PPO
        & {\nomark} 
        & {\nomark} 
        & {\nomark} 
        & {\nomark} 
        & {\nomark} 
        & {\nomark} 
        & {\nomark} 
        & {\cmark}  
        & \citep{schulman_ppo_2017}$^*$
        & {\uptri{.1}} 
        & {\uptri{.1}}  
        & {\uptri{.1}}
        & {\nomark} \\ 
        \cline{2-22}
        
        % Explorative
        & \citet{wang_maximizing_2020}( \citeyear{wang_maximizing_2020})
        & {\nomark} 
        & {\cmark}
        & {\cmark} 
        & {\nomark} 
        & {\nomark}
        & {\cmark} 
        & PPO
        & {\nomark} 
        & {\nomark} 
        & {\nomark} 
        & {\nomark} 
        & {\nomark} 
        & {\nomark} 
        & {\nomark} 
        & {\cmark}  
        & \citep{schulman_ppo_2017}$^*$
        & {\uptri{.15}} 
        & {\uptri{.1}}  
        & {\uptri{.15}}
        & {\nomark} \\ 

        % Explorative
        & \citet{xu_accelerating_2021}( \citeyear{xu_accelerating_2021})
        & {\nomark} 
        & {\cmark}
        & {\cmark} 
        & {\cmark} 
        & {\nomark}
        & {\nomark} 
        & Bayesian DQL % Soft DQL + Bayesian DQL
        & {\nomark} 
        & {\nomark} 
        & {\cmark} 
        & {\nomark} 
        & {\cmark} 
        & {\nomark} 
        & {\nomark} 
        & {\nomark}  
        & \citep{xiao_fresh_2020}, \citep{kamyar_bdqn_2018}$^*$
        & {\nomark}
        & {\uptri{.1}}  
        & {\uptri{0.15}}
        & {\nomark}  \\ 
        \cline{1-22}
    \end{tblr}
    \label{tab:irl_works}
    \normalsize
\end{sidewaystable}

%%%%%%%%%%%%%%%%%%%%%%%%%%%%%%%%%%%%%%%%%%%%%%%%%%%%%%%%%%%%%%%%%%%%%%%%%%%%%%%%%%%%%%%%%%%%%%%%%%%%%%%%%%%%%%%%%%%%%%%%%%%%%%%%%%%%
\subsection{Foundational Approaches}\label{sec:foundations-LfF}

To address \ref{item:irl-Q2} in more detail and provide examples of various approaches, we now look at foundational LfF approaches. Specifically, we will look at three foundational frameworks that are commonly referenced and utilized by other interactive RL works: TAMER, COACH, and learning from preferences (LfP). These frameworks therefore provide solid baselines for researchers interested in intrinsic feedback to utilize, build-upon, and draw inspiration from. The TAMER framework explores modeling quantitative feedback, shaping myopic and non-myopic RL agents, and addressing the human feedback credit assignment challenge \citep{knox_combining_2010, knox_framing_2015, warnell_deeptamer_2018, arakawa_dqntamer_2018}. The COACH framework provides a competitive alternative to that of the TAMER framework as it explores policy shaping without any feedback modeling \citep{macglashan_coach_2017, arumugam_dlcoach_2019}. Finally, the LfP framework explores a different, yet popular, approach which focuses on qualitative preference-based feedback \citep{christiano_preferences_2017, wirth_preference_2017, ouyang_training_2022}. 

Additionally, in order to address \ref{item:irl-Q3}, we provide a qualitative summary cornering the improvement or degradation in an algorithm's aptitude and alignment capabilities. In particular, a focus is placed on aptitude due to a lack of alignment analysis found in many existing works. This summary is formed by examining the comparisons made between a given work's proposed algorithm and the provided baseline algorithms. Table~\ref{tab:irl_works} summarizes the sampled LfF works along with the addition of works on intrinsic feedback, which will be discussed in the \nameref{sec:iirl} section.

%%%%%%%%%%%%%%%%%%%%%%%%%%%%%%%%%%%%%%%%%%%%%%%%%%%%%%%%%%%%%%%%%%%%%%%%%%%%%%%%%%%%%%%%%%%%%%%%%%%%%%%%%%%%%%%%%%%%%%%%%%%%%%%%%%%%%%%

\subsubsection{TAMER}\label{sec:TAMER}
Training an Agent Manually via Evaluative Reinforcement (TAMER) framework primarily focuses on modeling quantitative feedback and using human feedback to exert complete control over the RL agent. As we'll see, this approach can be interpreted as combining feedback modeling and a myopic RL agent into a single algorithm \citep{knox_tamer_2008, knox_interactively_2009, knox_thesis_2012}. TAMER goes a step further as it aims to address the credit assignment challenge by mapping delayed feedback to its corresponding behavior. The TAMER framework consists of three main modules: feedback modeling, action selection, and credit assignment.

As TAMER interprets feedback quantitatively such that $h: S \times A  \rightarrow \{-1, 0, 1\}$, modeling is formulated as a supervised learning regression problem. This formalizes TAMER's modeling error $\delta_t$ as
\begin{equation}\label{eq:TAMER_error}
    \delta_t = H_t - \hat{h}(S_t, A_t),
\end{equation}
where $t$ is the current time step, $H_t$ is the raw feedback, and $\hat{h}(S_t, A_t)$ is the predicted feedback. Learning of the parameters $w$ for $\hat{h}$ can be done by simply minimizing the squared error of \eqref{eq:TAMER_error} using stochastic gradient decent as follows:
\begin{equation}\label{eq:TAMER_update}
\begin{split}
    w_{t+1} & = w_t - \alpha \nabla \bigg [ \frac{1}{2} \big(H_t - \hat{h}_{w_t}(S_t, A_t) \big )^2 \bigg ] \\
    & = w_t - \alpha \delta_t  \nabla \hat{h}_{w_t}(S_t, A_t). \\
\end{split}
\end{equation}
Given the TAMER's goal of solely using feedback to control the agent, the action selection process involves greedily selecting the action that is estimated to receive the largest amount of feedback, 
\begin{equation}\label{eq:TAMER_action_selection}
    \pi(s) = \argmax{a} \hat{h}(s,a).
\end{equation}
Thus, one way to interpret TAMER's supervised learning formulation is as a form of reward shaping with myopic RL, where the agent and the modeling steps are combined using the TD error \eqref{eq:temporal_difference_error} \citep{knox_thesis_2012, najar_irlsurvey_2021, zhang_irlsurvey_2019, arzate_irlsurvey_2020}. \eqref{eq:TAMER_error} can then be derived by replacing $r$ with $h$, setting $\gamma = 0$, and substituting $Q$ with $\hat{h}$.

TAMER further addresses the temporal credit assignment problem of how to map feedback to the contributing state-action pairs \citep{knox_interactively_2009, knox_thesis_2012}. As feedback is not instantaneous, the delay between the occurrence of the targeted state-action pair(s) and the elicitation of $h$ needs to be accounted for. Depending on the rate at which state-action pairs are occur, many state-action pairs might appear between the delay of the feedback and the targeted state-action pair(s). Consequently, it can be challenging to precisely determine the exact state-action pair(s) that the feedback was intended for, necessitating an approximation of which state-action pair(s) might have contributed to $h$.
 
The proposed solution TAMER presents is to estimate the delay in feedback using a probability density function (PDF) \citep{knox_interactively_2009, knox_thesis_2012} to compute the probability that a $h$ can be ``credited'' to a given state-action pair. When an accurate PDF is used, it is possible to calculate the probability that a given state-action pair fell within the expected human delay period as
\begin{equation}\label{credit_prob}
c(\mathrm{T}^s, \mathrm{T}^e, \mathrm{T}^h) = \int^{\mathrm{T}^s - \mathrm{T}^h}_{\mathrm{T}^e - \mathrm{T}^h} f_{\text{delay}}(x)dx,
\end{equation}
where $c(\mathrm{T}^s, \mathrm{T}^e, \mathrm{T}^h)$ represents the credit or probability that $h$ belongs to a state-action pair. The PDF $f_{delay}(x)$ is bounded by when the state-action pair began and ended with respect to when $h$ was received. The first bound ${\mathrm{T}^s - \mathrm{T}^h}$ represents the difference between the time at which the state began $\mathrm{T}^s$ and the time at which $h$ was received $\mathrm{T}^h$. The second bound ${\mathrm{T}^e - \mathrm{T}^h}$ represents the difference between the time at which the state ended $\mathrm{T}^e$ and the time at which $h$ was received $\mathrm{T}^h$. Although the selection of the optimal PDF remains unclear and largely unexplored \citep{knox_thesis_2012}, the a uniform distribution bounded by the average human reaction time \citep{hockley_rt_1984} is often adopted \citep{knox_interactively_2009, knox_thesis_2012}.

Credit can be integrated into TAMER's error (\eqref{eq:TAMER_error}) through either of the methods denoted as delay-weighted, individual reward \citep{knox_interactively_2009} or delay-weighted, aggregate reward \citep{knox_thesis_2012}. Delay-weighted, individual reward applies credit to $\hat{h}(s, a)$ such that a weighted sum over $\hat{h}(s, a)$ is computed for each label $H_t$ as 
\begin{equation}\label{eq:credit_state_error}
    \begin{split}
        \delta_t & = H_t - \sum_{k} c(\mathrm{T}^s_{t-k}, \mathrm{T}^e_{t-k}, \mathrm{T}^h_{t}) \hat{h}(S_{t-k}, A_{t-k}),
    \end{split}
\end{equation}
where term $k$ allows for weighted summation using prior state-action pairs contained within a sliding window (i.e., a history of state-action pairs). Thus, only prior state-action pairs $(S_{t-k}, A_{t-k})$ are considered to contribute to the weighted sum as pairs that occurred after $t$ cannot be attributed to $H_t$. State-action pairs are pruned from the window when they are considered be old by having a near-zero credit for all feedback \citep{knox_thesis_2012, knox_interactively_2009}.

Delay-weighted, aggregate reward applies the credit directly to $h$ such that a weighted sum over $h$ is computed for each state-action pair. This weighted sum inherently incorporates all prior feedback that corresponds to the current state-action pair into a single label such that the TAMER error is computed as
\begin{equation}\label{eq:credit_dwar_error}
    \begin{split}
        \delta_t & = \bigg ( \sum_{k} c(\mathrm{T}^s_{t}, \mathrm{T}^e_{t}, \mathrm{T}^h_{t+k}) \, H_{t+k} \bigg ) - \hat{h}(S_{t}, A_{t})
    \end{split}
\end{equation}
where term $k$ allows for the weighted summation over future elicited feedback contained within the sliding window. Thus, for a given state-action pair $(S_t, A_t)$, only future feedback $H_{t+k}$ is considered to contribute to the weighted sum as feedback that occurred before $t$ cannot be attributed to $(S_t, A_t)$. Feedback is pruned from the window when it is considered old by having a near-zero credit for all state-action pairs \citep{knox_thesis_2012}.

TAMER's initial empirical results are conducted in simplistic 2D \citep{knox_interactively_2009, knox_thesis_2012} and robotics \citep{knox_robot_2013} environments. These results show an increase in aptitude performance when compared to the baseline SARSA($\lambda$). This is due to an increase in aptness from having good short-term performance and a likely increase in scalability as limited feedback is required for learning to occur. That being said, TAMER's overall robustness tends to be worse as SARSA($\lambda$) has better asymptotic performance. Additional studies have shown TAMER's overall increase in aptitude performance is quite limited as TAMER fails to learn as the complexity of environments increases \citep{macglashan_coach_2017, warnell_deeptamer_2018}. Specifically, TAMER tends to forget and fails to generalize due to receiving varying feedback in similar states \citep{macglashan_coach_2017, knox_framing_2015}. TAMER's inability to properly handle variation in feedback is associated with its myopic qualities (i.e., supervised learning) which only allows for a rigid encoding of the current policy \citep{macglashan_coach_2017, knox_framing_2015}. TAMER is also limited by its linear modeling of $\hat{h}$ as it is unable to properly handle high dimensional state spaces and non-linear relationships found within more complex environments (e.g., Atari) \citep{warnell_deeptamer_2018}. To address the innate drawbacks, newer iterations of the TAMER framework have been proposed to further increase the framework's aptitude. We discuss some of the more interesting and insightful approaches below. 

%%%%%%%%%%%%%%%%%%%%%%%%%%%%%%%%%%%%%%%%%%%%%%%%%%%%%%%%%%%%%%%%%%%%%%%%%%%%%%%%%%%%%%%%%%%%%%%%%%%%%%%%%%%%%%%%%%%%%%%%%%%%%%%%%%%%%%%

\paragraph{TAMER+RL}\label{sec:TAMER+RL}
TAMER+RL aims to exploit TAMER's short-term performance and RL's asymptotic performance \citep{knox_combining_2010, knox_reinforcement_2012, knox_thesis_2012} by taking advantage of both feedback and the reward function. To combine the benefits of TAMER and RL, the modeling and RL agent are fully separated. Thus, $h$ is modeled using TAMER's supervised learning error function (\eqref{eq:TAMER_error}) and different shaping methodologies are examined to identify the optimal shaping approach when using $\hat{h}$. Initially proposed were eight different shaping approaches \citep{knox_combining_2010} which were later narrowed down to the following four approaches in \citep{knox_reinforcement_2012}: reward shaping (\eqref{eq:reward_shaping}), Q-augmentation (\eqref{eq:value_shaping}), action biasing (\eqref{eq:action_biasing}), and control sharing (\eqref{eq:control_sharing}).

Using the mountain car and cart pole environments, all shaping methodologies show an increase in both aptness and robustness over SARSA($\lambda$). Yet, only some shaping methodologies maintain TAMER's original performance gains in aptness \citep{knox_reinforcement_2012}. In particular, shaping methodologies that gently push the RL agent towards the behaviors of TAMER (e.g., action biasing and control sharing) show increases to that of SARSA($\lambda$) while approximately maintaining TAMER's original performance \citep{knox_combining_2010, knox_reinforcement_2012}. While not shown directly, TAMER+RL shows a potential increase in scalability as it can continue learning without feedback as the reward function can still be utilized. However, there are a few caveats to consider. First, only classical benchmark environments such as cart-pole and mountain car are used for testing. This makes it hard to tell if these results will generalize to more complex environments. Second, while the results can show an increase in aptitude over both TAMER and SARSA($\lambda$), the results can be dependant on the human providing the feedback, the environment, and the shaping methodology chosen. Finally, while using the reward function can be seen as an advantage, it can be also be seen as a constraint if the reward function is unknown.

%%%%%%%%%%%%%%%%%%%%%%%%%%%%%%%%%%%%%%%%%%%%%%%%%%%%%%%%%%%%%%%%%%%%%%%%%%%%%%%%%%%%%%%%%%%%%%%%%%%%%%%%%%%%%%%%%%%%%%%%%%%%%%%%%%%%%%%

\paragraph{VI-TAMER}
VI-TAMER builds upon TAMER+RL by exploring effects of non-myopyic learning when combining TAMER and RL \citep{knox_learning_2013, knox_myopic_2012, knox_framing_2015, knox_thesis_2012}. This work offers valuable insights into potential aptitude-alignment trade-offs that are often overlooked in most LfF works. As TAMER's supervised learning modeling can be formulated as myopic RL problem, this can leave TAMER learning a rigid policy. In turn, this makes it hard for TAMER to generalize to unseen states. The authors postulate that non-myopic learning (e.g., combining TAMER with RL) improves aptitude due to better generalization to unseen states. As seen in TAMER+RL, VI-TAMER fully separates the modeling of feedback from the RL agent, but incorporates reward shaping with replacement as the shaping methodology.

Testing is in the grid world and mountain car environments where various discount values $\gamma$ are examined as well. Results show that non-myopic learning can enhance TAMER's robustness by allowing for better generalization, specifically in the grid world environment when the layout is changed. At the same time, non-myopic learning also exhibited a decrease in overall agent alignment as issues like reward hacking start to appear. Since human feedback tends to be positively biased \citep{thomaz_posneg_2007, thomaz_teachable_2008}, agents are able to exploit this positive feedback and engage in behaviors that do not align with the human's intentions \citep{macglashan_coach_2017, knox_learning_2013, knox_framing_2015, knox_thesis_2012}. While these unintended behaviors can often be corrected during interactive training with a human, novel unintended behaviors are less likely to be addressed when the human stops providing feedback and the model $\hat{h}$ is no longer updated. Interestingly, VI-TAMER overcomes the alignment trade-offs by formulating the tasks as a continual task instead of an episodic task. By switching to continual tasks, the intended behavior (i.e., reaching the goal) is no longer constrained by achieving it as the future reward is not limited to that specific achievement. In contrast, episodic tasks consider the achievement of the intended behavior as the end of the task, thereby limiting the future reward. This work, despite sharing the similar caveats as TAMER+RL, offers interesting opportunities for further trade-off investigations, such as exploring whether similar results can be replicated using increasingly more complex environments or non-linear feedback models.

 %%%%%%%%%%%%%%%%%%%%%%%%%%%%%%%%%%%%%%%%%%%%%%%%%%%%%%%%%%%%%%%%%%%%%%%%%%%%%%%%%%%%%%%%%%%%%%%%%%%%%%%%%%%%%%%%%%%%%%%%%%%%%%%%%%%%%%
 
\paragraph{Deep TAMER}\label{sec:Deep_TAMER}
To overcome the limitations of the TAMER's linear modeling of feedback, \citet{warnell_deeptamer_2018} integrate TAMER with Deep Q-Networks (DQN) \citep{mnih_dqn_2015}. First, a deep convolutional autoencoder \citep{masci_cae_2011} is used to model $\hat{h}$. This end-to-end approach allows for the use of high-dimensional feature spaces including raw input images. Next, the autoencoder is pretrained using reconstruction loss \citep{goodfellow_dl_2016} to enhance the sample efficiency of the deep neural network. Lastly, the experience replay \citep{lin_self_1992, mnih_dqn_2015} is added to store the feedback $h$ and state-tuples, enabling their reuse during the training process. During training, weights are fine-tuned by computing a weighted sum of TAMER's original loss where the weight is determined by the credit assignment $c(\mathrm{T}^s, \mathrm{T}^e, \mathrm{T}^h)$:
\begin{equation}\label{Deep_TAMER_loss}
\mathcal{L}(\hat{H}) = \sum_{k} c(\mathrm{T}^s_{t-k}, \mathrm{T}^e_{t-k}, \mathrm{T}^h_{t}) \big [ H_t -  \hat{h}(S_{t-k}, A_{t-k} )  \big ]^2.
\end{equation}
The weighting of the squared error is hypothesized to bias the agent's learning towards state-action pairs that were intended to receive more feedback \citep{warnell_deeptamer_2018}. 

Deep TAMER is tested in the Atari Bowling environment using TAMER, A3C \citep{mnih_a3c_2016}, Double-DQN \citep{hasselt_ddqn_2016} as baselines. The average learning curve over all subjects shows a large increase in aptness and robustness due to better short-term and asymptotic performance than the baselines. Further testing shows that Deep TAMER can even outperform human experts within 15 minutes of training. Still, additional studies have shown that Deep TAMER's overall aptitude increase is limited as it struggles in more complex 3D environments \citep{arumugam_dlcoach_2019}. Finally, evaluation of the PDF selection for the credit assignment problem reveals that a uniform distribution $U(.28, 4)$ leads to good performance while a gamma distribution $\text{Gamma}(2, .28)$ vastly underperform. This highlights the importance of selecting the proper PDF to model the delay in human feedback.

%%%%%%%%%%%%%%%%%%%%%%%%%%%%%%%%%%%%%%%%%%%%%%%%%%%%%%%%%%%%%%%%%%%%%%%%%%%%%%%%%%%%%%%%%%%%%%%%%%%%%%%%%%%%%%%%%%%%%%%%%%%%%%%%%%%%%%%

\paragraph{DQN-TAMER}
 \citet{arakawa_dqntamer_2018} aim to improve TAMER+RL by utilizing Deep TAMER for feedback modeling and a DQN \citep{mnih_dqn_2015} as the RL agent. As with TAMER+RL, it takes advantage of both the feedback and environmental reward while attempting to gain aptitude improvements often presented by the various deep RL techniques. DQN-TAMER essentially boils down to using the Deep TAMER to model the feedback $\hat{h}$ and action biasing as the shaping methodology (\eqref{eq:action_biasing}). The action biasing formulation is slightly changed such that $\alpha_q$ weighs $Q(s, a)$ and $\alpha_h$ weighs $\hat{h}(s, a)$ where $\alpha_h$ is annealed overtime. An action is selected in a greedy manner such that
\begin{equation}\label{DQN-TAMER-policy}
\pi(s) = \argmax{a}[\alpha_q Q(s, a) + \alpha_h \hat{h}(s, a) ].
\end{equation}

DQN-TAMER is tested in various grid world environments along with two baselines, Deep TAMER and a DQN. Empirical results that use synthetic human feedback indicate a slight increase in aptness and robustness compared to the baselines. As with TAMER+RL, the scalability of DQN-TAMER is likely to increase as learning can continue even when feedback is no longer provided as a reward function is available. Yet, there are a few caveats to consider. First, DQN-TAMER targets problems where a reward function is available, distinguishing it from Deep TAMER or TAMER, which targets problems where only feedback is available. Moreover, the overall aptitude increase DQN-TAMER presents remains largely unclear as the evaluation is conducted on a limited number of relatively simplistic environments. Another caveat arises from the use of synthetic feedback instead of real human feedback is used for training. While synthetic feedback is useful for testing how different variations in feedback affect performance (e.g., delay, sparsity, etc.), it raises an issue concerning how well the synthetic feedback represents real human feedback. As human feedback tends to be positively biased, policy dependent, and non-stationary, synthetic feedback provided via the Manhattan distance might not produce representative results \citep{knox_framing_2015, macglashan_coach_2017}.

%%%%%%%%%%%%%%%%%%%%%%%%%%%%%%%%%%%%%%%%%%%%%%%%%%%%%%%%%%%%%%%%%%%%%%%%%%%%%%%%%%%%%%%%%%%%%%%%%%%%%%%%%%%%%%%%%%%%%%%%%%%%%%%%%%%%%%%

\subsubsection{COACH}
Convergent Actor-Critic by Humans (COACH) \citep{macglashan_coach_2017} takes an alternative approach to TAMER by integrating human quantitative feedback with actor-critic algorithms through a unique policy shaping method. COACH is built on the assumption that both human feedback and the advantage function are policy dependent, allowing for feedback to directly replace the advantage function. Here policy dependence indicates that the feedback and advantage function values for the agent's current policy are dependent on the quality of prior observed policies. This policy dependence of human feedback is shown empirically by tracking feedback given in relation to humans observing a series of pre-defined good, average, or bad policies. The policy dependent assumption affords the emergence of three desirable training strategies that coincide with how humans tend to provide feedback:
\begin{enumerate}
    \item Diminishing Returns: There is a gradual decrease in positive feedback for good actions as the agent adopts said good actions. This is useful for decreasing the burden of how active the human trainer must be when providing feedback.  
    \item Differential Feedback: Feedback varies in magnitude with respect to the degree of improvement or diminishment. This allows for emphasis to be placed on important behaviors or to communicate urgency for learning.
    \item Policy Dependent Shaping: Positive feedback is provided for sub-optimal actions to improve the behavior and negative feedback is provided after the improvement is made. This form of feedback indicates improvement relative to the current baseline\footnote{\citet{macglashan_coach_2017} originally denoted this idea as ``policy shaping'', however, we adjusted this term to differentiate it from its more general use which we have defined prior.}
\end{enumerate}

The advantage function is then the focus as it is claimed to encapsulate these three policy-dependent training strategies as well. For example, the advantage function inherently possesses the differential feedback property since it provides information about how much better or worse the current action is compared to the previous policy $\pi$. Diminishing return arises due to the fact that $V_\pi(s)$ slowly approximates $Q_\pi(s, a)$ as learning occurs such that the advantage function is ultimately driven towards $\mathbb{A}_\pi(s,a) \rightarrow Q_\pi(s, a) - V_\pi(s)$ as an agent adopts an action $\mathbb{A}_\pi(s,a) \rightarrow 0$. Policy dependent shaping inherently arises as more optimal actions produce positive advantage values and less optimal actions produce negative advantage values. Whether the advantage function value is positive or negative is dependent on the current policy's relationship to prior policies. Thus, sub-optimal actions can be rewarded and then punished as more optimal actions are discovered. 

Given the policy dependent nature of human feedback and the advantage function $\mathbb{A}_\pi(s, a)$, feedback $h$ is then substituted for the TD error (\eqref{eq:temporal_difference_error}) which itself is an unbiased estimate of $\mathbb{A}_\pi(s, a)$ \citep{sutton_rl_2018}. The following equality then holds,
\begin{equation}\label{eq:human_advantage_function}
\begin{split}
\mathbb{A}(s, a) &=  Q(s, a) - V(s) \\
&=  r + \gamma V(s') - V(s) \\
&= h
\end{split}
\end{equation}
such that the human replaces the critic. 

Similarly to TAMER reward aggregation, COACH aggregates feedback ${H}^\prime$ so that the magnitude of feedback can be captured. While COACH accounts for the typical human delay of 0.2 to 0.8 seconds by allowing feedback to be associated with events $d$ steps ago, it does not scale the feedback using a PDF. Instead, COACH utilizes multiple eligibility traces with various decay rates in order to smoothly distribute feedback to past events \citep{sutton_rl_2018}. It allows for various strengths of policy gradient memory to be employed as each trace has a unique decay rate $\lambda$. Each individual eligibility trace vector $e_{\lambda}$ is then updated for each $d$ previous time steps,
\begin{equation}\label{rt_COACH_et}
    e_{\lambda} = \lambda e_{\lambda} + \nabla \log \pi_{\theta_t}(A_{t-d}|S_{t-d}).
\end{equation}
A single eligibility trace is either explicitly selected by the human trainer or implicitly selected based on the value of the aggregated feedback magnitude and used to update the actor's parameters as follows:
\begin{equation}\label{rt_COACH_et_eq:actor_update}
    \theta_{t+1} = \theta_t + \alpha \, {H}_t^\prime \, e_{\lambda},
\end{equation}
where ${H}_t^\prime$ denotes the aggregated feedback.

COACH performs well in a relatively complex environment that requires a human trainer to teach various behaviors (e.g., navigation, following, etc) to a TurtleBot robot using the aforementioned policy dependent training strategies and an additional strategy of compositional learning. Here compositional learning is a strategy where sub-behaviors are first learned and then stitched together to learn a new behavior. COACH shows an increase in robustness as it is able to learn all behaviors using all training strategies. This is unlike the baseline TAMER which is unable to learn behaviors requiring compositional learning. Likewise, COACH shows an increase in aptness as it learns all behaviors within 2 minutes. Despite the fact that COACH proves to be viable, and potentially better alternative to the TAMER framework, it remains unclear how it performs in comparison to the other variations of TAMER such as the actor-critic TAMER \citep{vien_actamer_2013} or TAMER+RL \citep{knox_combining_2010, knox_reinforcement_2012}. Just like TAMER, COACH's overall increase in aptitude is limited due to its linear modeling of the policy and poor generalization in more complex environments \citep{arumugam_dlcoach_2019}. Lastly, It is unknown to what degree COACH increases aptitude when compared to traditional RL algorithms due to the lack of baseline comparisons. Below we briefly discuss a variation of COACH which attempts to further address the shortcomings of vanilla COACH. 

%%%%%%%%%%%%%%%%%%%%%%%%%%%%%%%%%%%%%%%%%%%%%%%%%%%%%%%%%%%%%%%%%%%%%%%%%%%%%%%%%%%%%%%%%%%%%%%%%%%%%%%%%%%%%%%%%%%%%%%%%%%%%%%%%%%%%%%

\paragraph{Deep COACH}
Similar to Deep TAMER, \citet{arumugam_dlcoach_2019} extend COACH by incorporating a DQN \citep{mnih_dqn_2015} to increase aptitude and deal with high-dimensional state spaces. There are three new components that are added to the base COACH framework: a convolutional autoencoder, experience replay, and entropy regularization. Just as in Deep TAMER, a pretrained convolutional autoencoder is used to enable raw image input and improve the sample efficiency through the unsupervised pre-training with reconstruction loss.

Instead of using a basic replay buffer that stores single experiences (i.e., a state-tuple and feedback), Deep COACH modifies the buffer to store windows of experience (i.e., trajectories). A trajectory is stored whenever feedback is received. Upon storage, the feedback $h$ is mapped to the last state transition in the trajectory such that reward aggregation is no longer done. Importance sampling $\rho = \frac{\pi_{\theta_t} (A_{t-1} | S_{t-1})}{\pi_{\theta_{t-1}} (A_{t-1} | S_{t-1})}$ is also used to account for the discrepancy between the current policy $\pi_{\theta_t}$ and the behavior policy $\pi_{\theta_{t-1}}$ at $t-1$. The actor's eligibility trace (\eqref{rt_COACH_et}) is then updated as 
\begin{equation}\label{eq:deep_COACH_is_update}
 e_{\lambda} =  \lambda e_{\lambda} + \rho \nabla \log \pi_{\theta_{t}}(A_{t-1}|S_{t-1}),
\end{equation}
where $e_{\lambda}$ is updated for each transition in the trajectory. Once all transitions for a window have been used to update $e_{\lambda}$, $h$ is then applied to $e_{\lambda}$ such that
\begin{equation}\label{eq:deep_COACH_et_window_update}
    \bar{e}_{\lambda} = \bar{e}_{\lambda}  + h \ e_{\lambda},
\end{equation}
where $\bar{e}_{\lambda}$ is updated for every trajectory in a minibatch of size $m$.

To create an agent that is more responsive to feedback and not permanently biased by inconsistencies or natural errors, entropy regularization is utilized \citep{griffith_policy_2013}. The entropy regularization term is defined as $\beta \mathbb{H}(\pi(\cdot|S_t))$ where $\mathbb{H}(\pi(\cdot|S_t))$ is an entropy function that takes the action probability distribution $\pi(\cdot|S_t)$ and the regularization scaling parameter $\beta$. Entropy in actor-critic networks encourages exploration such that the agent might avoid convergence to a local minimum. The average eligibility trace over the minibatch $\bar{e}_{\lambda}$ and the entropy term are used to update the actor's parameter as 
\begin{equation}\label{eq:deep_COACH_eq:actor_update}
\theta_{t+1} = \theta_{t} + \alpha \bigg [  \frac{1}{m} \bar{e}_{\lambda} + \beta \nabla \mathbb{H}(\pi_{\theta_t}(\cdot|S_t)) \bigg ].
\end{equation}

Deep COACH is tested in the relatively complex simulated environment of Minecraft where different behaviors (e.g., navigating to a goal and walking the perimeter of a room) must be learned. Deep COACH shows an increase in aptness and robustness over the baselines of Deep TAMER and COACH as it achieves greater short-term and asymptotic performance for all behaviors. Deep TAMER learns to perform all behaviors at a sub-par level while COACH fails to learn some of the behaviors. Furthermore, Deep COACH shows potential increases in scalability as it requires significantly less feedback to learn all behaviors compared to that of the baselines. Two main caveats of Deep COACH's potential aptitude increase arise as follows. First, due to the limited testing environments, the overall increase in aptitude is unknown. Second, the degree of aptitude increase when compared to traditional RL algorithms is not examined and thus also unknown. 

%%%%%%%%%%%%%%%%%%%%%%%%%%%%%%%%%%%%%%%%%%%%%%%%%%%%%%%%%%%%%%%%%%%%%%%%%%%%%%%%%%%%%%%%%%%%%%%%%%%%%%%%%%%%%%%%%%%%%%%%%%%%%%%%%%%%%%%

\subsubsection{Learning from Preferences}\label{sec:learning_from_human_preferences}
 \citet{christiano_preferences_2017} present an alternative learning from preferences (LfP) framework that is quite distinct from the frameworks discussed so far. This novel approach specifically centers around preference-based learning, aiming to extract a latent reward function from qualitative feedback expressed through preferences between behaviors. Intuitively, this framework can be interpreted as combining feedback modeling with reward shaping. Here the feedback model is designed to learn a latent numerical representation $\hat{h}$ that is interpreted as the human's latent reward function. In order to train an RL agent, reward shaping with replacement is employed such that $\hat{h}(s, a)$ replaces the reward function $r(s, a)$.

In the proposed LfP framework, humans contribute preference feedback $h$ by evaluating pairwise trajectory segments drawn from the agent's experiences. Informally, a trajectory segment is just a short clip from an agent interacting with an environment (i.e., clip of a behavior). Formally, a trajectory segment $\sigma$ is a short sequence of observations (i.e., states) $O_t \in \mathcal{O}$ and actions $A_t \in \mathcal{A}$ of length $k$ such that
\begin{equation}\label{eq:trajectory_segment}
    \sigma = [(O_0, A_0), (O_1, A_1), ..., (O_{k-1}, A_{k-1})] \in (\mathcal{O}, \mathcal{A})^k.
\end{equation}
The first trajectory segment $\sigma^1$ is said to be preferred to another $\sigma^2$ if $\sigma^1 \succ \sigma^2$ such that
\begin{equation}\label{eq:trajectory_segment_preference}
    \begin{split}
        & [(O^1_0, A^1_0), (O^1_1, A^1_1), ..., (O^1_{k-1}, A^1_{k-1})] \\
        & \succ [(O^2_0, A^2_0), (O^2_1, A^2_1), ..., (O^2_{k-1}, A^2_{k-1})].
    \end{split}
\end{equation}
Given a reward function $r : \mathcal{O} \times \mathcal{A} \rightarrow \mathbb{R}$, the authors make the assumption that if a trajectory segment is preferred over another, the preferred trajectory segment must have a higher total reward:
\begin{equation}\label{eq:trajectory_segment_reward_preference}
    \begin{split}
        & r(O^1_0, A^1_0) + ... + r(O^1_{k-1}, A^1_{k-1}) \\
        & > r(O^2_0, A^2_0) +  ... + r(O^2_{k-1}, A^2_{k-1}).
    \end{split}
\end{equation}
As the true reward $r$ is unknown, it is assumed that $r$ can be approximated by modeling a latent reward function $\hat{h}$ using preferences, leaving only a traditional RL problem left to be solved. 

The estimated probability of preferring a trajectory segment over another is modeled with the softmax function of the sum of latent rewards as 
\begin{equation}\label{eq:trajectory_probability_preference}
\begin{split}
\hat{P}(\sigma^1 \succ \sigma^2) = \frac{\exp \sum \hat{h}(O^1_t, A^1_t)}{\exp \sum \hat{h}(O^1_t, A^1_t) + \exp \sum \hat{h}(O^2_t, A^2_t)}.
\end{split}
\end{equation}
Given the feedback label $h(\cdot)$ and the estimated preference probability $\hat{P}$, then $\hat{h}$ can be learned by minimizing the binary cross-entropy loss: 
\begin{equation}\label{eq:cross_entropy_preference}
\begin{split}
\mathcal{L}(\hat{h}) = - \sum_{\substack{(\sigma^1, \sigma^2, h) \in D}} & h(1)\log \hat{P}(\sigma^1 \succ \sigma^2) + h(2) \log \hat{P}(\sigma^2 \succ \sigma^1).
\end{split}
\end{equation}
The received preference feedback $h(\cdot)$ acts as a distribution over the pairwise trajectories where the mass is centered on the preferred trajectory segment. For example, if the first behavior is preferred, then $h(1) = 1$ and $h(2) = 0$. If both trajectories are said to be preferred, then distribution mass is distributed uniformly over the trajectory segments.

Using the benchmark environments of Atari and MuJoCo environments \citep{brockman_gym_2016}, empirical results with human preferences are mixed as the proposed algorithm does not always achieve comparable short-term or asymptotic performance with that of the Advantage Actor-Critic (A2C) \citep{mnih_a3c_2016} and Trust Region Policy Optimization (TRPO) \citep{schulman_trust_2017} baselines. Depending on the environment, these results suggest a decrease in both robustness and aptness. Albeit, there is a potential for a significant increase in scalability as preference feedback allows for reduction in the number of interactions needed by up to 3 orders of magnitude. Thus, whether there is a general increase in aptitude remains largely unclear as it hinges on what is considered ``sufficient'' performance in each environment. A further caveat here is that some environments still require a large number of preference feedback samples (e.g., 5.5k for Atari environments) for learning to occur. This brings about concerns surrounding how scalable the proposed algorithm is as more preference feedback can be expected to be needed as the environment complexity increases. As a comparative evaluation against other interactive RL algorithms has not been made, its standing among the other frameworks is unclear.   

%%%%%%%%%%%%%%%%%%%%%%%%%%%%%%%%%%%%%%%%%%%%%%%%%%%%%%%%%%%%%%%%%%%%%%%%%%%%%%%%%%%%%%%%%%%%%%%%%%%%%%%%%%%%%%%%%%%%%%%%%%%%%%%%%%%%%%%

\section{Learning from Intrinsic Feedback}\label{sec:iirl}
The point of intersection between interactive RL and BCI lies in the intrinsic communication of human input through informative brain signals extracted from neural activity. These informative brain signals can simply be interpreted as a new feedback medium, enabling the utilization of the previously discussed interactive RL and LfF concepts and frameworks. The interactive RL setup can be adjusted to account for the BCI Pipeline as depicted in \figref{fig:iirl-obs-all}. By leveraging the BCI pipeline, which captures and extracts informative brain signals to be interpreted as feedback, and using LfF frameworks, the feedback can be effectively utilized for learning. 

Now, we can further narrow the critical questions presented in the \nameref{sec:irl} section to specifically focus on intrinsic feedback. In doing so, we can inherit many of the motivations, approaches, and ideas previously discussed while also adding new ones that are specific to intrinsic feedback. Additionally, a new critical question specific to intrinsic feedback is needed in order to understand which brain signals can be interpreted as feedback:
\begin{enumerate}[label=(Q\arabic*)]
    \setItemnumber{4}
    \item What informative brain signals can be decoded from neural activity and used as feedback? \label{item:irl-Q4}
\end{enumerate}

The remainder of this section will be structured around these four critical questions, which will drive the  discussions on intrinsic feedback. The \nameref{sec:irrl_motivation} section will aim to address critical question \ref{item:irl-Q1}, now with a specific focus on motivations for utilizing intrinsic feedback. The \nameref{sec:errp_feedback} section will attempt to address critical question \ref{item:irl-Q4} by understanding how ErrPs, and potentially other known brain signals, can be interpreted as feedback. Lastly, the \nameref{sec:current_approaches} section aims to address critical questions \ref{item:irl-Q2} and \ref{item:irl-Q3} by covering the most recent approaches that have employed intrinsic feedback.

%%%%%%%%%%%%%%%%%%%%%%%%%%%%%%%%%%%%%%%%%%%%%%%%%%%%%%%%%%%%%%%%%%%%%%%%%%%%%%%%%%%%%%%%%%%%%%%%%%%%%%%%%%%%%%%%%%%%%%%%%%%%%%

\subsection{Intrinsic Feedback Motivations}\label{sec:irrl_motivation}
To address \ref{item:irl-Q1} with respect to intrinsic feedback, we can build upon the motivations outlined in the \nameref{sec:irl_motivation} section and introduce two additional motivations: leveraging the abundant latent information within the brain and the automatic elicitation of feedback. The primary motivation stems from the desire to tap into the vast pool of latent information present in the brain, which many BCI researchers aim to utilize. Extracting signals corresponding to error processing, affective states, and cognitive states can serve as a proxy to human beliefs \citep{kim_intrinsic_2017, alnafjan_emotionsurvey_2017, li_emotionsurvey_2022, myrden_cognitivestates_2017}. Albeit, effectively extracting this information remains a challenge. This is largely due to the paradoxical nature of the brain as it acts as an immense concentration of potential information while at the same time this means extracting a specific piece of information is extremely difficult. This problem is akin to the Cocktail Party Problem where the goal is the separation of the desired signal from as much of the surrounding noise as possible. While identification and extraction of informative brain signals is non-trivial, our ability to do so has nevertheless progressed over the past decade \citep{craik_bcisurvey_2019, lotte_bcisurvey_2018, luck_erpbook_2014}.

The secondary motivation, and potentially even more important one, arises from the fact that this latent information in the brain is automatically elicited. Although automatically elicited informative brain signals might not directly correspond to feedback, they can often be interpreted as an indirect or implied forms of feedback. One significant benefit of this automatically elicited feedback is that humans can freely engage in or observe a task without the need for feedback to be their primary concern. One can think of this as being able to ``unconsciously'' elicit feedback, although some degree of focus will always be required \citep{wessel_erroraware_2012}. Additionally, the automatic elicitation of feedback has generated interest in the notion that intrinsically provided feedback can be obtained at a faster and more consistent rate compared to explicit feedback \citep{luo_drlpref_2018, xu_accelerating_2021}. However, it is important to note that this area of research remains relatively inconclusive and necessitates further exploration. 

\begin{figure}[t!]
    \centering
    \subfloat[LfF setup using intrinsic feedback]{\includegraphics[scale=.35]{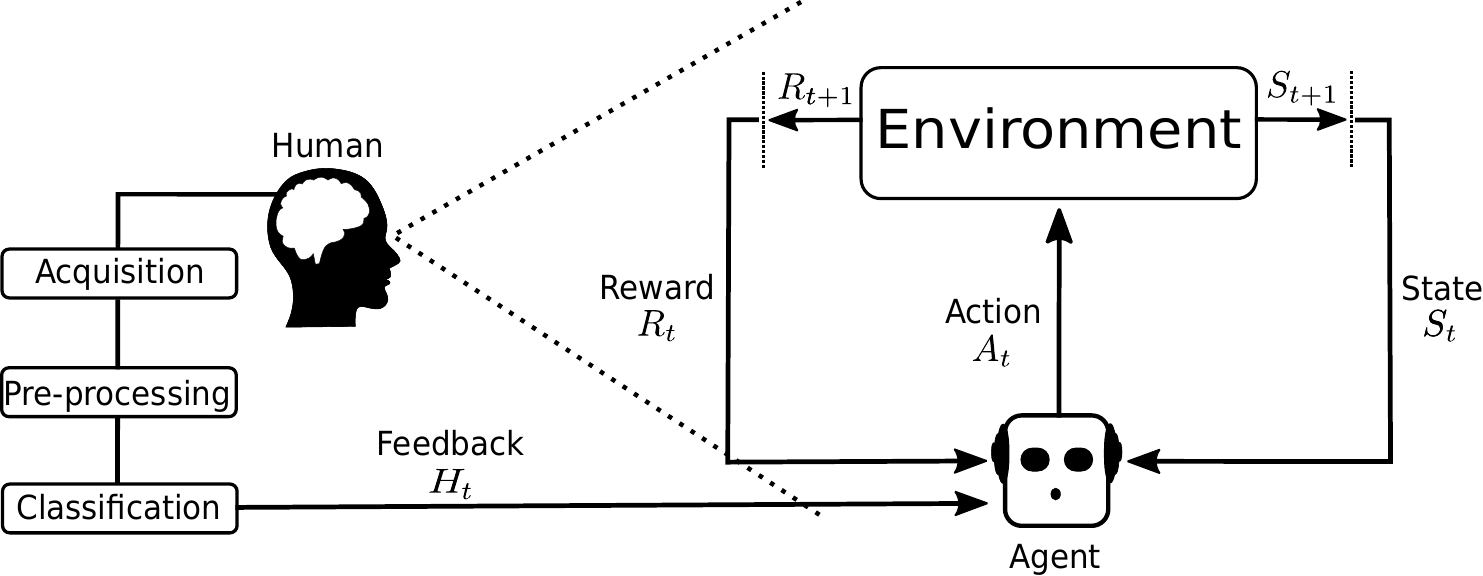}
        \label{fig:iirl-obs-all}}
        \hfil
    \subfloat[Setup combining the LfF and LfD setups using intrinsic feedback]{\includegraphics[scale=.35]{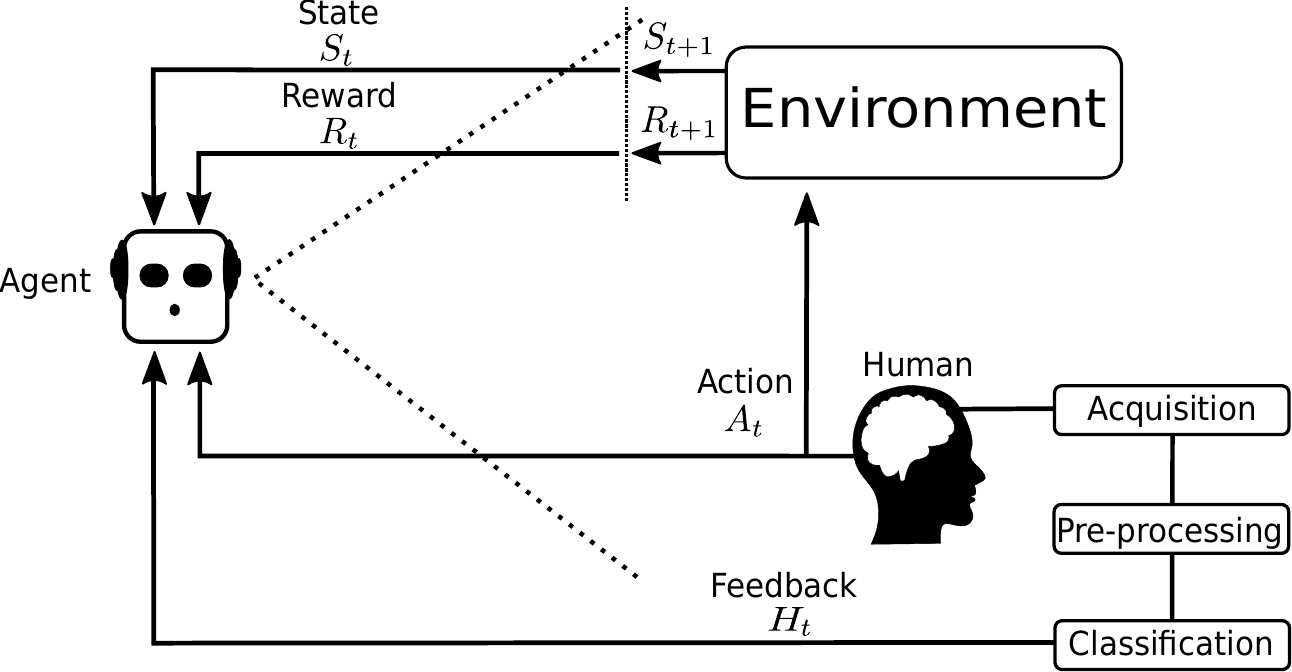}
        \label{fig:iirl-int-all}}
        \hfil
    \caption{Modified interactive RL setups that now account for intrinsic feedback. While the reward function is included in both setups, it is not required. (\subref{fig:iirl-obs-all}) The LfF setup where the human provides intrinsic feedback when observing an agent. This setup can be easily adapted for preference-based qualitative feedback as well. (\subref{fig:iirl-int-all}) A new interactive RL setup combining LfF and LfD setups by having the human perform a task while providing intrinsic feedback. }
    \label{fig:iirl}
\end{figure}

Another benefit of automatically elicited guidance arises by allowing for the formulation of new interactive RL setups. Typically, as depicted in \figref{fig:iirl-obs-all} and \figref{fig:fb}, LfF setups require the human to observe the agent and provide feedback based on their observation. For instance, let us consider how we might combine both feedback and demonstrations to allow for additional information to be transferred to the agent. In the conventional approach, this would require a sequential setup where the LfD setup (\figref{fig:lfd-all}) is used to initially train the agent, followed by further refinement of the agent using the LfF setup \citep{ibarz_rewardlearning_2018}. When using explicit feedback, it is extremely difficult to have a human perform a task while simultaneously providing feedback on their own actions. Thus, the context regarding the quality of a given demonstration remains unknown unless it is meticulously reviewed and labeled retrospectively. Intrinsic, and often implicit, feedback allows for the parallel combination of other forms of human input. In the case of combining demonstrations and intrinsic feedback, \figref{fig:iirl-int-all} shows how the human could perform a task while providing feedback regarding how good or bad their own actions were. Having access to such information could prevent LfD approaches like behavioral cloning from replicating undesirable behaviors. At a high level, this can be viewed as a form of automatic data labeling, where the human labels their demonstrations with their intrinsic feedback, resulting in two interconnected sources of human input (i.e., multi-modal). To our best knowledge, the implications of having two interconnected sources of human input have not been extensively explored.

%%%%%%%%%%%%%%%%%%%%%%%%%%%%%%%%%%%%%%%%%%%%%%%%%%%%%%%%%%%%%%%%%%%%%%%%%%%%%%%%%%%%%%%%%%%%%%%%%%%%%%%%%%%%%%%%%%%%%%%%%%%%%%

\subsection{Brain Signals as Feedback}\label{sec:errp_feedback}
To address \ref{item:irl-Q4}, we now delve into what types of brain signals can be interpreted as feedback, with a particular focus on ErrPs. The persistent interest of BCI researchers in ErrPs is driven by their connection to erroneous events encountered by humans. This makes it possible to utilize the presence or absence of an ErrP as an indication of an event's ``correctness.'' \citep{kim_intrinsic_2017, salazar_errprobot_2017, ferrez_errpwrong_2005, ferrez_errp_2008, chavarriaga_errplearning_2010, lopesdias_errp_2019, ehrlich_errpfeasibility_2019, ehrlich_errpcoadapt_2018, spuler_continuouserrp_2015, iturrate_robot_2010, iturrate_errprl_2015, iturrate_errptask_2013, luo_drlpref_2018, xu_accelerating_2021}. For example, ErrPs have been classically employed as a feedback mechanism to rectify erroneous responses made by mental typewriter applications when a wrong letter or symbol is selected by the interface \citep{schmidt_mentaltype_2012}. Additionally, ErrPs have been used to correct the actions of rudimentary agents. A notable example is the work by \citet{salazar_errprobot_2017}, where the detection of ErrPs is used to correct a robot's action when it chooses an incorrect action. Likewise, \citet{chavarriaga_errplearning_2010} use ErrPs to teach an agent a naive policy for a simple goal search task. When the agent makes the incorrect choice of moving away from the goal and an ErrP is detected, the probability of taking that particular action is subsequently decreased. 

Thus, it can be valuable to further interpret ErrPs as a proxy to a human's subjective ground truth regarding what they think is correct or incorrect \citep{kim_intrinsic_2017}. As one might begin to see, the utilization of ErrPs to assess the correctness of an agent's actions or behaviors aligns well with the concept of feedback. In particularly evaluative feedback, which aims to evaluate the quality of an action. Consequently, a majority of the intrinsic feedback approaches, basically all the approaches we will cover, adopt ErrPs as the brain signal of interest.

Typically, the presence of an ErrP can be interpreted as an action or behavior being wrong. Upon detection, negative quantitative (e.g., -1) or qualitative (e.g., a categorical variable for ``wrong'' or ``bad'') feedback could be provided to the agent \citep{kim_intrinsic_2017, iturrate_robot_2010, ehrlich_errpcoadapt_2018}. Research on ErrPs has even attempted to detect the severity of an erroneous event, potentially providing more fine-tuned information concerning how wrong an action was \citep{spuler_continuouserrp_2015, wirth_4way_2020}. Meanwhile, the absence of an ErrP can be interpreted in two ways: the absence of feedback or positive feedback. Firstly, the absence of an ErrP can be interpreted simply as the absence of feedback, meaning no feedback is given. Secondly, under an assumption similar to that of the legal principle of presumption of innocence, the absence of an ErrP can indicate what a human considers correct. Therefore, by default, positive quantitative (e.g., +1) or qualitative (e.g., a categorical variable for ``correct'' or ``good'') feedback could be provided to the agent until an ErrP is detected \citep{kim_intrinsic_2017, chavarriaga_errplearning_2010, ehrlich_errpcoadapt_2018, iturrate_robot_2010, iturrate_errprl_2015, xu_accelerating_2021}.

Given ErrPs can be treated as feedback, accurate detection of ErrPs is desirable as this determines the accuracy of the feedback. Poor detection of ErrPs means that inaccurate feedback will then be communicated to an agent, potentially leading to either slowed or disrupted learning. Even though classification of ErrPs has been a long standing area of research \citep{ferrez_errpwrong_2005, ferrez_errp_2008, lopesdias_errp_2019, spuler_continuouserrp_2015, kim_transfer_2013, kim_handlingtransfer_2016, abu_errpinvariance_2017, lawhern_eegnet_2018, ehrlich_errpfeasibility_2019, iturrate_errptask_2013, kakkos_errpcomplex_2020}, performance can vary greatly depending on factors such as the subject, data quality (e.g., SNR or non-stationarity), and data quantity \citep{lotte_bcisurvey_2018}. Yet, prior works that have paired relatively poor classification models with interactive RL have shown that the algorithms are often still able to learn, indicating that RL has some inherent robustness to a certain level of noisy reinforcement signals \citep{kim_intrinsic_2017, xu_accelerating_2021, wang_maximizing_2020, akinola_accelerated_2020, iturrate_robot_2010}.

In regards to alternative brain signals to be used as feedback, this has largely been an unexplored area. It could be desirable to use multiple types of brain signals as feedback, as this would allow for a wider array of feedback to be conveyed. One natural direction to investigate would be signals that can be interpreted as positive reinforcement. \citet{wirth_goal_2020} attempt use the P300 component as a form of positive feedback. They find it is possible to discriminate between different magnitudes of correctness when a simulated robot moves towards the goal versus when it reaches the goal (the latter being consider ``more correct''). \citet{wirth_4way_2020} extend this idea to detect multiple magnitudes of both positive and negative feedback using the P300 and ERN components. However, whether utilizing P300 as positive feedback is generalizable to other tasks/environments remains to be explored.

Furthermore, if we assume that ErrPs and other ERPs (e.g., P300) act as short-term feedback that allows for fine-tuned feedback to be applied event-by-event (i.e., state-by-state), then cognitive and affective states are likely a natural fit a medium- to long-term feedback as they can reflect a human's general feelings concerning the agent's trajectory \citep{chakraborti_hitlbci_2017}. In particular, negative and positive affective states naturally fit with being interpreted both as negative and positive feedback. For instance, emotions like frustration, stress, or other negative valence emotions can be interpreted as forms of negative feedback. On the other hand, emotions like happiness, excitement, and other positive valence emotions can be interpreted as forms of positive feedback.

%%%%%%%%%%%%%%%%%%%%%%%%%%%%%%%%%%%%%%%%%%%%%%%%%%%%%%%%%%%%%%%%%%%%%%%%%%%%%%%%%%%%%%%%%%%%%%%%%%%%%%%%%%%%%%%%%%%%%%%%%%%%%%%%%%%%%%%

\subsection{Current Approaches}\label{sec:current_approaches}
By harnessing the foundations established in the \nameref{sec:LfF} section, we can now address \ref{item:irl-Q2} and \ref{item:irl-Q3} in the context of intrinsic feedback. While the aforementioned LfF frameworks can be easily adapted to incorporate intrinsic feedback \citep{luo_drlpref_2018}, it is important to highlight recent works that specifically focus on new LfF approaches that take into consideration the downsides of intrinsic feedback. To do so, many of these approaches draw inspiration from the foundational LfF frameworks \citep{xu_accelerating_2021, akinola_accelerated_2020, wang_maximizing_2020}.

With the relatively recent emergence of intrinsic feedback, only a limited number of works have been produced \citep{iturrate_robot_2010, iturrate_errprl_2015, ehrlich_errpcoadapt_2018, kim_intrinsic_2017, kim_flexible_2020, xu_accelerating_2021, luo_drlpref_2018, akinola_accelerated_2020, wang_maximizing_2020}. Thus, we attempt to provide a comprehensive review of these known works. Nonetheless, the categorization of the intrinsic feedback literature provides valuable insights into current approaches and potential future directions. We classify these works into three categories: proof-of-concept, adaptive, and explorative. Proof-of-concept approaches focus on demonstrating the feasibility and potential benefits of using intrinsic feedback. This is done by testing the combination of BCI and interactive RL in different simulated and real-world toy environments. Adaptive approaches seek to adapt existing LfF frameworks to function with intrinsic feedback. Explorative approaches delve into the exploration of intrinsic feedback while also improving upon existing LfF ideas. This is typically done by introducing ideas that aim to compensate for the misclassification of intrinsic feedback so that the impact of a noisy feedback signal can be reduced. In the following sections, we provide a brief qualitative analysis of each work from the perspective of aptitude and alignment alleviations. To facilitate comparison, a summary of these intrinsic feedback works is provided in Table~\ref{tab:irl_works}. 

%%%%%%%%%%%%%%%%%%%%%%%%%%%%%%%%%%%%%%%%%%%%%%%%%%%%%%%%%%%%%%%%%%%%%%%%%%%%%%%%%%%%%%%%%%%%%%%%%%%%%%%%%%%%%%%%%%%%%%%%%%%%%%%%%%%%%%%

\subsubsection{Proof-of-Concept}
The majority of the prior intrinsic feedback works follow a proof-of-concept approach where the goal is to simply test if the combination of BCI and RL allows agents to learn in different simulated and real-world toy environments \citep{iturrate_robot_2010, iturrate_errprl_2015, ehrlich_errpcoadapt_2018, ehrlich_errpfeasibility_2019, kim_intrinsic_2017, chavarriaga_robust_2016}. It is important to note that these proof-of-concept works were primarily conducted by BCI researchers who recognized a connection to RL but may not have explicitly explored interactive RL, which was also developing around the same time. This can be inferred by observing the references cited in these works, which often lack references to interactive RL, with the exception of \citet{kim_intrinsic_2017}. Regardless, these approaches are often characterized by their use of reward shaping with replacement and standard RL algorithms such as Q-learning. Consequently, these works do not provide a detailed analysis that we can use to examine \ref{item:irl-Q3}, as their primary focus is not on conducting comparisons to other algorithms.

For example, \citet{iturrate_robot_2010} utilize Q-learning in a discrete action space task where a simulated robot arm learns to select the central object given a line of five objects. If the robot selects the wrong object and an ErrP is detected, quantitative feedback $h = \{-0.5, -1\}$ is given depending on the severity of the error. If the robot performs the correct action and no ErrP is detected, $h = +1$. Using Q-learning and reward shaping with replacement, the robot eventually learns the correct policy of grabbing the central object. 

Similarly, \citet{kim_intrinsic_2017} utilize ErrPs to teach a robotic arm to perform a predefined action when a specific hand gesture known only to the human trainer is performed. In the experiment, the robot has no prior knowledge of which predefined action belongs to which gesture, only the human trainers know the correct mapping. Thus, the robot must explore each action until the correct mapping is learned by using ErrPs which are elicited in response to the robot performing wrong actions. When an ErrP is detected due to either an incorrect action or misclassification, the quantitative feedback is given as $h = 0$. Otherwise, by default, $h = +1$. Training is done using Linear Upper Confidence Bound (LinUCB) and reward shaping with replacement until the robot converges to the human's selected gesture-to-action policy.

Lastly, \citet{ehrlich_errpcoadapt_2018} have a robot and human collaborate to find a common policy for selecting an object. The robot is in charge of selecting one of three objects and the human must figure out which object the robot has selected. To help the human, the robot must learn a gazing policy that maps the object it has selected to the object it is currently gazing at. If the robot learns a good gazing policy, the robot can communicate to the human which object it has selected. When the human makes a guess concerning which object the robot has selected, an interface indicates to the human whether they are right or wrong. If the human is right and no ErrP is detected,  quantitative feedback is given as $h = +1$. Otherwise, if the human was wrong and an ErrP is detected, $h = -1$. The robot learns a gazing policy through a naive formulation of policy gradient where reward shaping with replacement is used once again. 

%%%%%%%%%%%%%%%%%%%%%%%%%%%%%%%%%%%%%%%%%%%%%%%%%%%%%%%%%%%%%%%%%%%%%%%%%%%%%%%%%%%%%%%%%%%%%%%%%%%%%%%%%%%%%%%%%%%%%%%%%%%%%%%%%%%%%%%

\subsubsection{Adaptive and Explorative}
Moving beyond proof-of-concepts are what we define as adaptive and explorative approaches. Adaptive approaches integrate existing LfF approaches with intrinsic feedback. Adaptive approaches provide an opportunity show useful comparisons between intrinsic feedback and the other feedback mediums (often explicit feedback). On the other hand, explorative approaches propose new approaches for integrating intrinsic feedback. As mentioned before, these explorative approaches often try to compensate for the misclassification of the intrinsic feedback (i.e., noisy feedback signal). 

 \citet{luo_drlpref_2018} propose adapting the LfP setup proposed by \citet{christiano_preferences_2017} with intrinsic feedback. This is done by using ErrPs as intrinsic feedback for determining a subject's preference between two trajectory segments. The preference selection process is adapted to work with intrinsic feedback such that when two trajectory segments, $\sigma^1$ and $\sigma^2$, are presented, a subject must determine within their mind which trajectory segment is preferred. After observing the two segments, the interface randomly selects one. If the segment selected does not match the segment selected by the human, an ErrP is expected to be elicited. Likewise, no ErrP is expected to be elicited when selected trajectory matches the one selected by the interface. Once the preference is given through the intrinsic feedback $h$, the latent human reward function $\hat{h}$ is modeled and fed to an RL algorithm for learning as originally proposed by \citet{christiano_preferences_2017}.

Empirical results are given for intrinsic feedback and explicit feedback using a subset of the MuJoCo testing environments. Across a majority of subjects, it is found that intrinsic feedback shows a slight decrease in robustness and aptness to that of explicit feedback. This is surprising as the average ErrP classification accuracy across subjects is only $67\%$, meaning the preference feedback was quite noisy. It appears that enhanced classification performance may enable intrinsic feedback to achieve comparable or potentially superior performance to that of explicit feedback. Finally, scalability is suggested to be further increased as it is reported that intrinsic feedback can be captured at a faster rate than explicit feedback. That being said, these scalability results require further replication and testing in all of the original environments used by \citet{christiano_preferences_2017} is needed for a more comprehensive comparison with explicit feedback.

Moving towards explorative direction, \citet{xu_accelerating_2021} propose a LfF approach that takes advantages of ideas very similar to those used in the LfP setup proposed by \citet{christiano_preferences_2017}. The core of their approach aims to learn a latent numerical representation that is interpreted as the human Q-function $\hat{h} = Q_h$. Given $\hat{h}$, the human reward function and policy can then be extracted. This approach differs from \citet{christiano_preferences_2017} as general qualitative feedback is now used instead of preference-based feedback and no direct modeling of the human reward function is done. To model the latent human Q-function $\hat{h}_\theta$, the authors take advantage of a newer RL methodology denoted as soft Q-learning \citep{haarnoja_soft_2017}. Using $\hat{h}_\theta$, a soft human policy $\pi_{h}$ and human reward function $r_{h}(s, a)$ are further derived. Reward shaping is finally employed where human reward function $r_{h}(s, a)$ is used to perturb existing reward function $r(s, a)$. The augmented reward function $r^\prime(s, a)$ is then fed to any desired RL algorithm for learning.

The modeling of the soft human Q-values $\hat{h}_\theta$ is done to increase the overall resilience of the approach to noise in the feedback signal which is caused by the misclassification of ErrPs. The resilience of soft Q-learning algorithm inherently emerges from the entropy term $\mathbb{H}(\pi(\cdot|S_t))$ which is factored directly into the reward maximization formulation. Maximization of both the future reward and uncertainty helps to encourage exploration and prevent convergence to poor local minimums that can arise due to the misclassification of intrinsic feedback $h$ \citep{xu_accelerating_2021, haarnoja_soft_2017}. To estimate $\hat{h}_\theta$, ErrPs are formulated as qualitative feedback such that the detection of an ErrP indicates an agent's behavior is ``incorrect.'' Feedback is given based on generated trajectories but can also be given for every state-action pair where the former puts less burden on the human. With these ideas, it is possible to model $\hat{h}_\theta$ using a neural network with parameters $\theta$ by minimizing the binary cross-entropy using samples drawn from a replay buffer as follows:
\begin{equation}\label{eq:human_cross_entropy}
    \begin{split}
        L(\theta) = - \sum_{(s,a) \in D} (1 - h) \log \pi_{h}(a|s) + h \log (1 - \pi_{h}(a|s)).
    \end{split}
\end{equation}
 Here $\pi_{h}(a|s)$ represents the probability of the human taking the action (i.e., probability of receiving positive feedback) while $1 - \pi_{h}(a|s)$ acts as the probability the human does not take the action (i.e., probability of receiving negative feedback). Additionally, the labels $h \in \{0, 1\}$ correspond to whether an ErrP was detected $h = 1$ or not $h = 0$. The human policy $\pi_{h}(a|s)$ and state-value function $V_h(s)$ are derived by using the soft policy formulation \citep{haarnoja_soft_2017}:
\begin{equation}\label{eq:soft_human_policy}
    \begin{split}
        \pi_{h}(a|s) & = \exp \big [(\hat{h}_\theta(s, a) -  V_{h}(s)) / \alpha \big ], \\
        V_{h}(s) & = \alpha \log \sum_a \exp(\hat{h}_\theta(s,a) / \alpha),
    \end{split}
\end{equation}
where $\alpha$ is a manually selected energy coefficient. Using the Bellman equation, the human reward function $r_{h}(s, a)$ can be approximated by the difference between the current state's human Q-value and the discounted human Q-value of the next state as follows:
\begin{equation}\label{eq:estimate_human_feedback}
    r_{h}(s, a) = \hat{h}_\theta(s, a) - \gamma \max_{a'} \hat{h}_\theta(s',a').
\end{equation}
Due to the limited number of labeled ErrP examples, the variance of \eqref{eq:estimate_human_feedback} is updated in an attempt to lower the variance of $r_{h}(s, a)$ by utilizing a learned baseline function $t_\phi(s)$:
\begin{equation}\label{eq:aux_reward}
    r_{h}(s, a) = \hat{h}_\theta(s, a) + t_\phi(s) - \gamma \max_{a'}[\hat{h}_\theta(s',a') + t_\phi(s')].
\end{equation}
The baseline parameters $\phi$ are then learned by minimizing \eqref{eq:aux_reward} using both labeled and unlabeled data so that that state dynamics can be more efficiently learned. Finally, $r_{h}(s, a)$ is used to perturb $r(s, a)$ to create an augmented reward function 
\begin{equation}\label{eq:xu_augmented_reward_function}
    r'(s, a) = r(s, a) + \beta \, r_{h}(s, a),
\end{equation}
where the scaling factor $\beta$ is exponentially decayed over time. The augmented reward function $r'(s, a)$ is then passed to a Bayesian DQN for learning.

For the majority of subjects, empirical results show an increase in aptness as the algorithm converges to the optimal behavior more quickly than that of the Bayesian DQN baseline in a grid world (e.g., maze and wobble) and an Atari-like environments (e.g., catcher). Scalability is likely to be increased for three reasons. First, when available, the environmental reward can be utilized. Second, the utilization of unlabeled data has a positive impact on learning. Thirdly, the proposed approach only requires feedback on selected trajectories instead of every state-action pair. Furthermore, the authors compare their approach with an adaptive implementation of a newer explicit feedback approach called FRESH \citep{xiao_fresh_2020} where a similar increase in aptness and scalability can be observed as well. Overall, these results are promising, considering that the average subject's ErrP classification accuracy is only $73\%$. That being said, due to the simplistic testing environments, all algorithms eventually converge to optimal performance, making any increases in robustness harder to gleam. Moreover, the overall increase in aptitude remains undetermined as testing in more complex environments is needed. Lastly, it is unclear how this method would perform when no environmental reward function $r(s, a)$ is provided such that only the human reward function $r_{h}(s, a)$ could be utilized.

A simpler approach is proposed by \citet{akinola_accelerated_2020} focuses on modeling human action probabilities. This approach bears resemblance to TAMER+RL; however, it diverges by employing qualitative feedback instead of quantitative. The proposed approach models intrinsic feedback $h$ communicated via ErrPs by having $\hat{h}$ represent a probability distribution over the possible ``correct'' actions. Thus, the action with the highest predicted probability corresponds to the action the human is most likely to consider correct. Modeling of $\hat{h}$ is done by minimizing the binary cross-entropy loss for the samples drawn from a replay buffer. The policy $\pi_{h}$ of the agent is then greedily derived from $\hat{h}$ such that
\begin{equation}\label{eq:human_policy}
\pi_{h}(s) = \argmax{a}  \hat{h}(s,a).
\end{equation}
Once $\hat{h}$ is learned, any deep RL algorithm with access to a very simple sparse reward can be used for learning the policy $\pi_{RL}$ where the control sharing shaping methodology assists in influencing the policy. This works by having $\pi_{h}$ replace  $\pi_{RL}$ for a particular episode given some probability $\epsilon$ that decays linearly over time.
 
The empirical results from a simulated 3D robotic navigation environment are mixed due to the performance being dependent on the ErrP classification accuracy for each subject. Accuracies above 65\% show an general increase in both aptness and robustness over the proximal policy optimization (PPO) baseline \citep{schulman_ppo_2017}. When classification accuracies are less than $65\%$ the results become mixed while accuracies near $60\%$ perform the worst. Scalability is likely to be slightly increased as the proposed approach can utilize the reward function, given one is available. As per usual, the lack of testing environments leaves the overall increase in aptitude largely undetermined. Another potential caveat arises from the fact that the human policy and RL policy must be trained sequentially as no investigation into training the two policies in parallel is given.
  
 \citet{wang_maximizing_2020} build upon the work of \citet{akinola_accelerated_2020} by adding new components that aim to further increase the aptitude of the algorithm. In particular, uncertainty is introduced to inherently address the noisy intrinsic feedback decoding due to misclassification. Therefore, instead of only modeling $\hat{h}$ as the estimation of the probability of actions, the proposed algorithm uses Evidential Deep Learning \citep{sensoy_evidential_2018} to model $\hat{h}$ as a distribution over all such action probabilities. Meaning, the confidence of an action being considered ``correct'' can be measured.

Uncertainty enables active learning techniques and a purified replay buffer to be integrated into the approach. Active learning allows the agent to only query for intrinsic feedback when it is not confident in the action it took such that feedback is more informative. The purified replay buffer stores intrinsic feedback and state tuples only if the confidence of $\hat{h}(S, A)$ is greater than a pre-defined threshold $\epsilon$. This threshold helps to regulate the noise in feedback such that only consistent feedback is stored. Training batches are then sampled from both the purified and basic replay buffers. Further, $\pi_{RL}$ is further refined by using a LfD approach that helps the RL algorithm to reproduce good behaviors previously performed by either $\pi_{h}$ or $\pi_{RL}$.

Empirical results are gathered from same environment from \citet{akinola_accelerated_2020} and a new simulated robotic reaching task. First, ablation studies using simulated feedback verify the importance of each new component. Further results then show a large improvement in both aptness and robustness over original approach \citep{akinola_accelerated_2020} and the PPO baseline \citep{schulman_ppo_2017} for both simulated and real human feedback. Moreover, classification accuracies that fall between $55-60\%$ show increases in aptness and robustness as well. Scalability also shows potential for improvement due to active learning reducing the amount of feedback required. As before in \citet{akinola_accelerated_2020}, the remaining caveats in regards to the lack of testing environments and sequential training remain.

%%%%%%%%%%%%%%%%%%%%%%%%%%%%%%%%%%%%%%%%%%%%%%%%%%%%%%%%%%%%%%%%%%%%%%%%%%%%%%%%%%%%%%%%%%%%%%%%%%%%%%%%%%%%%%%%%%%%%%%%%%%%%%%%%%%%%%%

\section{Challenges and Future Directions}\label{sec:challenges}

Despite the promising nature of intrinsic feedback, it still remains a relatively nascent research area. Therefore, it is crucial to acknowledge and discuss some of the significant challenges that it faces. Throughout this section, we delve into the potential challenges and future directions of intrinsic feedback. When applicable, we relate intrinsic feedback challenges to that of its parent fields of interactive RL, LfF, and BCI. This is necessary as some of the challenges can interconnected and can be seen as ``bad habits'' that intrinsic feedback approaches have inherited.

%%%%%%%%%%%%%%%%%%%%%%%%%%%%%%%%%%%%%%%%%%%%%%%%%%%%%%%%%%%%%%%%%%%%%%%%%%%%%%%%%%%%%%%%%%%%%%%%%%%%%%%%%%%%%%%%%%%%%%%%%%%%%%%%%%%%%%%

\subsection{Viability Limitations}\label{sec:challenges_viability}
One of the prominent challenges for works that use intrinsic feedback is establishing their viability. We define ``viability'' as the endeavor to fulfil the motivation of effectively alleviating aptitude and alignment challenges associated with integrating human input. Establishing viability requires not only rigorous and in-depth experiments but also thorough analysis. While we have provided a high-level analysis of viability for both intrinsic feedback and LfF approaches, it remains relatively difficult to provide a more satisfactory analysis. This is largely because there are three gleaming issues that make the analysis of viability challenging. Each of these three issues is an example of a bad habit that the intrinsic feedback literature has inherited from its RL related parent fields. Shedding light on these issues is valuable as addressing them will pave the way for better experimental design and analysis, not only for intrinsic feedback works but LfF works as well.
 
First, is the lack of standardized and complex testing environments. Current intrinsic feedback and, more generally, LfF works employ inconsistent and rather basic testing environments. This can clearly be seen in Table~\ref{tab:irl_works} where differing LfF frameworks and intrinsic feedback works rarely use the same environment. Although some works might appear to use the same environment, they actually employ different variations or implementations of it (e.g., \citet{luo_drlpref_2018} and \citet{xu_accelerating_2021} use different Atari or Atari-like environments). This issue is compounded by the use of classical RL environments (e.g., mountain car or gridworld). Even in cases where more complex environments like Atari or basic robotics behavioral tasks (e.g., goal search) are employed, the investigation only focuses on one or two of these environments. In turn, more extensive experiments using a variety of complex environments will be required to further validate viability. 

While this is acceptable and expected for initial proof-of-concept approaches, modern day approaches aiming for practical applications will need to establish standardized and increasingly complex benchmark environments. This is in comparison to the wider RL community which maintains benchmarks that are progressively increasing in complexity. Initially, RL started with the same classical environments, but over time, the complexity has slowly increased to encompass more advanced robotic control tasks \citep{openai_hand_2020}, gaming \citep{brockman_gym_2016, openai_dota_2019, vinyals_alphastar_2019}, and simulated tasks within physic engines \citep{todorov_mujoco_2012}. Thus, it is crucial to address the question of which environment(s) should intrinsic feedback, and LfF or interactive RL in general, use as a standardized test bed. The easiest and most likely good starting point would be to simply follow the wider RL literature. Albeit, it is worth considering that intrinsic feedback might not be optimally useful in alleviating aptitude or alignment issues in certain environments (e.g., gaming). In such cases, more complex human-agent interactive environments, similar to applications often explored by BCI research \citep{arico_pbci_2018}, may be required. Yet, with little to no exploration into either of these directions, it is challenging to determine which holds the most promise.

Secondly, there is a notable lack of baseline comparisons against both RL and interactive RL approaches. In the realm of intrinsic feedback studies, it is common to compare results with at least one baseline RL approach. However, there is still a significant absence of comparisons with other LfF works. Due to this, and the aforementioned environment issues, it remains quite unclear what the current state-of-the-art approach is for either intrinsic feedback or LfF.  Moreover, the lack of baseline comparisons is not helped by the lack of reproducibility that has characterized LfF works and now intrinsic feedback works \citep{knox_interactively_2009, warnell_deeptamer_2018, macglashan_coach_2017, arumugam_dlcoach_2019, arakawa_dqntamer_2018, xu_accelerating_2021, akinola_accelerated_2020, wang_maximizing_2020}. While many of these works do at least provide pseudo-code, this still leaves the replication of results and further exploration of these approaches in an unnecessarily difficult state. This is in stark contrast to the broader RL field, where open-sourced code has become commonplace. 

Third, and more general to RL, is the lack of quantification measures for alignment improvements over baselines. This is crucial as human input aims to better align agents with human intentions. The biggest issue is that the assessment of alignment performance cannot rely solely on the performance measures, as is typically done for aptitude analysis. Oftentimes, a misaligned agent can still show an increase in performance measures even though it might not accomplish the task in the manner the human intended it to \citep{amodei_aisafty_2016, leike_alignment_2018}. Even more unclear is when the human finds the agent's performance sufficient which might differ from classical RL performance metrics (e.g., total return or completion percentage) altogether. Therefore, assessment of alignment alleviation often requires finer inspection of the agent's behaviors and more specified evaluation metrics that are often not thoroughly discussed or provided \citep{nahian_training_2021, leike_alignment_2018}.

%%%%%%%%%%%%%%%%%%%%%%%%%%%%%%%%%%%%%%%%%%%%%%%%%%%%%%%%%%%%%%%%%%%%%%%%%%%%%%%%%%%%%%%%%%%%%%%%%%%%%%%%%%%%%%%%%%%%%%%%%%%%%%%%%%%%%%%

\subsection{Decoding Non-stationary Brain Signals}\label{sec:challenges_non-stationary}
One of the defining properties of brain signals is that they follow a non-stationary distribution. This presents a rather difficult challenge for accurate decoding as the distribution of informative brain signal can shift, causing the performance of the classification model to degrade \citep{klonowski_eegquest_2009, shenoy_nonstation_2006, samek_erpnonstation_2015}. This non-stationarity is largely due to constant changes in the underlying noise that is combined with the brain signal of interest. As fully separating a brain signal from the surrounding biological and external noise is not currently possible, a brain signal's distribution is tightly coupled to the accompanying noise. Changes in cognitive and affective processes, biological artifacts, and external noise all play a role in influencing a signals distribution \citep{samek_erpnonstation_2015, millan_onlinebci_2004}. Thus, the sensitivity to noise can shift a signal's distributions even only after a short period of time (i.e., from trial to trial) \citep{blankertz_erpvary_2011}. The non-stationarity of brain signal distributions is only further pronounced when comparing distributions between humans as individual characteristics play a role in influencing distributions \citep{gu_brainstruct_2014, hoffmann_errpage_2012}.

A very clear example of this non-stationary distribution is found in ErrPs. In part, ErrP variations have been influenced by different cognitive processes that are elicited due to different tasks \citep{abu_errpinvariance_2017, ehrlich_errpfeasibility_2019, iturrate_errptask_2013}, changes in the complexity of a task \citep{kakkos_errpcomplex_2020}, or changes in the way a task interacts with \citep{kim_transfer_2013, kim_handlingtransfer_2016, poole_errp_2022}. For example, if an error occurs either while performing a task or while observing the same task being performed, two slightly different ErrPs are elicited \citep{kim_transfer_2013, kim_handlingtransfer_2016, abu_errpinvariance_2017, poole_errp_2022}. Further potential variation can arise from the frequency of errors. For instance, frequent elicitation has been shown to alter the grand signal average of an ErrP, indicating a likely deterioration in classification as the underlying ErrP signal has been altered \citep{abu_errpinvariance_2017, ferrez_errp_2008}. Overcoming this challenge would lead to an improvement in the classification accuracy such that intrinsic feedback would be more accurately conveyed to agents \citep{kim_handlingtransfer_2016, iturrate_robot_2010, luo_drlpref_2018, xu_accelerating_2021}. However, it is vastly unclear to what degree agent performance scales with classification accuracy as some works report successful learning with classification accuracies as low as $55\%$ \citep{wang_maximizing_2020}.

Nevertheless, non-stationarity has been a long ongoing challenge faced by practically all BCI applications, making it an active area of research today. For example, transfer learning has been adopted to deal with these distribution shifts by using data from a source domain to either find latent features or learn parameters that capture the invariant underlying signal structure such that they can be transferred to some desired target domain \citep{kim_transfer_2013, kim_handlingtransfer_2016, abu_errpinvariance_2017, ehrlich_errpfeasibility_2019, iturrate_errptask_2013,poole_errp_2022}. The source domain serves as the initial signal distribution and the target domain serves as the shifted signal distribution. Alternatively, adaptive classification where classifiers try to continuously learn from newly collected data in order to account for the shifts in signal distribution as they occur has been applied as well \citep{millan_onlinebci_2004, shenoy_nonstation_2006}. 

%%%%%%%%%%%%%%%%%%%%%%%%%%%%%%%%%%%%%%%%%%%%%%%%%%%%%%%%%%%%%%%%%%%%%%%%%%%%%%%%%%%%%%%%%%%%%%%%%%%%%%%%%%%%%%%%%%%%%%%%%%%%%%%%%%%%%%%

\subsection{Sample Efficiency of Feedback}\label{sec:challenges_sample_inefficiency}
Collecting human feedback is often expensive endeavor, resulting in limited availability of feedback. Under the assumption that as the complexity of the environment increases, more feedback will be required for learning to occur; the challenge becomes efficiently learning from limited feedback even as complexity scales up. Although most LfF approaches seem to possess potential for increasing scalability \citep{knox_thesis_2012, warnell_deeptamer_2018, christiano_preferences_2017, luo_drlpref_2018, xu_accelerating_2021, wang_maximizing_2020}, the question remains: to what extent and how much feedback is required when introducing more complex environments? In other words, to what degree are these approaches actually scalable? These questions are asked not only regarding training an agent with raw feedback but also when a model of feedback is required to be built. In turn, the question becomes how much feedback is required to build a sufficient feedback model. These answers remain relatively unknown as only further experimental investigations can really begin to shed any light on these questions.

Intrinsic feedback in particular introduces additional considerations for sample efficiency. First, intrinsic feedback requires pre-training or simultaneous training of a classification model to extract the informative brain signal that will act as feedback. If the classification performance is relatively poor, the classifier is likely to present larger amounts of noisy feedback, reducing sample efficiency \citep{xu_accelerating_2021, luo_drlpref_2018, wang_maximizing_2020}. Thus, providing sufficient training examples to build a good classifier needs to be considered as well. Second, collecting intrinsic feedback can be far more expensive than collecting explicit feedback. The setup of EEG equipment is often time consuming and requires frequent maintenance during an elongated session. Consequently, this limitation can hinder both the quantity and rate at which feedback can be collected. Despite these challenges, there are claims that intrinsic feedback can be captured at a faster rate compared to explicit feedback \citep{luo_drlpref_2018}, which adds an additional layer of uncertainty to the state of sample efficiency for intrinsic feedback.

%%%%%%%%%%%%%%%%%%%%%%%%%%%%%%%%%%%%%%%%%%%%%%%%%%%%%%%%%%%%%%%%%%%%%%%%%%%%%%%%%%%%%%%%%%%%%%%%%%%%%%%%%%%%%%%%%%%%%%%%%%%%%%%%%%%%%%%

\subsection{Feedback Credit Assignment}\label{sec:challenges_credit_assignment}
LfF approaches often present a reduced credit assignment problem as human feedback is still delayed from the original contributing events \citep{arzate_irlsurvey_2020}. Albeit, the reduction in the difficulty of the credit assignment problem is relegated to how long the human decides to delay feedback. Thus, prior LfF works have attempted to address this reduced credit assignment problem although it still remains unclear which methods produce the best results or how crucial solving the problem actually is. For instance, \citet{knox_interactively_2009} attempt to address the credit assignment by estimating the delay between feedback and the corresponding events using a PDF \citep{knox_interactively_2009, knox_thesis_2012}. However, \citet{warnell_deeptamer_2018} show that the selection of a PDF matters as changing the PDF can produce widely varying results. This entails that if the assumed credit assignment PDF does not match the human's delay distribution, performance can be greatly affected. Likewise, \citet{macglashan_coach_2017} propose using eligibility traces with reward aggregation. Despite the promising nature of both approaches, it remains uncertain how either approach compares to one another.

Furthermore, the impact of intrinsic feedback on the feedback credit assignment problem remains unclear. For instance,  ErrPs, and ERPs in general, tend to have a relatively consistent onset and duration \citep{luck_erpbook_2014, falkenstein_ern_2000}. This implies that a specific ErrP can be expected to occur approximately 200 ms after any erroneous event \citep{falkenstein_ern_2000}. Moreover, a sliding window lasting up to 1 second long is needed to capture the full ErrP signal where even shorter windows have been used \citep{chavarriage_errpbci_2014, kim_intrinsic_2017}. Theoretically, intrinsic feedback is likely to further simplify the credit assignment problem as feedback elicitation and duration remain relatively static. Keep in mind that all forms of feedback require the accumulation of evidence (i.e., the buildup of sufficient evidence that something is actually an error) before feedback can be provided \citep{batzianoulis_customizing_2021}. Intrinsic feedback is not magically freed from this natural constraint. Thus, long-term feedback signals that accumulate over time (e.g., difficult to discern errors or affective states) might present more difficult credit assignment problems. 

An additional caveat of intrinsic feedback that remains relatively unexplored is the problem of future credit assignment. This problem entails the mapping of anticipatory feedback to future events that the human foresees. For instance, a human might provide negative feedback right before the agent collides with an incoming obstacle. Although this problem was originally recognized by \citet{thomaz_reinforcement_2006} in the explicit feedback literature, it has also been of interest to BCI researchers as well. In particular, BCI researchers have been interested in detecting and utilizing ErrPs that are elicited in response to foreseen erroneous events \citep{iwane_inferring_2019, batzianoulis_customizing_2021}. Lastly, it is worth considering whether this problem actually meets the definition of feedback or if it is better suited as being interpreted as a form of guidance. Regardless, it remains unclear how to handle this type of advice but intrinsically occurring brain signals present a well sited use-case for additional investigations into this problem.

%%%%%%%%%%%%%%%%%%%%%%%%%%%%%%%%%%%%%%%%%%%%%%%%%%%%%%%%%%%%%%%%%%%%%%%%%%%%%%%%%%%%%%%%%%%%%%%%%%%%%%%%%%%%%%%%%%%%%%%%%%%%%%%%%%%%%%%

\subsection{Lack of Positive Feedback}\label{sec:challenges_multi_feedback}
Recall that most of the intrinsic feedback works covered have dealt with only detecting and providing a negative feedback signal. Conversely, prior explicit quantitative feedback works that we have covered have largely used both positive and negative feedback to provide more information to the agent \citep{knox_thesis_2012, warnell_deeptamer_2018, macglashan_coach_2017, arumugam_dlcoach_2019}. This brings into question the importance of utilizing a positive feedback signal. There is an argument to be made for excluding positive feedback to prevent alignment challenges (e.g., reward hacking or positive circuits) \citep{knox_thesis_2012, knox_combining_2010, knox_reinforcement_2012}. At the same time, removing positive feedback limits the human's degrees of freedom as human trainers tend to utilize both positive and negative feedback in different ways \citep{macglashan_coach_2017, arumugam_dlcoach_2019}. Particularly, human feedback tends to be positively biased \citep{thomaz_posneg_2007, knox_thesis_2012, knox_combining_2010, knox_reinforcement_2012, arumugam_dlcoach_2019}. Thus, excluding positive feedback may reduce the amount of information available to the agent making complex tasks with more abstract goals harder to learn, although this requires further investigation. 

As it stands, intrinsic feedback works have largely been focused on utilizing ErrPs as negative qualitative feedback where positive feedback or no feedback is assumed by default \citep{xu_accelerating_2021, akinola_accelerated_2020, wang_maximizing_2020, iturrate_errprl_2015, ehrlich_errpfeasibility_2019}. In doing so, these intrinsic feedback works have still been able to learn. Whether or not learning could be improved using an intrinsic signal that can be directly interpreted as positive feedback (i.e., not assumed by default) remains to be seen. As briefly discussed in the \nameref{sec:errp_feedback} section, there are potential options for signals that can be interpreted as positive feedback. The caveat of multi-signal feedback is that it inherently introduces multi-class classification which is likely to introduce even more difficulties in training a well performing classifier \citep{wirth_4way_2020}. It also remains to be seen if the necessity of positive feedback applies only to quantitative, or to qualitative feedback as well. Recall that qualitative works can model a latent numerical representation that can inherently take on either positive or negative values \citep{xu_accelerating_2021, christiano_preferences_2017}.

%%%%%%%%%%%%%%%%%%%%%%%%%%%%%%%%%%%%%%%%%%%%%%%%%%%%%%%%%%%%%%%%%%%%%%%%%%%%%%%%%%%%%%%%%%%%%%%%%%%%%%%%%%%

\subsection{Multi-Channel Communication}\label{sec:challenges_multi_guidance}
As was covered in the \nameref{sec:irl_motivation} section, increasing the number of channels in which a human can communicate the agent is crucial for alleviating both aptitude and alignment issues. Intrinsic feedback presents an interesting opportunity that allows for new LfF setups where various types of human input are conveyed in parallel. Thus, deeper investigation into how to utilize multiple types of human input and feedback mediums for learning will be required.

As inspiration, we can first look at a handful of initial works that have already begun to investigate this direction without intrinsic feedback. As mentioned before, \citet{ibarz_rewardlearning_2018} look at combing explicit feedback with demonstrations in a sequential manner. This is done by first pre-training the policy using behavioral cloning and then fine-tuning using preference-based feedback. Alternately, \citet{guan_widening_2021} look at combining explicit feedback and guidance. This is done by having the human provide feedback while also highlighting objects in an environment that are important or relate to the given feedback. Likewise, \citet{zhang_head_2019, zhang_agil_2018} take an intriguing approach through a combination of implied guidance and demonstrations. Implied guidance communicated via human gaze is collected in parallel with demonstrations allowing for two inter-connected sources of information to be related. Here, gaze implies information about what the agent should be attending to while learning from the human demonstrations.

Similar ideas could be used in conjunction with intrinsic feedback. Take for instance the parallel combination of intrinsic feedback and demonstrations (\figref{fig:iirl-int-all}). Doing so would allow the agent to have access to feedback regarding which actions or behaviors the human thinks they performed incorrectly. In turn, this could allow LfD algorithms to learn using noisy or imperfect demonstrations. Alternatively, \citet{batzianoulis_customizing_2021} explore a different method for combining both intrinsic feedback with demonstration. This is done by using intrinsic feedback where the probability of detecting an ErrP in the agent's trajectory is used to weigh the inverse RL loss. Additionally, the combination of implicit guidance (e.g., gaze) and intrinsic feedback could help alleviate the problem of whether the intrinsically occurring signals are actually being elicited due to the task or some unrelated distractor in the human's line of sight. This would also allow for the intrinsic feedback to be mapped to the location in the task which is eliciting the feedback. Further exploration into how to combine intrinsic feedback with other forms of human input is likely to be a promising future direction of research. 

Lastly, an investigation into the benefits and disadvantages of intrinsic feedback in comparison to the other feedback mediums (i.e., explicit and implicit) could prove useful for determining how or when to use each medium. For example, \citet{luo_drlpref_2018} suggest that more intrinsic feedback can be captured in less amount of time than that of explicit feedback while \citet{xu_accelerating_2021} claim intrinsic feedback can be captured more accurately when extremely fast feedback rates are required. More in-depth and thorough comparisons are needed to ensure these trends can be replicated in diverse settings. In particular, thoroughly understanding what effects time pressure might have on the intrinsic feedback classifier is needed.

%%%%%%%%%%%%%%%%%%%%%%%%%%%%%%%%%%%%%%%%%%%%%%%%%%%%%%%%%%%%%%%%%%%%%%%%%%%%%%%%%%%%%%%%%%%%%%%%%%%%%%%%%%%%%%%%%%%%%%%%%%%%%%%%%%%%%%%

\section{Conclusion}
This paper presented a tutorial-style review on intrinsic feedback and its foundational concepts, offering insights into the intersection of RL and BCI. In doing so, we hope to better empower researchers who wish to use intrinsic feedback by providing them with knowledge about how feedback has been utilized and the viability of these approaches. In particular, we have focused on interactive RL and LfF as intrinsically occurring biological signals that can be more readily interpreted as a new medium of feedback which we denote as intrinsic feedback. To do so, we established the groundwork for the field of interactive RL to provide an understanding of the field and how feedback fits into it. Moreover, we highlight foundational LfF approaches and their viability to provide an understanding of how feedback has classically been utilized. We further discuss intrinsic feedback as a concept and provide a comprehensive review of current intrinsic feedback works, including their viability and connections with existing LfF approaches. Finally, we provide insights into the challenges and future research directions for intrinsic feedback where we include relevant challenges that have been inherited from its parent fields of RL and BCI.

%%%%%%%%%%%%%%%%%%%%%%%%%%%%%%%%%%%%%%%%%%%%%%%%%%%%%%%%%%%%%%%%%%%%%%%%%%%%%%%%%%%%%%%%%%%%%%%%%%%%%%%%%%%%%%%%%%%%%%%%%%%%%%%%%%%%%%%
\section*{Author Contributions}
\textbf{Benjamin Poole}: Conceptualization, Writing - Original Draft, Writing - Reviewing and Editing, Visualization \textbf{Minwoo Lee}: Writing - Reviewing and Editing, Supervision 

\section*{Declaration of Interests}
The authors declare that they have no known competing financial interests or personal relationships that could have appeared to influence the work reported in this paper.

\section*{Acknowledgements}
The authors would like to especially thank Jos\'e del R. Mill\'an, Bradley Knox, and Peter Stone for taking the time to provide valuable feedback on early iterations of the paper.

%%%%%%%%%%%%%%%%%%%%%%%%%%%%%%%%%%%%%%%%%%%%%%%%%%%%%%%%%%%%%%%%%%%%%%%%%%%%%%%%%%%%%%%%%%%%%%%%%%%%%%%%%%%%%%%%%%%%%%%%%%%%%%%%%%%%%%%
\bibliographystyle{plainnat}
\bibliography{manuscript} 

\begin{thebibliography}{133}
\providecommand{\natexlab}[1]{#1}
\providecommand{\url}[1]{\texttt{#1}}
\expandafter\ifx\csname urlstyle\endcsname\relax
  \providecommand{\doi}[1]{doi: #1}\else
  \providecommand{\doi}{doi: \begingroup \urlstyle{rm}\Url}\fi

\bibitem[Abu-Alqumsan et~al.(2017)Abu-Alqumsan, Kapeller, Hinterm{\"u}ller,
  Guger, and Peer]{abu_errpinvariance_2017}
Mohammad Abu-Alqumsan, Christoph Kapeller, Christoph Hinterm{\"u}ller,
  Christoph Guger, and Angelika Peer.
\newblock Invariance and variability in interaction error-related potentials
  and their consequences for classification.
\newblock 14\penalty0 (6):\penalty0 066015, November 2017.
\newblock ISSN 1741-2552.
\newblock \doi{10.1088/1741-2552/aa8416}.
\newblock URL \url{https://doi.org/10.1088\%2F1741-2552\%2Faa8416}.

\bibitem[Akinola et~al.(2020)Akinola, Wang, Shi, He, Lapborisuth, Xu,
  Watkins-Valls, Sajda, and Allen]{akinola_accelerated_2020}
Iretiayo Akinola, Zizhao Wang, Junyao Shi, Xiaomin He, Pawan Lapborisuth,
  Jingxi Xu, David Watkins-Valls, Paul Sajda, and Peter Allen.
\newblock Accelerated robot learning via human brain signals.
\newblock In \emph{IEEE International Conference on Robotics and Automation},
  pages 3799--3805. IEEE, 2020.
\newblock \doi{10.1109/ICRA40945.2020.9196566}.

\bibitem[Alnafjan et~al.(2017)Alnafjan, Hosny, Al-Ohali, and
  Al-Wabil]{alnafjan_emotionsurvey_2017}
Abeer Alnafjan, Manar Hosny, Yousef Al-Ohali, and Areej Al-Wabil.
\newblock Review and {Classification} of {Emotion} {Recognition} {Based} on
  {EEG} {Brain}-{Computer} {Interface} {System} {Research}: {A} {Systematic}
  {Review}.
\newblock 7:\penalty0 1239, November 2017.
\newblock \doi{10.3390/app7121239}.

\bibitem[Amodei et~al.(2016)Amodei, Olah, Steinhardt, Christiano, Schulman, and
  Man{\'e}]{amodei_aisafty_2016}
Dario Amodei, Chris Olah, Jacob Steinhardt, Paul Christiano, John Schulman, and
  Dan Man{\'e}.
\newblock Concrete {Problems} in {AI} {Safety}, July 2016.

\bibitem[Andrychowicz et~al.(2020)Andrychowicz, Baker, Chociej, J{\`o}zefowicz,
  McGrew, Pachocki, Petron, Plappert, Powell, Ray, Schneider, Sidor, Tobin,
  Welinder, Weng, and Zaremba]{openai_hand_2020}
OpenAI:~Marcin Andrychowicz, Bowen Baker, Maciek Chociej, Rafal J{\`o}zefowicz,
  Bob McGrew, Jakub Pachocki, Arthur Petron, Matthias Plappert, Glenn Powell,
  Alex Ray, Jonas Schneider, Szymon Sidor, Josh Tobin, Peter Welinder, Lilian
  Weng, and Wojciech Zaremba.
\newblock Learning dexterous in-hand manipulation.
\newblock 39\penalty0 (1):\penalty0 3--20, 2020.
\newblock \doi{10.1177/0278364919887447}.
\newblock URL \url{https://doi.org/10.1177/0278364919887447}.

\bibitem[Arakawa et~al.(2018)Arakawa, Kobayashi, Unno, Tsuboi, and
  Maeda]{arakawa_dqntamer_2018}
Riku Arakawa, Sosuke Kobayashi, Yuya Unno, Yuta Tsuboi, and Shin{-}ichi Maeda.
\newblock {DQN-TAMER:} human-in-the-loop reinforcement learning with
  intractable feedback.
\newblock abs/1810.11748, 2018.

\bibitem[Aric{\`{o}} et~al.(2018)Aric{\`{o}}, Borghini, Flumeri, Sciaraffa, and
  Babiloni]{arico_pbci_2018}
Pietro Aric{\`{o}}, Gianluca Borghini, Gianluca~Di Flumeri, Nicolina Sciaraffa,
  and Fabio Babiloni.
\newblock Passive {BCI} beyond the lab: current trends and future directions.
\newblock 39\penalty0 (8):\penalty0 08TR02, August 2018.
\newblock ISSN 0967-3334.
\newblock \doi{10.1088/1361-6579/aad57e}.
\newblock URL \url{https://doi.org/10.1088\%2F1361-6579\%2Faad57e}.

\bibitem[Arumugam et~al.(2019)Arumugam, Lee, Saskin, and
  Littman]{arumugam_dlcoach_2019}
Dilip Arumugam, Jun~Ki Lee, Sophie Saskin, and Michael~L. Littman.
\newblock Deep {Reinforcement} {Learning} from {Policy}-{Dependent} {Human}
  {Feedback}, February 2019.

\bibitem[Arzate~Cruz and Igarashi(2020)]{arzate_irlsurvey_2020}
Christian Arzate~Cruz and Takeo Igarashi.
\newblock A {Survey} on {Interactive} {Reinforcement} {Learning}: {Design}
  {Principles} and {Open} {Challenges}.
\newblock In \emph{Proceedings of the 2020 {ACM} {Designing} {Interactive}
  {Systems} {Conference}}, {DIS} '20, pages 1195--1209. Association for
  Computing Machinery, July 2020.
\newblock ISBN 978-1-4503-6974-9.
\newblock \doi{10.1145/3357236.3395525}.
\newblock URL \url{https://doi.org/10.1145/3357236.3395525}.

\bibitem[Azizzadenesheli et~al.(2018)Azizzadenesheli, Brunskill, and
  Anandkumar]{kamyar_bdqn_2018}
Kamyar Azizzadenesheli, Emma Brunskill, and Animashree Anandkumar.
\newblock Efficient exploration through bayesian deep q-networks.
\newblock In \emph{ITA}, pages 1--9. IEEE, 2018.
\newblock \doi{10.1109/ITA.2018.8503252}.

\bibitem[Batzianoulis et~al.(2021)Batzianoulis, Iwane, Wei, Correia,
  Chavarriaga, Millan, and Billard]{batzianoulis_customizing_2021}
Iason Batzianoulis, Fumiaki Iwane, Shupeng Wei, Carolina Gaspar Pinto~Ramos
  Correia, Ricardo Chavarriaga, Jose del~R. Millan, and Aude Billard.
\newblock Customizing skills for assistive robotic manipulators, an inverse
  reinforcement learning approach with error-related potentials.
\newblock 4\penalty0 (1):\penalty0 1--14, December 2021.
\newblock ISSN 2399-3642.
\newblock \doi{10.1038/s42003-021-02891-8}.
\newblock URL \url{https://www.nature.com/articles/s42003-021-02891-8}.

\bibitem[Blankertz et~al.(2011)Blankertz, Lemm, Treder, Haufe, and
  M{\"u}ller]{blankertz_erpvary_2011}
Benjamin Blankertz, Steven Lemm, Matthias Treder, Stefan Haufe, and
  Klaus-Robert M{\"u}ller.
\newblock Single-trial analysis and classification of {ERP} components - {A}
  tutorial.
\newblock 56\penalty0 (2):\penalty0 814--825, May 2011.
\newblock ISSN 1053-8119.
\newblock \doi{10.1016/j.neuroimage.2010.06.048}.
\newblock URL
  \url{http://www.sciencedirect.com/science/article/pii/S1053811910009067}.

\bibitem[Bostrom(2020)]{bostrom_ethical_2020}
Nick Bostrom.
\newblock Ethical issues in advanced artificial intelligence.
\newblock \emph{Machine Ethics and Robot Ethics}, pages 69--75, 2020.

\bibitem[Brockman et~al.(2016)Brockman, Cheung, Pettersson, Schneider,
  Schulman, Tang, and Zaremba]{brockman_gym_2016}
Greg Brockman, Vicki Cheung, Ludwig Pettersson, Jonas Schneider, John Schulman,
  Jie Tang, and Wojciech Zaremba.
\newblock {OpenAI} {Gym}, 2016.

\bibitem[Celemin and Ruiz-del Solar(2019)]{celemin_interactive_2019}
Carlos Celemin and Javier Ruiz-del Solar.
\newblock An interactive framework for learning continuous actions policies
  based on corrective feedback.
\newblock 95\penalty0 (1):\penalty0 77--97, 2019.
\newblock ISSN 1573-0409.
\newblock \doi{10.1007/s10846-018-0839-z}.
\newblock URL \url{https://doi.org/10.1007/s10846-018-0839-z}.

\bibitem[Chakraborti et~al.(2017)Chakraborti, Sreedharan, Kulkarni, and
  Kambhampati]{chakraborti_hitlbci_2017}
Tathagata Chakraborti, Sarath Sreedharan, Anagha Kulkarni, and Subbarao
  Kambhampati.
\newblock Alternative {Modes} of {Interaction} in {Proximal}
  {Human}-in-the-{Loop} {Operation} of {Robots}, March 2017.

\bibitem[Chavarriaga and Millan(2010)]{chavarriaga_errplearning_2010}
Ricardo Chavarriaga and Jose del~R. Millan.
\newblock Learning {From} {EEG} {Error}-{Related} {Potentials} in {Noninvasive}
  {Brain}-{Computer} {Interfaces}.
\newblock 18\penalty0 (4):\penalty0 381--388, August 2010.
\newblock ISSN 1534-4320.
\newblock \doi{10.1109/TNSRE.2010.2053387}.

\bibitem[Chavarriaga et~al.(2014)Chavarriaga, Sobolewski, and
  Millan]{chavarriage_errpbci_2014}
Ricardo Chavarriaga, Aleksander Sobolewski, and Jose del~R. Millan.
\newblock Errare machinale est: the use of error-related potentials in
  brain-machine interfaces.
\newblock 8:\penalty0 208, July 2014.
\newblock ISSN 1662-4548.
\newblock \doi{10.3389/fnins.2014.00208}.
\newblock URL \url{https://www.ncbi.nlm.nih.gov/pmc/articles/PMC4106211/}.

\bibitem[Chavarriaga et~al.(2016)Chavarriaga, Iturrate, and
  Millan]{chavarriaga_robust_2016}
Ricardo Chavarriaga, I{\~n}aki Iturrate, and Jose del~R. Millan.
\newblock Robust, accurate spelling based on error-related potentials.
\newblock In \emph{6th Int. BCI Meeting}, May 2016.
\newblock \doi{10.3217/978-3-85125-467-9-15}.
\newblock URL \url{https://dx.doi.org/10.3217/978-3-85125-467-9-15}.

\bibitem[Christiano et~al.(2017)Christiano, Leike, Brown, Martic, Legg, and
  Amodei]{christiano_preferences_2017}
Paul~F Christiano, Jan Leike, Tom Brown, Miljan Martic, Shane Legg, and Dario
  Amodei.
\newblock Deep reinforcement learning from human preferences.
\newblock 30, 2017.

\bibitem[Craik et~al.(2019)Craik, He, and
  Contreras-Vidal]{craik_bcisurvey_2019}
Alexander Craik, Yongtian He, and Jose~L. Contreras-Vidal.
\newblock Deep learning for electroencephalogram ({EEG}) classification tasks:
  a review.
\newblock 16\penalty0 (3):\penalty0 031001, April 2019.
\newblock ISSN 1741-2552.
\newblock \doi{10.1088/1741-2552/ab0ab5}.
\newblock URL \url{https://doi.org/10.1088\%2F1741-2552\%2Fab0ab5}.

\bibitem[Cruz and Igarashi(2021)]{cruz_interactive_2021}
Christian~Arzate Cruz and Takeo Igarashi.
\newblock Interactive {Explanations}: {Diagnosis} and {Repair} of
  {Reinforcement} {Learning} {Based} {Agent} {Behaviors}.
\newblock In \emph{2021 {IEEE} {Conference} on {Games} ({CoG})}, pages 01--08,
  August 2021.
\newblock \doi{10.1109/CoG52621.2021.9618999}.

\bibitem[Cruz et~al.(2016)Cruz, Parisi, Twiefel, and
  Wermter]{cruz_multimodal_2016}
Francisco Cruz, German~I. Parisi, Johannes Twiefel, and Stefan Wermter.
\newblock Multi-modal integration of dynamic audiovisual patterns for an
  interactive reinforcement learning scenario.
\newblock pages 759--766, October 2016.
\newblock \doi{10.1109/IROS.2016.7759137}.
\newblock ISSN: 2153-0866.

\bibitem[Dabas et~al.(2020)Dabas, Saxena, Nordlund, and
  Ahamed]{dabas_bciappsurvey_2020}
S.~Dabas, P.~Saxena, N.~Nordlund, and S.~I. Ahamed.
\newblock A step closer to becoming symbiotic with {AI} through {EEG}: A review
  of recent {BCI} technology.
\newblock In \emph{{IEEE} 44th Annual Computers, Software, and Applications
  Conference}, {COMPSAC} '20, pages 361--368. Institute of Electrical and
  Electronics Engineers, 2020.
\newblock \doi{10.1109/COMPSAC48688.2020.0-220}.

\bibitem[Dulac-Arnold et~al.(2020)Dulac-Arnold, Levine, Mankowitz, Li,
  Paduraru, Gowal, and Hester]{dulacarnold_challenges_2020}
Gabriel Dulac-Arnold, Nir Levine, Daniel~J. Mankowitz, Jerry Li, Cosmin
  Paduraru, Sven Gowal, and Todd Hester.
\newblock An empirical investigation of the challenges of real-world
  reinforcement learning, March 2020.

\bibitem[Ehrlich and Cheng(2018)]{ehrlich_errpcoadapt_2018}
Stefan~K. Ehrlich and Gordon Cheng.
\newblock Human-agent co-adaptation using error-related potentials.
\newblock 15\penalty0 (6):\penalty0 066014, September 2018.
\newblock ISSN 1741-2552.
\newblock \doi{10.1088/1741-2552/aae069}.
\newblock URL \url{https://doi.org/10.1088\%2F1741-2552\%2Faae069}.

\bibitem[Ehrlich and Cheng(2019)]{ehrlich_errpfeasibility_2019}
Stefan~K. Ehrlich and Gordon Cheng.
\newblock A {Feasibility} {Study} for {Validating} {Robot} {Actions} {Using}
  {EEG}-{Based} {Error}-{Related} {Potentials}.
\newblock 11\penalty0 (2):\penalty0 271--283, April 2019.
\newblock ISSN 1875-4805.
\newblock \doi{10.1007/s12369-018-0501-8}.
\newblock URL \url{https://doi.org/10.1007/s12369-018-0501-8}.

\bibitem[Falkenstein et~al.(2000)Falkenstein, Hoormann, Christ, and
  Hohnsbein]{falkenstein_ern_2000}
Michael Falkenstein, Jorg Hoormann, Stefan Christ, and Joachim Hohnsbein.
\newblock {ERP} components on reaction errors and their functional
  significance: a tutorial.
\newblock 51\penalty0 (2):\penalty0 87--107, January 2000.
\newblock ISSN 0301-0511.
\newblock \doi{10.1016/S0301-0511(99)00031-9}.
\newblock URL
  \url{http://www.sciencedirect.com/science/article/pii/S0301051199000319}.

\bibitem[Farwell and Donchin(1988)]{farwell_p3_1988}
Lawrence~A. Farwell and Emanuel Donchin.
\newblock Talking off the top of your head: toward a mental prosthesis
  utilizing event-related brain potentials.
\newblock 70\penalty0 (6):\penalty0 510--523, December 1988.
\newblock ISSN 0013-4694.
\newblock \doi{10.1016/0013-4694(88)90149-6}.
\newblock URL
  \url{http://www.sciencedirect.com/science/article/pii/0013469488901496}.

\bibitem[Ferrez and Millan(2005)]{ferrez_errpwrong_2005}
Pierre~W. Ferrez and Jose del~R. Millan.
\newblock You {Are} {Wrong}! - {Automatic} {Detection} of {Interaction}
  {Errors} from {Brain} {Waves}.
\newblock In \emph{In Proceedings of the International Joint Conferences on
  Artificial Intelligence}, IJCAI '05, pages 1413--1418. Morgan Kaufmann
  Publishers Inc., 2005.

\bibitem[Ferrez and Millan(2008)]{ferrez_errp_2008}
Pierre~W. Ferrez and Jose del~R. Millan.
\newblock Error-{Related} {EEG} {Potentials} {Generated} {During} {Simulated}
  {Brain}-{Computer} {Interaction}.
\newblock 55\penalty0 (3):\penalty0 923--929, March 2008.
\newblock ISSN 0018-9294.
\newblock \doi{10.1109/TBME.2007.908083}.

\bibitem[Gabriel(2020)]{gabriel_artificial_2020}
Iason Gabriel.
\newblock Artificial {Intelligence}, {Values}, and {Alignment}.
\newblock 30\penalty0 (3):\penalty0 411--437, September 2020.
\newblock ISSN 1572-8641.
\newblock \doi{10.1007/s11023-020-09539-2}.
\newblock URL \url{https://doi.org/10.1007/s11023-020-09539-2}.

\bibitem[Goodfellow et~al.(2016)Goodfellow, Bengio, and
  Courville]{goodfellow_dl_2016}
Ian Goodfellow, Yoshua Bengio, and Aaron Courville.
\newblock \emph{Deep {Learning}}.
\newblock MIT Press, November 2016.
\newblock ISBN 978-0-262-33737-3.

\bibitem[Griffith et~al.(2013)Griffith, Subramanian, Scholz, Isbell, and
  Thomaz]{griffith_policy_2013}
Shane Griffith, Kaushik Subramanian, Jonathan Scholz, Charles~L. Isbell, and
  Andrea~L. Thomaz.
\newblock Policy {Shaping}: {Integrating} {Human} {Feedback} with
  {Reinforcement} {Learning}.
\newblock In C.~J.~C. Burges, L.~Bottou, M.~Welling, Z.~Ghahramani, and K.~Q.
  Weinberger, editors, \emph{Advances in {Neural} {Information} {Processing}
  {Systems}}, pages 2625--2633. Curran Associates, Inc., 2013.

\bibitem[Gu and Kanai(2014)]{gu_brainstruct_2014}
Jenny Gu and Ryota Kanai.
\newblock What contributes to individual differences in brain structure?
\newblock 8:\penalty0 262, 2014.
\newblock ISSN 1662-5161.
\newblock \doi{10.3389/fnhum.2014.00262}.
\newblock URL
  \url{https://www.frontiersin.org/articles/10.3389/fnhum.2014.00262/full}.

\bibitem[Gu et~al.(2021)Gu, Cao, Jolfaei, Xu, Wu, Jung, and
  Lin]{gu_bcisurvey_2020}
Xiaotong Gu, Zehong Cao, Alireza Jolfaei, Peng Xu, Dongrui Wu, Tzyy-Ping Jung,
  and Chin-Teng Lin.
\newblock Eeg-based brain-computer interfaces (bcis): A survey of recent
  studies on signal sensing technologies and computational intelligence
  approaches and their applications, 2021.
\newblock ISSN 1545-5963.
\newblock URL \url{https://doi.org/10.1109/TCBB.2021.3052811}.

\bibitem[Guan et~al.(2021)Guan, Verma, Guo, Zhang, and
  Kambhampati]{guan_widening_2021}
Lin Guan, Mudit Verma, Suna~(Sihang) Guo, Ruohan Zhang, and Subbarao
  Kambhampati.
\newblock Widening the pipeline in human-guided reinforcement learning with
  explanation and context-aware data augmentation.
\newblock In \emph{Advances in Neural Information Processing Systems},
  volume~34, pages 21885--21897. Curran Associates, Inc., 2021.
\newblock URL
  \url{https://proceedings.neurips.cc/paper/2021/hash/b6f8dc086b2d60c5856e4ff517060392-Abstract.html}.

\bibitem[Gui et~al.(2019)Gui, Ruiz-Blondet, Laszlo, and
  Jin]{gui_bcisecuritysurvey_2019}
Qiong Gui, Maria~V. Ruiz-Blondet, Sarah Laszlo, and Zhanpeng Jin.
\newblock A {Survey} on {Brain} {Biometrics}.
\newblock 51\penalty0 (6):\penalty0 112:1--112:38, February 2019.
\newblock ISSN 0360-0300.
\newblock \doi{10.1145/3230632}.
\newblock URL \url{https://doi.org/10.1145/3230632}.

\bibitem[Haarnoja et~al.(2017)Haarnoja, Tang, Abbeel, and
  Levine]{haarnoja_soft_2017}
Tuomas Haarnoja, Haoran Tang, Pieter Abbeel, and Sergey Levine.
\newblock Reinforcement learning with deep energy-based policies.
\newblock In \emph{International Conference on Machine Learning}, pages
  1352--1361. PMLR, 2017.

\bibitem[Hendrycks et~al.(2021)Hendrycks, Mazeika, Zou, Patel, Zhu, Navarro,
  Song, Li, and Steinhardt]{hendrycks_what_2021}
Dan Hendrycks, Mantas Mazeika, Andy Zou, Sahil Patel, Christine Zhu, Jesus
  Navarro, Dawn Song, Bo~Li, and Jacob Steinhardt.
\newblock What {Would} {Jiminy} {Cricket} {Do}? {Towards} {Agents} {That}
  {Behave} {Morally}.
\newblock In \emph{Advances in {Neural} {Information} {Processing} {Systems}},
  volume~1, 2021.

\bibitem[Hockley(1984)]{hockley_rt_1984}
William~E. Hockley.
\newblock Analysis of response time distributions in the study of cognitive
  processes.
\newblock 10\penalty0 (4):\penalty0 598--615, 1984.
\newblock ISSN 1939-1285(Electronic),0278-7393(Print).
\newblock \doi{10.1037/0278-7393.10.4.598}.

\bibitem[Hoffmann and Falkenstein(2012)]{hoffmann_errpage_2012}
Sven Hoffmann and Michael Falkenstein.
\newblock Predictive information processing in the brain: {Errors} and response
  monitoring.
\newblock 83\penalty0 (2):\penalty0 208--212, February 2012.
\newblock ISSN 0167-8760.
\newblock \doi{10.1016/j.ijpsycho.2011.11.015}.
\newblock URL
  \url{http://www.sciencedirect.com/science/article/pii/S016787601100362X}.

\bibitem[Holroyd and Coles(2002)]{holroyd_frn_2002}
Clay~B. Holroyd and Michael G.~H. Coles.
\newblock The neural basis of human error processing: reinforcement learning,
  dopamine, and the error-related negativity.
\newblock 109\penalty0 (4):\penalty0 679--709, October 2002.
\newblock ISSN 0033-295X.
\newblock \doi{10.1037/0033-295X.109.4.679}.

\bibitem[Ibarz et~al.(2018)Ibarz, Leike, Pohlen, Irving, Legg, and
  Amodei]{ibarz_rewardlearning_2018}
Borja Ibarz, Jan Leike, Tobias Pohlen, Geoffrey Irving, Shane Legg, and Dario
  Amodei.
\newblock Reward learning from human preferences and demonstrations in {Atari}.
\newblock In S.~Bengio, H.~Wallach, H.~Larochelle, K.~Grauman, N.~Cesa-Bianchi,
  and R.~Garnett, editors, \emph{Advances in {Neural} {Information}
  {Processing} {Systems}}, pages 8011--8023. Curran Associates, Inc., 2018.

\bibitem[Iturrate et~al.(2010)Iturrate, Montesano, and
  Minguez]{iturrate_robot_2010}
Inaki Iturrate, Luis Montesano, and Javier Minguez.
\newblock Robot reinforcement learning using {EEG}-based reward signals.
\newblock In \emph{{IEEE} {International} {Conference} on {Robotics} and
  {Automation}}, {ICRA} `10, pages 4822--4829. IEEE, May 2010.
\newblock \doi{10.1109/ROBOT.2010.5509734}.

\bibitem[Iturrate et~al.(2013)Iturrate, Montesano, and
  Minguez]{iturrate_errptask_2013}
Inaki Iturrate, Luis Montesano, and Javier Minguez.
\newblock Task-dependent signal variations in {EEG} error-related potentials
  for brain-computer interfaces.
\newblock 10\penalty0 (2):\penalty0 026024, March 2013.
\newblock ISSN 1741-2552.
\newblock \doi{10.1088/1741-2560/10/2/026024}.
\newblock URL \url{https://doi.org/10.1088\%2F1741-2560\%2F10\%2F2\%2F026024}.

\bibitem[Iturrate et~al.(2015)Iturrate, Chavarriaga, Montesano, Minguez, and
  Millan]{iturrate_errprl_2015}
Inaki Iturrate, Ricardo Chavarriaga, Luis Montesano, Javier Minguez, and Jose
  del~R. Millan.
\newblock Teaching brain-machine interfaces as an alternative paradigm to
  neuroprosthetics control.
\newblock 5:\penalty0 13893, September 2015.
\newblock ISSN 2045-2322.
\newblock \doi{10.1038/srep13893}.
\newblock URL \url{https://www.nature.com/articles/srep13893}.

\bibitem[Iwane et~al.(2019)Iwane, Halvagal, Iturrate, Batzianoulis,
  Chavarriaga, Billard, and Millan]{iwane_inferring_2019}
F.~Iwane, M.~S. Halvagal, I.~Iturrate, I.~Batzianoulis, R.~Chavarriaga,
  A.~Billard, and J.~d~R. Millan.
\newblock Inferring subjective preferences on robot trajectories using {EEG}
  signals.
\newblock In \emph{9th International {IEEE}/{EMBS} Conference on Neural
  Engineering}, pages 255--258, 2019.
\newblock \doi{10.1109/NER.2019.8717025}.

\bibitem[Jiang et~al.(2019)Jiang, Stocco, Losey, Abernethy, Prat, and
  Rao]{jiang_brainnet_2019}
Linxing Jiang, Andrea Stocco, Darby~M. Losey, Justin~A. Abernethy, Chantel~S.
  Prat, and Rajesh P.~N. Rao.
\newblock {BrainNet}: {A} {Multi}-{Person} {Brain}-to-{Brain} {Interface} for
  {Direct} {Collaboration} {Between} {Brains}.
\newblock 9\penalty0 (1):\penalty0 1--11, April 2019.
\newblock ISSN 2045-2322.
\newblock \doi{10.1038/s41598-019-41895-7}.
\newblock URL \url{https://www.nature.com/articles/s41598-019-41895-7}.

\bibitem[Kakkos et~al.(2020)Kakkos, Ventouras, Asvestas, Karanasiou, and
  Matsopoulos]{kakkos_errpcomplex_2020}
Ioannis Kakkos, Errikos~M. Ventouras, Pantelis~A. Asvestas, Irene~S.
  Karanasiou, and George~K. Matsopoulos.
\newblock A condition-independent framework for the classification of
  error-related brain activity.
\newblock 58\penalty0 (3):\penalty0 573--587, March 2020.
\newblock ISSN 1741-0444.
\newblock \doi{10.1007/s11517-019-02116-5}.
\newblock URL \url{https://doi.org/10.1007/s11517-019-02116-5}.

\bibitem[Kerous et~al.(2018)Kerous, Skola, and
  Liarokapis]{kerous_bcigames_2018}
Bojan Kerous, Filip Skola, and Fotis Liarokapis.
\newblock {EEG}-based {BCI} and video games: a progress report.
\newblock 22\penalty0 (2):\penalty0 119--135, June 2018.
\newblock ISSN 1434-9957.
\newblock \doi{10.1007/s10055-017-0328-x}.
\newblock URL \url{https://doi.org/10.1007/s10055-017-0328-x}.

\bibitem[Kim and Kirchner(2013)]{kim_transfer_2013}
Su~Kyoung Kim and Elsa~A. Kirchner.
\newblock Classifier {Transferability} in the {Detection} of {Error} {Related}
  {Potentials} from {Observation} to {Interaction}.
\newblock In \emph{{IEEE} {International} {Conference} on {Systems}, {Man}, and
  {Cybernetics}}, {SMC} '13, pages 3360--3365. IEEE, October 2013.
\newblock \doi{10.1109/SMC.2013.573}.

\bibitem[Kim and Kirchner(2016)]{kim_handlingtransfer_2016}
Su~Kyoung Kim and Elsa~A. Kirchner.
\newblock Handling {Few} {Training} {Data}: {Classifier} {Transfer} {Between}
  {Different} {Types} of {Error}-{Related} {Potentials}.
\newblock 24\penalty0 (3):\penalty0 320--332, March 2016.
\newblock ISSN 1558-0210.
\newblock \doi{10.1109/TNSRE.2015.2507868}.

\bibitem[Kim et~al.(2017)Kim, Kirchner, Stefes, and
  Kirchner]{kim_intrinsic_2017}
Su~Kyoung Kim, Elsa~A. Kirchner, Arne Stefes, and Frank Kirchner.
\newblock Intrinsic interactive reinforcement learning - {Using} error-related
  potentials for real world human-robot interaction.
\newblock 7\penalty0 (1):\penalty0 17562, December 2017.
\newblock ISSN 2045-2322.
\newblock \doi{10.1038/s41598-017-17682-7}.
\newblock URL \url{https://www.nature.com/articles/s41598-017-17682-7}.

\bibitem[Kim et~al.(2020)Kim, Kirchner, and Kirchner]{kim_flexible_2020}
Su~Kyoung Kim, Elsa~A. Kirchner, and Frank Kirchner.
\newblock Flexible online adaptation of learning strategy using {EEG}-based
  reinforcement signals in real-world robotic applications.
\newblock In \emph{{IEEE} {International} {Conference} on {Robotics} and
  {Automation}}, {ICRA} '20, pages 4885--4891. IEEE, May 2020.
\newblock \doi{10.1109/ICRA40945.2020.9197538}.

\bibitem[Kirchner et~al.(2019)Kirchner, Fairclough, and
  Kirchner]{kirchner_robointerface_2019}
Elsa~A. Kirchner, Stephen~H. Fairclough, and Frank Kirchner.
\newblock \emph{Embedded Multimodal Interfaces in Robotics: Applications,
  Future Trends, and Societal Implications}, pages 523--576.
\newblock Association for Computing Machinery and Morgan \& Claypool, 2019.
\newblock ISBN 9781970001754.
\newblock URL \url{https://doi.org/10.1145/3233795.3233810}.

\bibitem[Klonowski(2009)]{klonowski_eegquest_2009}
Wlodzimierz Klonowski.
\newblock Everything you wanted to ask about {EEG} but were afraid to get the
  right answer.
\newblock 3:\penalty0 2, May 2009.
\newblock ISSN 1753-4631.
\newblock \doi{10.1186/1753-4631-3-2}.
\newblock URL \url{https://www.ncbi.nlm.nih.gov/pmc/articles/PMC2698918/}.

\bibitem[Knox et~al.(2013)Knox, Stone, and Breazeal]{knox_robot_2013}
W.~Bradely Knox, Peter Stone, and Cynthia Breazeal.
\newblock Training a robot via human feedback: A case study.
\newblock In \emph{Social Robotics}, pages 460--470. Springer International
  Publishing, 2013.
\newblock ISBN 978-3-319-02675-6.

\bibitem[Knox(2012)]{knox_thesis_2012}
W.~Bradley Knox.
\newblock \emph{Learning from human-generated reward}.
\newblock PhD thesis, December 2012.
\newblock URL \url{https://repositories.lib.utexas.edu/handle/2152/19472}.

\bibitem[Knox and Stone(2008)]{knox_tamer_2008}
W.~Bradley Knox and Peter Stone.
\newblock {TAMER}: {Training} an {Agent} {Manually} via {Evaluative}
  {Reinforcement}.
\newblock In \emph{7th {IEEE} {International} {Conference} on {Development} and
  {Learning}}, {ICDL} '08, pages 292--297. IEEE, August 2008.
\newblock \doi{10.1109/DEVLRN.2008.4640845}.

\bibitem[Knox and Stone(2009)]{knox_interactively_2009}
W.~Bradley Knox and Peter Stone.
\newblock Interactively shaping agents via human reinforcement: the {TAMER}
  framework.
\newblock In \emph{Proceedings of the fifth international conference on
  {Knowledge} capture}, K-{CAP} '09, pages 9--16. Association for Computing
  Machinery, September 2009.
\newblock ISBN 978-1-60558-658-8.
\newblock \doi{10.1145/1597735.1597738}.
\newblock URL \url{https://doi.org/10.1145/1597735.1597738}.

\bibitem[Knox and Stone(2010)]{knox_combining_2010}
W.~Bradley Knox and Peter Stone.
\newblock Combining manual feedback with subsequent mdp reward signals for
  reinforcement learning.
\newblock In \emph{Proceedings of the 9th International Conference on
  Autonomous Agents and Multiagent Systems}, volume~1 of \emph{AAMAS '10},
  pages 5--12. International Foundation for Autonomous Agents and Multiagent
  Systems, 2010.
\newblock ISBN 9780982657119.

\bibitem[Knox and Stone(2012{\natexlab{a}})]{knox_myopic_2012}
W.~Bradley Knox and Peter Stone.
\newblock Reinforcement learning from human reward: {Discounting} in episodic
  tasks.
\newblock In \emph{{IEEE} {RO}-{MAN}: {The} 21st {IEEE} {International}
  {Symposium} on {Robot} and {Human} {Interactive} {Communication}}, RO-MAN
  '12, pages 878--885. IEEE, September 2012{\natexlab{a}}.
\newblock \doi{10.1109/ROMAN.2012.6343862}.

\bibitem[Knox and Stone(2012{\natexlab{b}})]{knox_reinforcement_2012}
W.~Bradley Knox and Peter Stone.
\newblock Reinforcement learning from simultaneous human and {MDP} reward.
\newblock In \emph{Proceedings of the 11th {International} {Conference} on
  {Autonomous} {Agents} and {Multiagent} {Systems}}, volume~1 of \emph{{AAMAS}
  '12}, pages 475--482. International Foundation for Autonomous Agents and
  Multiagent Systems, June 2012{\natexlab{b}}.
\newblock ISBN 978-0-9817381-1-6.

\bibitem[Knox and Stone(2013)]{knox_learning_2013}
W.~Bradley Knox and Peter Stone.
\newblock Learning non-myopically from human-generated reward.
\newblock In \emph{Proceedings of the international conference on {Intelligent}
  user interfaces}, {IUI} '13, pages 191--202. Association for Computing
  Machinery, March 2013.
\newblock ISBN 978-1-4503-1965-2.
\newblock \doi{10.1145/2449396.2449422}.
\newblock URL \url{https://doi.org/10.1145/2449396.2449422}.

\bibitem[Knox and Stone(2015)]{knox_framing_2015}
W.~Bradley Knox and Peter Stone.
\newblock Framing reinforcement learning from human reward: {Reward}
  positivity, temporal discounting, episodicity, and performance.
\newblock 225:\penalty0 24--50, August 2015.
\newblock ISSN 0004-3702.
\newblock \doi{10.1016/j.artint.2015.03.009}.
\newblock URL
  \url{http://www.sciencedirect.com/science/article/pii/S0004370215000557}.

\bibitem[Kosmyna et~al.(2016)Kosmyna, Tarpin-Bernard, Bonnefond, and
  Rivet]{kosmyna_bcienv_2016}
Nataliya Kosmyna, Franck Tarpin-Bernard, Nicolas Bonnefond, and Bertrand Rivet.
\newblock Feasibility of {BCI} {Control} in a {Realistic} {Smart} {Home}
  {Environment}.
\newblock 10:\penalty0 416, August 2016.
\newblock ISSN 1662-5161.
\newblock \doi{10.3389/fnhum.2016.00416}.
\newblock URL \url{https://www.ncbi.nlm.nih.gov/pmc/articles/PMC4999433/}.

\bibitem[Lawhern et~al.(2018)Lawhern, Solon, Waytowich, Gordon, Hung, and
  Lance]{lawhern_eegnet_2018}
Vernon~J. Lawhern, Amelia~J. Solon, Nicholas~R. Waytowich, Stephen~M. Gordon,
  Chou~P. Hung, and Brent~J. Lance.
\newblock {EEGNet}: a compact convolutional neural network for {EEG}-based
  brain-computer interfaces.
\newblock 15\penalty0 (5):\penalty0 056013, July 2018.
\newblock ISSN 1741-2552.
\newblock \doi{10.1088/1741-2552/aace8c}.
\newblock URL \url{https://doi.org/10.1088\%2F1741-2552\%2Faace8c}.

\bibitem[Leike et~al.(2018)Leike, Krueger, Everitt, Martic, Maini, and
  Legg]{leike_alignment_2018}
Jan Leike, David Krueger, Tom Everitt, Miljan Martic, Vishal Maini, and Shane
  Legg.
\newblock Scalable agent alignment via reward modeling: a research direction,
  November 2018.

\bibitem[Li et~al.(2018)Li, He, Gomez, and Nakamura]{li_interactive_2018}
Guangliang Li, Bo~He, Randy Gomez, and Keisuke Nakamura.
\newblock Interactive {Reinforcement} {Learning} from {Demonstration} and
  {Human} {Evaluative} {Feedback}.
\newblock In \emph{27th {IEEE} {International} {Symposium} on {Robot} and
  {Human} {Interactive} {Communication}}, {RO}-{MAN} '18, pages 1156--1162.
  Institute of Electrical and Electronics Engineers, August 2018.
\newblock \doi{10.1109/ROMAN.2018.8525837}.

\bibitem[Li et~al.(2019)Li, Gomez, Nakamura, and He]{li_irlsurvey_2019}
Guangliang Li, Randy Gomez, Keisuke Nakamura, and Bo~He.
\newblock Human-{Centered} {Reinforcement} {Learning}: {A} {Survey}.
\newblock 49\penalty0 (4):\penalty0 337--349, August 2019.
\newblock ISSN 2168-2305.
\newblock \doi{10.1109/THMS.2019.2912447}.

\bibitem[Li et~al.(2022)Li, Zhang, Tiwari, Song, Hu, Yang, Zhao, Kumar, and
  Marttinen]{li_emotionsurvey_2022}
Xiang Li, Yazhou Zhang, Prayag Tiwari, Dawei Song, Bin Hu, Meihong Yang,
  Zhigang Zhao, Neeraj Kumar, and Pekka Marttinen.
\newblock {EEG} based emotion recognition: A tutorial and review.
\newblock 2022.
\newblock ISSN 0360-0300.
\newblock \doi{10.1145/3524499}.
\newblock URL \url{https://doi.org/10.1145/3524499}.

\bibitem[{Li, Guangliang}(2016)]{li_thesis_2016}
{Li, Guangliang}.
\newblock \emph{Socially intelligent autonomous agents that learn from human
  reward}.
\newblock PhD thesis, University of Amsterdam, 2016.

\bibitem[Lin et~al.(2020)Lin, Ma, Gomez, Nakamura, He, and
  Li]{lin_irlsocial_2020}
Jinying Lin, Zhen Ma, Randy Gomez, Keisuke Nakamura, Bo~He, and Guangliang Li.
\newblock A {Review} on {Interactive} {Reinforcement} {Learning} {From} {Human}
  {Social} {Feedback}.
\newblock 8:\penalty0 120757--120765, 2020.
\newblock ISSN 2169-3536.
\newblock \doi{10.1109/ACCESS.2020.3006254}.

\bibitem[Lin(1992)]{lin_self_1992}
Long-Ji Lin.
\newblock Self-improving reactive agents based on reinforcement learning,
  planning and teaching.
\newblock 8\penalty0 (3):\penalty0 293--321, May 1992.
\newblock ISSN 1573-0565.
\newblock \doi{10.1007/BF00992699}.
\newblock URL \url{https://doi.org/10.1007/BF00992699}.

\bibitem[Lopes-Dias et~al.(2019)Lopes-Dias, Sburlea, and
  M{\"u}ller-Putz]{lopesdias_errp_2019}
Catarina Lopes-Dias, Andreea~I. Sburlea, and Gernot~R. M{\"u}ller-Putz.
\newblock Online asynchronous decoding of error-related potentials during the
  continuous control of a robot.
\newblock 9\penalty0 (1):\penalty0 17596, November 2019.
\newblock ISSN 2045-2322.
\newblock \doi{10.1038/s41598-019-54109-x}.
\newblock URL \url{https://www.nature.com/articles/s41598-019-54109-x}.

\bibitem[Lotte(2014)]{lotte_tutorial_2014}
Fabien Lotte.
\newblock A {Tutorial} on {EEG} {Signal}-processing {Techniques} for
  {Mental}-state {Recognition} in {Brain}-{Computer} {Interfaces}.
\newblock In Eduardo~Reck Miranda and Julien Castet, editors, \emph{Guide to
  {Brain}-{Computer} {Music} {Interfacing}}, pages 133--161. Springer, 2014.
\newblock ISBN 978-1-4471-6584-2.
\newblock \doi{10.1007/978-1-4471-6584-2_7}.
\newblock URL \url{https://doi.org/10.1007/978-1-4471-6584-2_7}.

\bibitem[Lotte et~al.(2018)Lotte, Bougrain, Cichocki, Clerc, Congedo,
  Rakotomamonjy, and Yger]{lotte_bcisurvey_2018}
Fabien Lotte, Laurent Bougrain, Andrzej Cichocki, Maureen Clerc, Marco Congedo,
  Alain Rakotomamonjy, and Florian Yger.
\newblock A review of classification algorithms for {EEG}-based brain-computer
  interfaces: a 10 year update.
\newblock 15\penalty0 (3):\penalty0 031005, April 2018.
\newblock ISSN 1741-2552.
\newblock \doi{10.1088/1741-2552/aab2f2}.
\newblock URL \url{https://doi.org/10.1088\%2F1741-2552\%2Faab2f2}.

\bibitem[Luck(2014)]{luck_erpbook_2014}
Steven~J. Luck.
\newblock \emph{An {Introduction} to the {Event}-{Related} {Potential}
  {Technique}}.
\newblock MIT Press, June 2014.
\newblock ISBN 978-0-262-32406-9.

\bibitem[Luo et~al.(2018)Luo, Fan, Lv, and Zhou]{luo_drlpref_2018}
Tian-jian Luo, Ya-chao Fan, Ji-tu Lv, and Chang-le Zhou.
\newblock Deep reinforcement learning from error-related potentials via an
  {EEG}-based brain-computer interface.
\newblock In \emph{2018 {IEEE} {International} {Conference} on {Bioinformatics}
  and {Biomedicine}}, {BIBM} '18, pages 697--701. IEEE, December 2018.
\newblock \doi{10.1109/BIBM.2018.8621183}.

\bibitem[MacGlashan et~al.(2017)MacGlashan, Ho, Loftin, Peng, Wang, Roberts,
  Taylor, and Littman]{macglashan_coach_2017}
James MacGlashan, Mark~K Ho, Robert Loftin, Bei Peng, Guan Wang, David~L.
  Roberts, Matthew~E. Taylor, and Michael~L. Littman.
\newblock Interactive learning from policy-dependent human feedback.
\newblock In \emph{International Conference on Machine Learning}, volume~70 of
  \emph{ICML'17}, pages 2285--2294. JMLR.org, 2017.

\bibitem[Masci et~al.(2011)Masci, Meier, Cire{\c s}an, and
  Schmidhuber]{masci_cae_2011}
Jonathan Masci, Ueli Meier, Dan Cire{\c s}an, and J{\"u}rgen Schmidhuber.
\newblock Stacked {Convolutional} {Auto}-{Encoders} for {Hierarchical}
  {Feature} {Extraction}.
\newblock In \emph{Artificial {Neural} {Networks} and {Machine} {Learning}},
  ICANN '11, pages 52--59. Springer, 2011.
\newblock ISBN 978-3-642-21735-7.
\newblock \doi{10.1007/978-3-642-21735-7_7}.

\bibitem[Millan(2004)]{millan_onlinebci_2004}
Jose del~R. Millan.
\newblock On the need for on-line learning in brain-computer interfaces.
\newblock In \emph{{IEEE} {International} {Joint} {Conference} on {Neural}
  {Networks} ({IEEE} {Cat}. {No}.{04CH37541})}, volume~4 of \emph{{IJCNN} '04},
  pages 2877--2882 vol.4. IEEE, July 2004.
\newblock \doi{10.1109/IJCNN.2004.1381116}.

\bibitem[Mnih et~al.(2015)Mnih, Kavukcuoglu, Silver, Rusu, Veness, Bellemare,
  Graves, Riedmiller, Fidjeland, Ostrovski, Petersen, Beattie, Sadik,
  Antonoglou, King, Kumaran, Wierstra, Legg, and Hassabis]{mnih_dqn_2015}
Volodymyr Mnih, Koray Kavukcuoglu, David Silver, Andrei~A. Rusu, Joel Veness,
  Marc~G. Bellemare, Alex Graves, Martin Riedmiller, Andreas~K. Fidjeland,
  Georg Ostrovski, Stig Petersen, Charles Beattie, Amir Sadik, Ioannis
  Antonoglou, Helen King, Dharshan Kumaran, Daan Wierstra, Shane Legg, and
  Demis Hassabis.
\newblock Human-level control through deep reinforcement learning.
\newblock 518\penalty0 (7540):\penalty0 529--533, February 2015.
\newblock ISSN 1476-4687.
\newblock \doi{10.1038/nature14236}.
\newblock URL \url{https://www.nature.com/articles/nature14236}.

\bibitem[Mnih et~al.(2016)Mnih, Badia, Mirza, Graves, Harley, Lillicrap,
  Silver, and Kavukcuoglu]{mnih_a3c_2016}
Volodymyr Mnih, Adri\`{a}~Puigdom\`{e}nech Badia, Mehdi Mirza, Alex Graves, Tim
  Harley, Timothy~P. Lillicrap, David Silver, and Koray Kavukcuoglu.
\newblock Asynchronous methods for deep reinforcement learning.
\newblock In \emph{International Conference on Machine Learning}, pages
  1928--1937. JMLR, 2016.

\bibitem[Myrden and Chau(2017)]{myrden_cognitivestates_2017}
Andrew Myrden and Tom Chau.
\newblock A {Passive} {EEG}-{BCI} for {Single}-{Trial} {Detection} of {Changes}
  in {Mental} {State}.
\newblock 25\penalty0 (4):\penalty0 345--356, April 2017.
\newblock ISSN 1558-0210.
\newblock \doi{10.1109/TNSRE.2016.2641956}.

\bibitem[Nahian et~al.(2021)Nahian, Frazier, Harrison, and
  Riedl]{nahian_training_2021}
Md~Sultan~Al Nahian, Spencer Frazier, Brent Harrison, and Mark Riedl.
\newblock Training {Value}-{Aligned} {Reinforcement} {Learning} {Agents}
  {Using} a {Normative} {Prior}, April 2021.
\newblock URL \url{http://arxiv.org/abs/2104.09469}.

\bibitem[Najar and Chetouani(2021)]{najar_irlsurvey_2021}
Anis Najar and Mohamed Chetouani.
\newblock Reinforcement {Learning} {With} {Human} {Advice}: {A} {Survey}.
\newblock 8:\penalty0 74, 2021.
\newblock ISSN 2296-9144.
\newblock \doi{10.3389/frobt.2021.584075}.
\newblock URL
  \url{https://www.frontiersin.org/article/10.3389/frobt.2021.584075}.

\bibitem[Ng and Russell(2000)]{ng_inverse_2000}
Andrew~Y. Ng and Stuart Russell.
\newblock Algorithms for {Inverse} {Reinforcement} {Learning}.
\newblock In \emph{{International} {Conference} on {Machine} {Learning}}, ICML
  '00, pages 663--670. Morgan Kaufmann, 2000.

\bibitem[Ng et~al.(1999)Ng, Harada, and Russell]{ng_reward_1999}
Andrew~Y. Ng, Daishi Harada, and Stuart Russell.
\newblock Policy invariance under reward transformations: {Theory} and
  application to reward shaping.
\newblock In \emph{{International} {Conference} on {Machine} {Learning}}, ICML
  '99, pages 278--287. Morgan Kaufmann, 1999.

\bibitem[OpenAI et~al.(2019)OpenAI, Berner, Brockman, Chan, Cheung,
  D{\k{e}}biak, Dennison, Farhi, Fischer, Hashme, Hesse, Józefowicz, Gray,
  Olsson, Pachocki, Petrov, de~Oliveira~Pinto, Raiman, Salimans, Schlatter,
  Schneider, Sidor, Sutskever, Tang, Wolski, and Zhang]{openai_dota_2019}
OpenAI, Christopher Berner, Greg Brockman, Brooke Chan, Vicki Cheung,
  Przemys{\l}aw D{\k{e}}biak, Christy Dennison, David Farhi, Quirin Fischer,
  Shariq Hashme, Chris Hesse, Rafal Józefowicz, Scott Gray, Catherine Olsson,
  Jakub Pachocki, Michael Petrov, Henrique~Pondé de~Oliveira~Pinto, Jonathan
  Raiman, Tim Salimans, Jeremy Schlatter, Jonas Schneider, Szymon Sidor, Ilya
  Sutskever, Jie Tang, Filip Wolski, and Susan Zhang.
\newblock Dota 2 with large scale deep reinforcement learning, 2019.

\bibitem[Osa et~al.(2018)Osa, Pajarinen, Neumann, Bagnell, Abbeel, and
  Peters]{osa_imitation_2018}
Takayuki Osa, Joni Pajarinen, Gerhard Neumann, J.~Andrew Bagnell, Pieter
  Abbeel, and Jan Peters.
\newblock An algorithmic perspective on imitation learning.
\newblock 7\penalty0 (1):\penalty0 1--179, 2018.
\newblock ISSN 1935-8253, 1935-8261.
\newblock \doi{10.1561/2300000053}.
\newblock URL \url{https://www.nowpublishers.com/article/Details/ROB-053}.

\bibitem[Ouyang et~al.(2022)Ouyang, Wu, Jiang, Almeida, Wainwright, Mishkin,
  Zhang, Agarwal, Slama, Ray, Schulman, Hilton, Kelton, Miller, Simens, Askell,
  Welinder, Christiano, Leike, and Lowe]{ouyang_training_2022}
Long Ouyang, Jeffrey Wu, Xu~Jiang, Diogo Almeida, Carroll Wainwright, Pamela
  Mishkin, Chong Zhang, Sandhini Agarwal, Katarina Slama, Alex Ray, John
  Schulman, Jacob Hilton, Fraser Kelton, Luke Miller, Maddie Simens, Amanda
  Askell, Peter Welinder, Paul~F. Christiano, Jan Leike, and Ryan Lowe.
\newblock Training language models to follow instructions with human feedback.
\newblock 35:\penalty0 27730--27744, 2022.
\newblock URL
  \url{https://proceedings.neurips.cc/paper_files/paper/2022/hash/b1efde53be364a73914f58805a001731-Abstract-Conference.html}.

\bibitem[Polich(2007)]{polich_p3review_2007}
John Polich.
\newblock Updating {P300}: {An} {Integrative} {Theory} of {P3a} and {P3b}.
\newblock 118\penalty0 (10):\penalty0 2128--2148, October 2007.
\newblock ISSN 1388-2457.
\newblock \doi{10.1016/j.clinph.2007.04.019}.
\newblock URL \url{https://www.ncbi.nlm.nih.gov/pmc/articles/PMC2715154/}.

\bibitem[Poole and Lee(2022)]{poole_errp_2022}
Benjamin Poole and Minwoo Lee.
\newblock Error-related potential variability: Exploring the effects on
  classification and transferability.
\newblock In \emph{IEEE Symposium Series on Computational Intelligence in
  Brain-Computer Interfaces}. IEEE, 2022.

\bibitem[Ramadan and Vasilakos(2017)]{ramadan_bcisurvey_2017}
Rabie~A. Ramadan and Athanasios~V. Vasilakos.
\newblock Brain computer interface: control signals review.
\newblock 223:\penalty0 26--44, February 2017.
\newblock ISSN 0925-2312.
\newblock \doi{10.1016/j.neucom.2016.10.024}.
\newblock URL
  \url{http://www.sciencedirect.com/science/article/pii/S0925231216312152}.

\bibitem[Rao et~al.(2014)Rao, Stocco, Bryan, Sarma, Youngquist, Wu, and
  Prat]{rao_braintobrain_2014}
Rajesh P.~N. Rao, Andrea Stocco, Matthew Bryan, Devapratim Sarma, Tiffany~M.
  Youngquist, Joseph Wu, and Chantel~S. Prat.
\newblock A {Direct} {Brain}-to-{Brain} {Interface} in {Humans}.
\newblock 9\penalty0 (11):\penalty0 e111332, November 2014.
\newblock ISSN 1932-6203.

\bibitem[Rashid et~al.(2020)Rashid, Sulaiman, P.~P. Abdul~Majeed, Musa,
  Ab.~Nasir, Bari, and Khatun]{rashid_bcisurvey_2020}
Mamunur Rashid, Norizam Sulaiman, Anwar P.~P. Abdul~Majeed, Rabiu~Muazu Musa,
  Ahmad~Fakhri Ab.~Nasir, Bifta~Sama Bari, and Sabira Khatun.
\newblock Current {Status}, {Challenges}, and {Possible} {Solutions} of
  {EEG}-{Based} {Brain}-{Computer} {Interface}: {A} {Comprehensive} {Review}.
\newblock 14:\penalty0 25, 2020.
\newblock ISSN 1662-5218.
\newblock \doi{10.3389/fnbot.2020.00025}.
\newblock URL
  \url{https://www.frontiersin.org/articles/10.3389/fnbot.2020.00025/full}.

\bibitem[Rodriguez-Soto et~al.(2022)Rodriguez-Soto, Serramia, Lopez-Sanchez,
  and Rodriguez-Aguilar]{rodriguez_instilling_2022}
Manel Rodriguez-Soto, Marc Serramia, Maite Lopez-Sanchez, and Juan~Antonio
  Rodriguez-Aguilar.
\newblock Instilling moral value alignment by means of multi-objective
  reinforcement learning.
\newblock 24\penalty0 (1):\penalty0 9, January 2022.
\newblock ISSN 1572-8439.
\newblock \doi{10.1007/s10676-022-09635-0}.
\newblock URL \url{https://doi.org/10.1007/s10676-022-09635-0}.

\bibitem[Salazar-Gomez et~al.(2017)Salazar-Gomez, DelPreto, Gil, Guenther, and
  Rus]{salazar_errprobot_2017}
Andres~F. Salazar-Gomez, Joseph DelPreto, Stephanie Gil, Frank~H. Guenther, and
  Daniela Rus.
\newblock Correcting robot mistakes in real time using {EEG} signals.
\newblock In \emph{{IEEE} {International} {Conference} on {Robotics} and
  {Automation}}, {ICRA} '17, pages 6570--6577. samek\_erpnonstation\_2015, May
  2017.
\newblock \doi{10.1109/ICRA.2017.7989777}.

\bibitem[Samek and M{\"u}ller(2015)]{samek_erpnonstation_2015}
Wojciech Samek and Klaus-Robert M{\"u}ller.
\newblock Tackling noise, artifacts and nonstationarity in {BCI} with robust
  divergences.
\newblock In \emph{23rd {European} {Signal} {Processing} {Conference}},
  {EUSIPCO} '15, pages 2741--2745. IEEE, August 2015.
\newblock \doi{10.1109/EUSIPCO.2015.7362883}.

\bibitem[Schmidt et~al.(2012)Schmidt, Blankertz, and
  Treder]{schmidt_mentaltype_2012}
Nico~M. Schmidt, Benjamin Blankertz, and Matthias~S. Treder.
\newblock Online detection of error-related potentials boosts the performance
  of mental typewriters.
\newblock 13:\penalty0 19, February 2012.
\newblock ISSN 1471-2202.
\newblock \doi{10.1186/1471-2202-13-19}.
\newblock URL \url{https://www.ncbi.nlm.nih.gov/pmc/articles/PMC3315432/}.

\bibitem[Schulman et~al.(2015)Schulman, Levine, Moritz, Jordan, and
  Abbeel]{schulman_trust_2017}
John Schulman, Sergey Levine, Philipp Moritz, Michael Jordan, and Pieter
  Abbeel.
\newblock Trust region policy optimization.
\newblock In \emph{International Conference on Machine Learning}, ICML'15,
  pages 1889--1897. JMLR, 2015.

\bibitem[Schulman et~al.(2017)Schulman, Wolski, Dhariwal, Radford, and
  Klimov]{schulman_ppo_2017}
John Schulman, Filip Wolski, Prafulla Dhariwal, Alec Radford, and Oleg Klimov.
\newblock Proximal {Policy} {Optimization} {Algorithms}, August 2017.

\bibitem[Sensoy et~al.(2018)Sensoy, Kaplan, and
  Kandemir]{sensoy_evidential_2018}
Murat Sensoy, Lance Kaplan, and Melih Kandemir.
\newblock Evidential {Deep} {Learning} to {Quantify} {Classification}
  {Uncertainty}.
\newblock In S.~Bengio, H.~Wallach, H.~Larochelle, K.~Grauman, N.~Cesa-Bianchi,
  and R.~Garnett, editors, \emph{Advances in {Neural} {Information}
  {Processing} {Systems}}, volume~31. Curran Associates, Inc., 2018.
\newblock URL
  \url{https://proceedings.neurips.cc/paper/2018/file/a981f2b708044d6fb4a71a1463242520-Paper.pdf}.

\bibitem[Shenoy et~al.(2006)Shenoy, Krauledat, Blankertz, Rao, and
  M{\"u}ller]{shenoy_nonstation_2006}
Pradeep Shenoy, Matthias Krauledat, Benjamin Blankertz, Rajesh~P. Rao, and
  Klaus-Robert M{\"u}ller.
\newblock Towards adaptive classification for {BCI}.
\newblock 3\penalty0 (1):\penalty0 R13--R23, 2006.
\newblock \doi{10.1088/1741-2560/3/1/R02}.

\bibitem[Spuler and Niethammer(2015)]{spuler_continuouserrp_2015}
Martin Spuler and Christian Niethammer.
\newblock Error-related potentials during continuous feedback: using {EEG} to
  detect errors of different type and severity.
\newblock 9:\penalty0 155, 2015.
\newblock ISSN 1662-5161.
\newblock \doi{10.3389/fnhum.2015.00155}.
\newblock URL
  \url{https://www.frontiersin.org/articles/10.3389/fnhum.2015.00155/full}.

\bibitem[Sutton(1984)]{sutton_temporal_1984}
Richard~S. Sutton.
\newblock \emph{Temporal {Credit} {Assignment} in {Reinforcement} {Learning}}.
\newblock PhD thesis, University of Massachusetts Amherst, January 1984.
\newblock URL \url{https://scholarworks.umass.edu/dissertations/AAI8410337}.

\bibitem[Sutton and Barto(2018)]{sutton_rl_2018}
Richard~S. Sutton and Andrew~G. Barto.
\newblock \emph{Reinforcement {Learning}: {An} {Introduction}}.
\newblock MIT Press, October 2018.
\newblock ISBN 978-0-262-35270-3.

\bibitem[Tenorio-Gonzalez et~al.(2010)Tenorio-Gonzalez, Morales, and
  Villasenor-Pineda]{kuri_dynamic_2010}
Ana~C. Tenorio-Gonzalez, Eduardo~F. Morales, and Luis Villasenor-Pineda.
\newblock Dynamic {Reward} {Shaping}: {Training} a {Robot} by {Voice}.
\newblock In Angel Kuri-Morales and Guillermo~R. Simari, editors,
  \emph{Advances in {Artificial} {Intelligence} - {IBERAMIA} 2010}, volume
  6433, pages 483--492. Springer Berlin Heidelberg, 2010.
\newblock \doi{10.1007/978-3-642-16952-6\_49}.

\bibitem[Thomaz and Breazeal(2006)]{thomaz_reinforcement_2006}
Andrea~L. Thomaz and Cynthia Breazeal.
\newblock Reinforcement learning with human teachers: evidence of feedback and
  guidance with implications for learning performance.
\newblock In \emph{Proceedings of the 21st national conference on {Artificial}
  intelligence - {Volume} 1}, {AAAI}'06, pages 1000--1005. AAAI Press, July
  2006.
\newblock ISBN 978-1-57735-281-5.

\bibitem[Thomaz and Breazeal(2007)]{thomaz_posneg_2007}
Andrea~L. Thomaz and Cynthia Breazeal.
\newblock Asymmetric {Interpretations} of {Positive} and {Negative} {Human}
  {Feedback} for a {Social} {Learning} {Agent}.
\newblock In \emph{{The} 16th {IEEE} {International} {Symposium} on {Robot} and
  {Human} {Interactive} {Communication}}, RO-MAN '07, pages 720--725. IEEE,
  2007.
\newblock ISBN 978-1-4244-1634-9.
\newblock \doi{10.1109/ROMAN.2007.4415180}.
\newblock URL \url{http://ieeexplore.ieee.org/document/4415180/}.

\bibitem[Thomaz and Breazeal(2008)]{thomaz_teachable_2008}
Andrea~L. Thomaz and Cynthia Breazeal.
\newblock Teachable robots: {Understanding} human teaching behavior to build
  more effective robot learners.
\newblock 172\penalty0 (6):\penalty0 716--737, April 2008.
\newblock ISSN 0004-3702.
\newblock \doi{10.1016/j.artint.2007.09.009}.
\newblock URL
  \url{https://www.sciencedirect.com/science/article/pii/S000437020700135X}.

\bibitem[Tiwari et~al.(2018)Tiwari, Edla, Dodia, and
  Bablani]{tiwari_bcisurvey_2018}
Neha Tiwari, Damodar~Reddy Edla, Shubham Dodia, and Annushree Bablani.
\newblock Brain computer interface: {A} comprehensive survey.
\newblock 26:\penalty0 118--129, October 2018.
\newblock ISSN 2212-683X.
\newblock \doi{10.1016/j.bica.2018.10.005}.
\newblock URL
  \url{http://www.sciencedirect.com/science/article/pii/S2212683X18301142}.

\bibitem[Todorov et~al.(2012)Todorov, Erez, and Tassa]{todorov_mujoco_2012}
Emanuel Todorov, Tom Erez, and Yuval Tassa.
\newblock Mujoco: A physics engine for model-based control.
\newblock In \emph{2012 IEEE/RSJ International Conference on Intelligent Robots
  and Systems}, pages 5026--5033. IEEE, 2012.
\newblock \doi{10.1109/IROS.2012.6386109}.

\bibitem[van Hasselt et~al.(2016)van Hasselt, Guez, and
  Silver]{hasselt_ddqn_2016}
Hado van Hasselt, Arthur Guez, and David Silver.
\newblock Deep reinforcement learning with double q-learning.
\newblock 30\penalty0 (1), 2016.
\newblock URL \url{https://ojs.aaai.org/index.php/AAAI/article/view/10295}.

\bibitem[van Schie et~al.(2004)van Schie, Mars, Coles, and
  Bekkering]{van_ern_2004}
Hein~T. van Schie, Rogier~B. Mars, Michael G.~H. Coles, and Harold Bekkering.
\newblock Modulation of activity in medial frontal and motor cortices during
  error observation.
\newblock 7\penalty0 (5):\penalty0 549--554, May 2004.
\newblock ISSN 1546-1726.
\newblock \doi{10.1038/nn1239}.
\newblock URL \url{https://www.nature.com/articles/nn1239}.

\bibitem[Vansteensel et~al.(2017)Vansteensel, Kristo, Aarnoutse, and
  Ramsey]{vansteensel_questionnaire_2017}
Mariska~J. Vansteensel, Gert Kristo, Erik~J. Aarnoutse, and Nick~F. Ramsey.
\newblock The brain-computer interface researcher's questionnaire: from
  research to application.
\newblock 4\penalty0 (4):\penalty0 236--247, October 2017.
\newblock ISSN 2326-263X.
\newblock \doi{10.1080/2326263X.2017.1366237}.
\newblock URL \url{https://doi.org/10.1080/2326263X.2017.1366237}.

\bibitem[Vien et~al.(2013)Vien, Ertel, and Chung]{vien_actamer_2013}
Ngo~Anh Vien, Wolfgang Ertel, and Tae~Choong Chung.
\newblock Learning via human feedback in continuous state and action spaces.
\newblock 39\penalty0 (2):\penalty0 267--278, September 2013.
\newblock ISSN 1573-7497.
\newblock \doi{10.1007/s10489-012-0412-6}.
\newblock URL \url{https://doi.org/10.1007/s10489-012-0412-6}.

\bibitem[Vinyals et~al.(2019)Vinyals, Babuschkin, Czarnecki, Mathieu, Dudzik,
  Chung, Choi, Powell, Ewalds, Georgiev, Oh, Horgan, Kroiss, Danihelka, Huang,
  Sifre, Cai, Agapiou, Jaderberg, Vezhnevets, Leblond, Pohlen, Dalibard,
  Budden, Sulsky, Molloy, Paine, Gulcehre, Wang, Pfaff, Wu, Ring, Yogatama,
  W{\"u}nsch, McKinney, Smith, Schaul, Lillicrap, Kavukcuoglu, Hassabis, Apps,
  and Silver]{vinyals_alphastar_2019}
Oriol Vinyals, Igor Babuschkin, Wojciech~M. Czarnecki, Micha{\"e}l Mathieu,
  Andrew Dudzik, Junyoung Chung, David~H. Choi, Richard Powell, Timo Ewalds,
  Petko Georgiev, Junhyuk Oh, Dan Horgan, Manuel Kroiss, Ivo Danihelka, Aja
  Huang, Laurent Sifre, Trevor Cai, John~P. Agapiou, Max Jaderberg,
  Alexander~S. Vezhnevets, R{\'e}mi Leblond, Tobias Pohlen, Valentin Dalibard,
  David Budden, Yury Sulsky, James Molloy, Tom~L. Paine, Caglar Gulcehre, Ziyu
  Wang, Tobias Pfaff, Yuhuai Wu, Roman Ring, Dani Yogatama, Dario W{\"u}nsch,
  Katrina McKinney, Oliver Smith, Tom Schaul, Timothy Lillicrap, Koray
  Kavukcuoglu, Demis Hassabis, Chris Apps, and David Silver.
\newblock Grandmaster level in {StarCraft} {II} using multi-agent reinforcement
  learning.
\newblock 575\penalty0 (7782):\penalty0 350--354, November 2019.
\newblock ISSN 1476-4687.
\newblock \doi{10.1038/s41586-019-1724-z}.
\newblock URL \url{https://www.nature.com/articles/s41586-019-1724-z}.

\bibitem[Wang et~al.(2020)Wang, Shi, Akinola, and Allen]{wang_maximizing_2020}
Zizhao Wang, Junyao Shi, Iretiayo Akinola, and Peter Allen.
\newblock Maximizing {BCI} {Human} {Feedback} using {Active} {Learning}.
\newblock In \emph{{IEEE}/{RSJ} {International} {Conference} on {Intelligent}
  {Robots} and {Systems}}, pages 10945--10951. IEEE, October 2020.
\newblock ISBN 978-1-72816-212-6.
\newblock \doi{10.1109/IROS45743.2020.9341669}.
\newblock URL \url{https://ieeexplore.ieee.org/document/9341669/}.

\bibitem[Warnell et~al.(2018)Warnell, Waytowich, Lawhern, and
  Stone]{warnell_deeptamer_2018}
Garrett Warnell, Nicholas Waytowich, Vernon Lawhern, and Peter Stone.
\newblock Deep {TAMER}: {Interactive} {Agent} {Shaping} in {High}-{Dimensional}
  {State} {Spaces}.
\newblock In \emph{Thirty-Second AAAI Conference on Artificial Intelligence},
  volume~32. AAAI, 2018.
\newblock URL \url{https://ojs.aaai.org/index.php/AAAI/article/view/11485}.

\bibitem[Wessel(2012)]{wessel_erroraware_2012}
Jan~R. Wessel.
\newblock Error awareness and the error-related negativity: evaluating the
  first decade of evidence.
\newblock 6:\penalty0 88, April 2012.
\newblock ISSN 1662-5161.
\newblock \doi{10.3389/fnhum.2012.00088}.
\newblock URL \url{https://www.ncbi.nlm.nih.gov/pmc/articles/PMC3328124/}.

\bibitem[Wirth et~al.(2017)Wirth, Akrour, Neumann, and
  F{\"u}rnkranz]{wirth_preference_2017}
Christopher Wirth, Riad Akrour, Gerhard Neumann, and Johannes F{\"u}rnkranz.
\newblock A survey of preference-based reinforcement learning methods.
\newblock 18\penalty0 (1):\penalty0 4945--4990, January 2017.
\newblock ISSN 1532-4435.

\bibitem[Wirth et~al.(2020{\natexlab{a}})Wirth, Toth, and
  Arvaneh]{wirth_4way_2020}
Christopher Wirth, Jake Toth, and Mahnaz Arvaneh.
\newblock Four-{Way} {Classification} of {EEG} {Responses} {To} {Virtual}
  {Robot} {Navigation}.
\newblock In \emph{2020 42nd {Annual} {International} {Conference} of the
  {IEEE} {Engineering} in {Medicine} {Biology} {Society}}, EMBC '20, pages
  3050--3053. IEEE, July 2020{\natexlab{a}}.
\newblock \doi{10.1109/EMBC44109.2020.9176230}.

\bibitem[Wirth et~al.(2020{\natexlab{b}})Wirth, Toth, and
  Arvaneh]{wirth_goal_2020}
Christopher Wirth, Jake Toth, and Mahnaz Arvaneh.
\newblock {You} {Have} {Reached} {Your} {Destination}: {A} {Single} {Trial}
  {EEG} {Classification} {Study}.
\newblock 14:\penalty0 66, 2020{\natexlab{b}}.
\newblock ISSN 1662-453X.
\newblock \doi{10.3389/fnins.2020.00066}.
\newblock URL
  \url{https://www.frontiersin.org/articles/10.3389/fnins.2020.00066/full?report=reader}.

\bibitem[Xiao et~al.(2020)Xiao, Lu, Ramasubramanian, Clark, Bushnell, and
  Poovendran]{xiao_fresh_2020}
Baicen Xiao, Qifan Lu, Bhaskar Ramasubramanian, Andrew Clark, Linda Bushnell,
  and Radha Poovendran.
\newblock {FRESH}: Interactive reward shaping in high-dimensional state spaces
  using human feedback.
\newblock In \emph{Proceedings of the 19th International Conference on
  Autonomous Agents and {MultiAgent} Systems}, {AAMAS} '20. International
  Foundation for Autonomous Agents and Multiagent Systems, 2020.
\newblock ISBN 978-1-4503-7518-4.

\bibitem[Xu et~al.(2021)Xu, Agarwal, Gupta, Fekri, and
  Sivakumar]{xu_accelerating_2021}
Duo Xu, Mohit Agarwal, Ekansh Gupta, Faramarz Fekri, and Raghupathy Sivakumar.
\newblock Accelerating reinforcement learning using eeg-based implicit human
  feedback.
\newblock 460:\penalty0 139--153, 2021.
\newblock ISSN 0925-2312.
\newblock \doi{https://doi.org/10.1016/j.neucom.2021.06.064}.
\newblock URL
  \url{https://www.sciencedirect.com/science/article/pii/S0925231221009887}.

\bibitem[Yger et~al.(2017)Yger, Berar, and Lotte]{yger_rgbcisurvey_2017}
Florian Yger, Maxime Berar, and Fabien Lotte.
\newblock Riemannian {Approaches} in {Brain}-{Computer} {Interfaces}: {A}
  {Review}.
\newblock 25\penalty0 (10):\penalty0 1753--1762, October 2017.
\newblock ISSN 1558-0210.
\newblock \doi{10.1109/TNSRE.2016.2627016}.

\bibitem[Zhang et~al.(2018)Zhang, Liu, Zhang, Whritner, Muller, Hayhoe, and
  Ballard]{zhang_agil_2018}
Ruohan Zhang, Zhuode Liu, Luxin Zhang, Jake~A. Whritner, Karl~S. Muller,
  Mary~M. Hayhoe, and Dana~H. Ballard.
\newblock Agil: Learning attention from human for visuomotor tasks.
\newblock In \emph{European Conference on Computer Vision}, pages 692--707,
  Cham, 2018. Springer International Publishing.
\newblock ISBN 978-3-030-01252-6.

\bibitem[Zhang et~al.(2019{\natexlab{a}})Zhang, Torabi, Guan, Ballard, and
  Stone]{zhang_irlsurvey_2019}
Ruohan Zhang, Faraz Torabi, Lin Guan, Dana~H. Ballard, and Peter Stone.
\newblock Leveraging {Human} {Guidance} for {Deep} {Reinforcement} {Learning}
  {Tasks}.
\newblock In \emph{Proceedings of the {Twenty}-{Eighth} {International} {Joint}
  {Conference} on {Artificial} {Intelligence}}, pages 6339--6346, August
  2019{\natexlab{a}}.
\newblock ISBN 978-0-9992411-4-1.
\newblock \doi{10.24963/ijcai.2019/884}.
\newblock URL \url{https://www.ijcai.org/proceedings/2019/884}.

\bibitem[Zhang et~al.(2020)Zhang, Walshe, Liu, Guan, Muller, Whritner, Zhang,
  Hayhoe, and Ballard]{zhang_head_2019}
Ruohan Zhang, Calen Walshe, Zhuode Liu, Lin Guan, Karl Muller, Jake Whritner,
  Luxin Zhang, Mary Hayhoe, and Dana Ballard.
\newblock Atari-head: Atari human eye-tracking and demonstration dataset.
\newblock 34\penalty0 (04):\penalty0 6811--6820, 2020.
\newblock \doi{10.1609/aaai.v34i04.6161}.

\bibitem[Zhang et~al.(2019{\natexlab{b}})Zhang, Yao, Wang, Monaghan, Mcalpine,
  and Zhang]{zhang_bcisurvey_2019}
Xiang Zhang, Lina Yao, Xianzhi Wang, Jessica Monaghan, David Mcalpine, and
  Yu~Zhang.
\newblock A {Survey} on {Deep} {Learning} based {Brain} {Computer} {Interface}:
  {Recent} {Advances} and {New} {Frontiers}, October 2019{\natexlab{b}}.

\end{thebibliography}
\end{document}